\documentclass{article}

\usepackage[numbers]{natbib}

\usepackage[nonatbib,preprint]{neurips_2024}
\usepackage{sidecap, caption}
\usepackage{wrapfig}
\usepackage{subcaption}

\usepackage{amsmath}
\usepackage{amssymb}
\usepackage{mathtools}
\usepackage{amsthm}

\usepackage{thm-restate}

\theoremstyle{plain}

\theoremstyle{definition}

\theoremstyle{remark}

\usepackage{algorithm}
\usepackage[noend]{algpseudocode}
\usepackage{wrapfig}

\usepackage{verbatim}
\usepackage{multirow}
\usepackage{multicol}
\usepackage{booktabs, multirow} 
\usepackage{soul}
\usepackage[table]{xcolor} 
\usepackage{changepage,threeparttable} 
\usepackage{thmtools}
\usepackage{graphicx}

\usepackage{xspace}

\newcommand{\customfootnotetext}[2]{{%
  \renewcommand{\thefootnote}{#1}%
  \footnotetext[0]{#2}}}%

\usepackage{tikz}

\usepackage{pifont}

\definecolor{darkgreen}{rgb}{0,0.5,0}
\newcommand{\cmark}{\textcolor{darkgreen}{\ding{51}}}
\newcommand{\xmark}{\textcolor{red}{\ding{55}}}
\definecolor{newblue}{rgb}{0.21,0.49,0.74}

\usepackage[pagebackref,breaklinks,colorlinks,allcolors=newblue]{hyperref}
\newcommand{\videophy}[1]{\textsc{VideoPhy}\xspace}
\newcommand{\flash}[1]{Gemini-2.0-Flash-Exp\xspace}
\newcommand{\name}[1]{\textsc{VideoPhy-2}\xspace}
\newcommand{\auto}[1]{\textsc{VideoPhy-2-Autoeval}\xspace}

\definecolor{backblue}{RGB}{210, 230, 250}
\definecolor{lightyellow}{cmyk}{0, 0.0, 0.3, 0}

\newcommand{\best}{\cellcolor{backblue}}
\newcommand{\second}{\cellcolor{lightyellow}}

\title{\name{}: A Challenging Action-Centric Physical Commonsense Evaluation in Video Generation}

\author{
\normalsize Hritik Bansal$^{*1}$
\hspace{0.4em}
\normalsize Clark Peng$^{*1}$
\hspace{0.4em}
\normalsize Yonatan Bitton$^{*2}$\\ 
\normalsize \textbf{Roman Goldenberg}$^{2}$ 
\hspace{0.4em}
\normalsize \textbf{Aditya Grover}$^{1}$
\hspace{0.4em}
\normalsize \textbf{Kai-Wei Chang}$^{1}$\\\\
\textbf{$^{1}$University of California Los Angeles} 
\hspace{0.5em}
\textbf{$^{2}$Google Research}\\
}

\begin{document}
\customfootnotetext{}{$^{*}$ Equal Contribution.}
\maketitle
\begin{abstract}

Large-scale video generative models, capable of creating realistic videos of diverse visual concepts, are strong candidates for general-purpose physical world simulators. However, their adherence to physical commonsense across real-world actions remains unclear (e.g., playing tennis, backflip). Existing benchmarks suffer from limitations such as limited size, lack of human evaluation, sim-to-real gaps, and absence of fine-grained physical rule analysis. To address this, we introduce \name{}, an action-centric dataset for evaluating physical commonsense in generated videos. We curate 200 diverse actions and detailed prompts for video synthesis from modern generative models. We perform human evaluation that assesses semantic adherence, physical commonsense, and grounding of physical rules in the generated videos. Our findings reveal major shortcomings, with even the best model achieving only $22\%$ joint performance (i.e., high semantic and physical commonsense adherence) on the hard subset of \name{}. We find that the models particularly struggle with conservation laws like mass and momentum. Finally, we also train \auto{}, an automatic evaluator for fast, reliable assessment on our dataset. Overall, \name{} serves as a rigorous benchmark, exposing critical gaps in video generative models and guiding future research in physically-grounded video generation. The data and code is available at \url{https://videophy2.github.io/}.
\end{abstract}

\sidecaptionvpos{figure}{c}
\begin{SCfigure}[50][h]
    \includegraphics[scale=0.52]{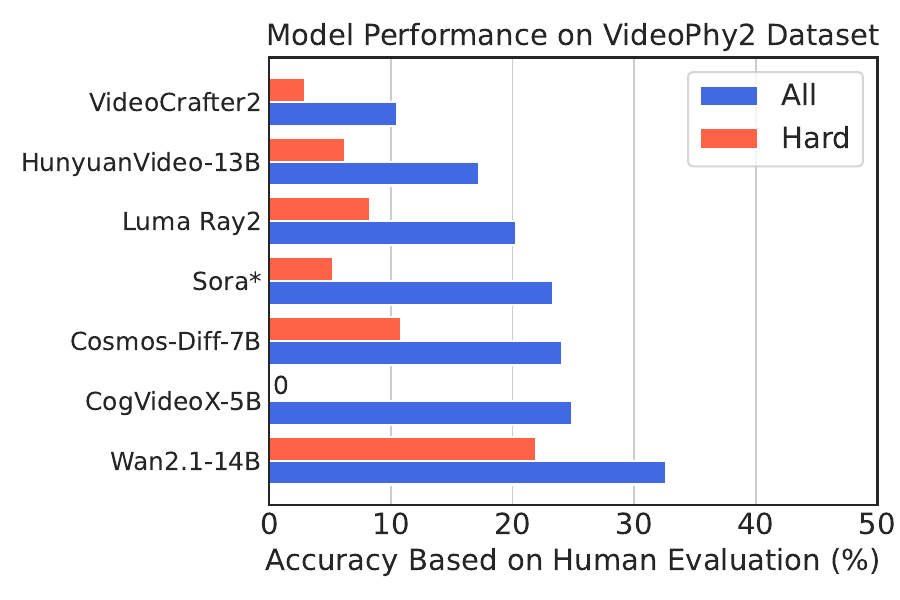}
    \caption{\textbf{Performance on the \name{} dataset using human evaluation.} We evaluate the physical commonsense and semantic adherence to text conditioning prompts for  diverse real-world actions. We observe that even the best-performing model Wan2.1-14B achieves $32.6\%$ and $22\%$ on the entire and hard subset of the data, respectively. * represents the evaluation on a very small subset of the dataset.}
    \label{fig:teaser_graph}
\end{SCfigure}

\begin{figure*}[htbp]
    \centering
    \includegraphics[width=\textwidth]{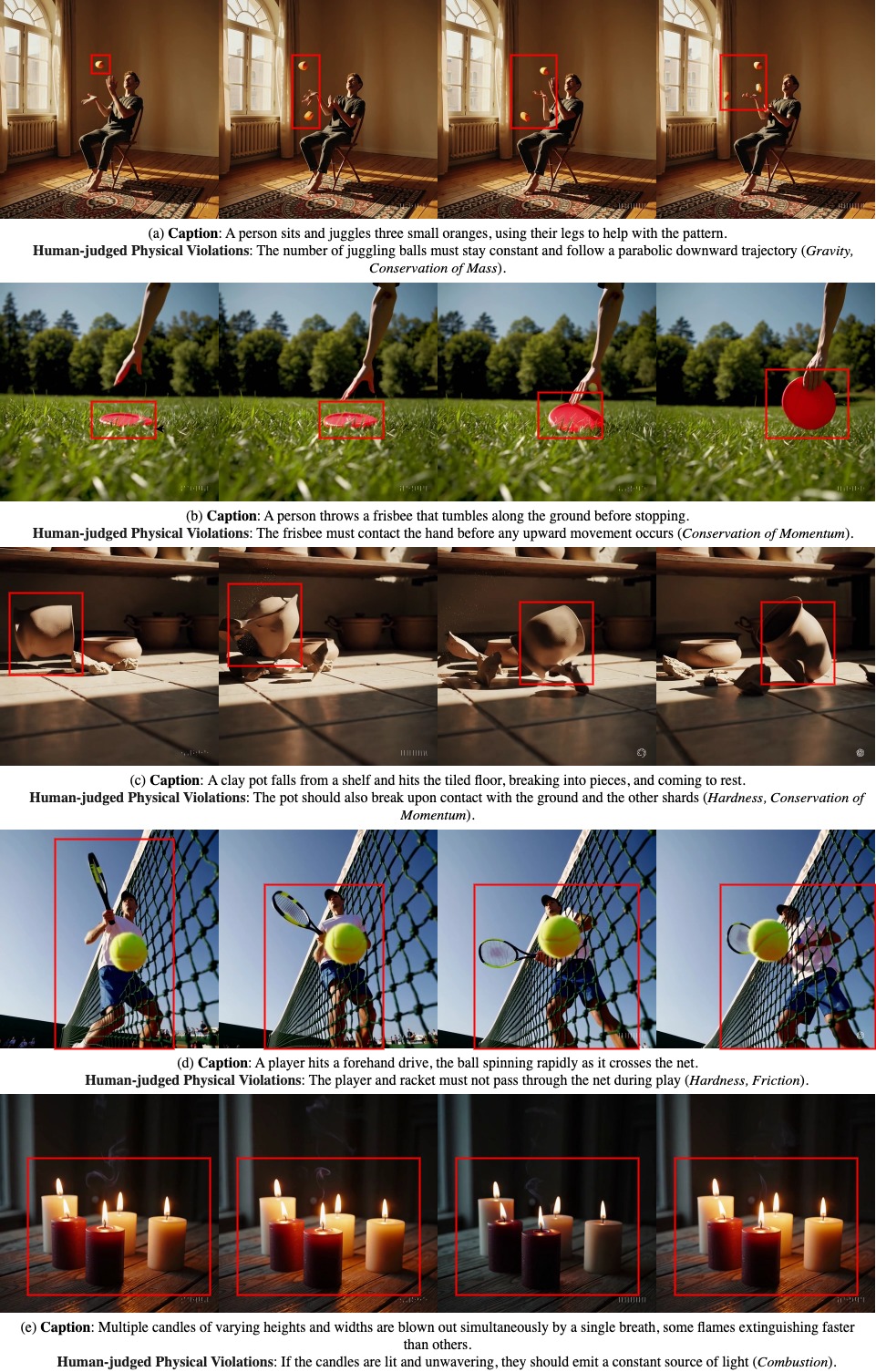}
    \caption{\textbf{Examples of physically unlikely video generations from Sora.} Each case demonstrates violations of physical rules and their laws.}
    \label{fig:sora_full_bad_examples}
\end{figure*}   

\begin{figure*}[t]
    \centering
    \includegraphics[width=0.9\linewidth]{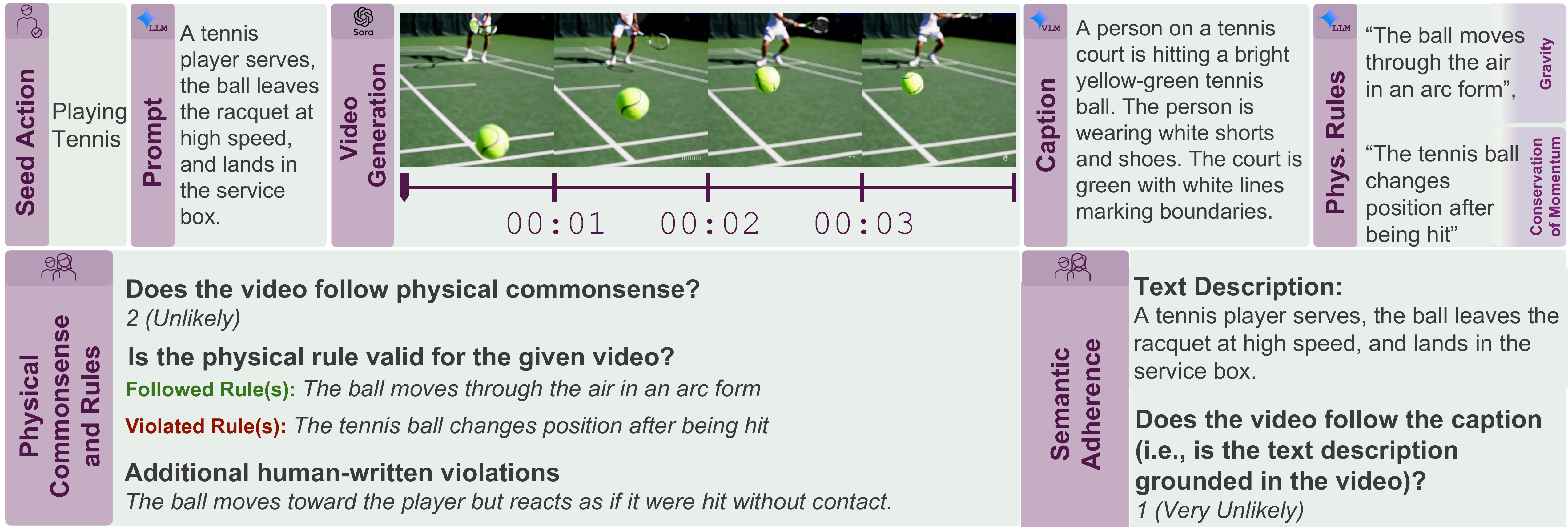}
    \caption{\small{\textbf{\name{} pipeline.} \textbf{Top:} We generate a text prompt from the seed action using an LLM, create a video with a text-to-video model, and caption it with a VLM to extract candidate physical rules. \textbf{Bottom:} Human annotators rate the video's physical likelihood, verify rule violations, suggest missing rules, and assess semantic adherence to the input prompt.}}
    \label{fig:pipeline}
\end{figure*}

\section{Introduction}
\label{sec:intro}

Recent advancements in large-scale video generative modeling offer the potential to simulate the physical world accurately \cite{videoworldsimulators2024,deepmindveo2}. In particular, this capability can enable learning general-purpose visuomotor policies \cite{liang2024dreamitate,du2024learning}, predicting future frames for robotic manipulation \cite{agarwal2025cosmos}, autonomous driving \cite{agarwal2025cosmos}, and game playing \cite{bruce2024genie,deepmindGenie2}. In daily life, humans rely on their sophisticated physics intuition to interact with the world \cite{duan2022survey} (e.g., predicting the trajectory of football after being hit). However, the extent to which existing video models can generate physically likely worlds across diverse real-world actions remains unclear.


A naive approach to evaluating generated videos is to compare them with ground-truth physical simulations \cite{agarwal2025cosmos, qin2024worldsimbench}. Furthermore, there is a lack of mature methods for rendering diverse real-world materials \cite{bansal2024videophy, klar2016drucker, o2002graphical} and for accurately simulating complex physical interactions \cite{liu2024physgen}. For instance, simulating a scenario like `a child kicking a ball against a wall' requires precise estimation of the foot’s pose and geometry relative to the ball at impact, an exact model of the kick’s dynamics, and considerations of the ball's air pressure and material properties. While we focus on evaluating the physical likelihood of generated videos, an assessment that can often be made by humans without formal physics education by relying on their real-world experience.


Recent work such as Physics-IQ \cite{motamed2025generative} conditions video models on the first few frames of real videos and evaluates their similarity by comparing predicted videos with ground-truth completions. However, this approach faces several challenges: (a) the extent to which it agrees with human judgment remains unclear, and (b) extending it to more complex scenarios depicting multiple events is non-trivial.
Another work PhyGenBench \cite{liu2024physgen} curates a small set of 160 manually crafted prompts, which is not scalable. Additionally, their evaluation approach simplifies the problem by designing text prompts that explicitly associate with a single physical law (e.g., `A stone placed on the surface of a water pool' is linked to law of Buoyancy). Although, this strict one-to-one association between a prompt and a physical law is problematic, as video models often exhibit imperfect semantic adherence. For instance, a video model might generate a video that does not strictly follow the prompt but still adheres to physical commonsense (e.g., producing a video where `a stone is dropped from a height into the pool', where gravity is more crucial than buoyancy). 
Further, \videophy{} \cite{bansal2024videophy} focuses on semantic adherence and physical commonsense on video generation for diverse material types and their interactions. However, it does not provide insights into the physical law violations in the videos. We note the difference between \name{} and existing work in Appendix Table \ref{tab:related_work_comparison}.


To address these gaps, we propose \name{}, a challenging physical commonsense evaluation dataset for real-world actions. Specifically, we curate a list of \textbf{197 actions} across diverse physical activities (e.g., hula-hooping, playing tennis, gymnastics) and object interactions (e.g., bending an object until it breaks). Then, we generate $3940$ detailed prompts from these seed actions using a large language model (LLM). Further, these prompts are used to synthesize videos with modern video generative models. Finally, we compile a list of \textbf{candidate physical rules} (and laws) that should be satisfied in the generated videos, using vision-language models in the loop. For example, in a video of \textit{sportsperson playing tennis}, a physical rule would be that \textit{a tennis ball should follow a parabolic trajectory under gravity}. For gold-standard judgments, we ask human annotators to score each video based on overall semantic adherence and physical commonsense, and to mark its compliance with various physical rules. We present the entire pipeline in Figure \ref{fig:pipeline}.

In our experiments, we find that the best-performing model, Wan2.1-14B \cite{wan14b}, achieves a joint performance score (high semantic adherence and physical commonsense) of only $32.6\%$. 
To further increase the dataset's difficulty, we create a \textbf{hard subset} which decreases the performance of 
Wan2.1-14B drops from $32.6\%$ to $22\%$. Furthermore, our fine-grained analysis of human-annotated physical rule violations reveals that video models struggle the most with \textit{conservation laws}, such as those governing mass and momentum. Overall, we demonstrate that \name{} is a high-quality dataset that presents a formidable challenge for modern video models. 

While human evaluation serves as the gold standard for real-world physical commonsense judgment, it is expensive and difficult to scale. To address this, we train an automatic evaluation model, \textbf{\auto{}}, capable of performing a wide range of tasks—including scoring semantic adherence, physical commonsense, and classifying physical rule grounding in the generated video. In our experiments, we find that \auto{} outperforms a capable multimodal foundation model, \flash{} \citep{GeminiFlash2p0}, with a relative correlation improvement of $81\%$ and $236\%$ on the semantic adherence and physical commonsense tasks, respectively, on the unseen prompts. 
Overall, our dataset represents a significant improvement over prior work and lays the foundation for assessing next-generation physical simulators on real-world tasks.












\begin{figure*}[t]
 \begin{minipage}{0.45\textwidth} 
 \centering
 \captionof{table}{\small{\textbf{Data statistics.} We present the number of instances for diverse features (e.g., captions,) in the \name{}.}}
 \label{tab:statistics}
\begin{tabular}{ccc}\toprule
\textbf{Feature} &\textbf{Number} \\\hline

Captions & $3940$ \\
Unique actions & $197$ \\
Generated videos & $6800$ \\
Human annotations & $102$K \\ 
\hline
Avg. words in original caption & $16$ \\
Avg. words in upsampled caption & $138$\\
\hline
\textit{Category}  & \textit{\# Actions} \\\hline
Sports and Physical Activities & $143$ \\
Object Interactions &  $54$\\
Hard subset & $60$ \\
\bottomrule
\end{tabular}
 \end{minipage} 
 \hfill
 \begin{minipage}{0.5\textwidth}
 \centering
\includegraphics[width=0.9\textwidth]{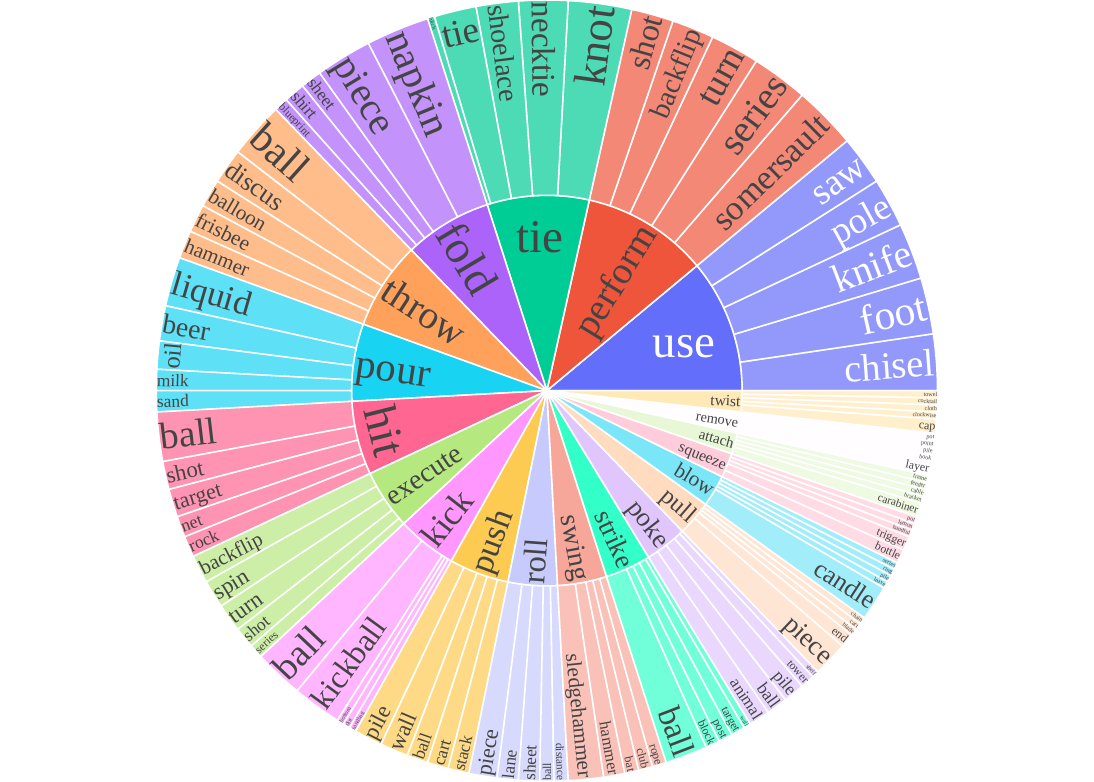}
 \caption{\small{\textbf{Diversity of \name{} prompts.} Top-20 frequently occurring verbs (inner) and their top-5 direct nouns (outer).}}
 \label{fig:source_dataset_vn}
 \end{minipage}
\end{figure*}

\section{\name{} Dataset}

\label{sec:dataset}


In this work, \name{} aims to assess the ability of modern text-to-video generative models to adhere to the input prompt and generate physically realistic videos. 
We present the steps for data construction below:

\paragraph{Seed Actions (Stage 1):} First, we curate a set of actions relevant to physical commonsense evaluation. Specifically, we compile a diverse list of over 600 actions from popular video datasets that capture a wide range of real-world activities, particularly those involving sports, physical activities, and object interactions. These datasets include Kinetics \cite{carreira2018short}, UCF-101 \cite{soomro2012ucf101}, and SSv2 \cite{goyal2017something}. Next, we divide the student authors, with undergraduate or more degree in STEM, into two groups, each of which independently reviews the list and marks actions deemed relevant for physical commonsense evaluation. Our goal is to include actions that test various physical laws (e.g., gravity, elasticity, buoyancy, reflection, conservation of mass and momentum). Importantly, we filter out actions that do not elicit significant motion or are unlikely to be compelling for physical commonsense evaluation in videos (e.g., typing, applying cream, arguing, auctioning, chewing, playing instruments, petting a cat). Finally, we retain only the actions deemed relevant by both groups of annotators. After this filtering process, we obtain a list of 232 actions, which we further refine using \flash{} to remove semantic duplicates, resulting in a final set of \textbf{197} actions. Among these, 54 actions focus on object interactions, while 143 pertain to physical and sports activities. We present the list of all the actions in Appendix Table \ref{app_table:all_actions_list}.

\paragraph{LLM-Generated Prompts (Stage 2):} In this stage, we query the \flash{} LLM to independently generate 20 prompts for each action in our dataset. Specifically, the prompt generation follows several key principles: (a) focus on visible physical interactions between objects that can be clearly grounded in a generated video (e.g., an arrow hitting a target), (b) exclusion on non-visual details, such as mental states (e.g., emotions and intent), sensory details (e.g., smells and sounds), and abstract or poetic language that does not translate into a clear visual representation, (c) incorporating diverse characters and objects, and (d) depiction of multiple events within a prompt to increase the challenge for modern video generation models (e.g., we encourage the LLM to generate `An archer draws the bowstring back to full tension, then releases the arrow, which flies straight and strikes a bullseye on a paper target' instead of a simpler prompt `An archer releases the arrow'). 
Our prompt generation template is presented in Appendix \ref{app:llm_generated_caption_prompt}. In total, we curated \textbf{$3940$} prompts covering a wide range of actions. Since the modern video models can understand long video descriptions, we also generate dense captions from the original captions using the Mistral-NeMo-12B-Instruct prompt upsampler from \cite{agarwal2025cosmos}. In particular, these dense captions add more visual details to the original caption without changing its semantic meaning (e.g., main characters and actions). We present some of the generated captions and underlying actions in Appendix Table \ref{tab:action_category_examples}, and the upsampled captions in Appendix Table \ref{tab:upsampled_prompts}.

\paragraph{Candidate physical rules and laws (Stage 3):} In this stage, we aim to generate candidate physical rules and associated laws that could be followed (or violated) in the generated video. Since video models often struggle to adhere to conditioning text prompts, we do not derive physical rules directly from them. Instead, we first generate videos using generative models conditioned on prompts from the \name{} dataset. Then, we create captions for these videos using the strong video captioning capabilities of \flash{}. This ensures that the physical rules are constructed based on details grounded in the video itself.\footnote{We observe that prompting \flash{} to generate physical rules directly from the video did not yield high-quality outputs. Therefore, we prefer a two-step process: first captioning the video, then generating the rules.} Subsequently, we ask \flash{} to generate a set of three physical rules (and laws) that should be followed for a given video. Since a video may violate physical rules that are not covered in the pre-defined rules, we further ask the human annotators to write additional violated rules during physical commonsense evaluation. We present the rule generation prompt in Appendix Table \ref{tab:video_physical_rule_prompt}. 

\paragraph{Construction of the Hard Subset (Stage 4):} While we collect diverse and lengthy captions to make the task more challenging, we further employ a model-based strategy to identify a subset of particularly difficult actions. Specifically, we generate videos using a strong open video model, CogVideoX-5B \cite{yang2024cogvideox}, conditioned on captions from the \name{} dataset. From this, we select \textbf{$60$} actions (out of 197) for which the model fails to generate videos that accurately adhere to the prompts and follow physical commonsense (Appendix Table \ref{app_table:hard_actions_list}). On examination, we find that these actions focus on physics-rich interactions (e.g., momentum transfer in throwing discus or passing football), state changes (e.g., bending something until it breaks), balancing (e.g., tightrope walking), and complex motions (e.g., backflip, pole vault, and pizzatossing). In total, we designate \textbf{$1200$} prompts making the dataset more challenging. We present the list of hard actions in Appendix Table \ref{app_table:hard_actions_list}.\footnote{We note that a similar model-based strategy is also adopted in recent works like Humanity's Last Exam \cite{phan2025humanity} and ZeroBench \cite{roberts2025zerobench} to collect hard instances for model evaluation.}

\paragraph{Data Analysis:} We present the dataset statistics in Table \ref{tab:statistics}. Specifically, \name{} contains $3940$ captions, which is $5.72\times$ more than those in the \videophy{} dataset. Additionally, the average lengths of original and upsampled captions are $16$ and $138$ tokens, respectively—$1.88\times$ and $16.2\times$ longer than those in \videophy{}. Furthermore, \name{} includes $102$K human annotations across various video generative models and their semantic adherence, physical commonsense, and physical rule annotations. Finally, we show the distribution of the root verbs and direct nouns in the original captions of \name{} in Figure \ref{fig:source_dataset_vn}, demonstrating the high diversity of the dataset. We also illustrate the diversity of multiple captions for a specific action in Appendix Figure \ref{app_fig:verb_noun_specific_actions}. Overall, our analysis highlights that \name{} significantly enhances data diversity and richness compared to its counterparts. 


 

\section{Evaluation}
\label{sec:evaluation}

\subsection{Metric}
\label{sec:eval_metrics}

In practice, we want the generated video to adhere to several constraints, including high video quality \cite{liu2024evalcrafter}, temporal consistency \cite{huang2024vbench}, entity and background consistency \cite{bansal2024talc}. While many of these metrics are intertwined, it is crucial to evaluate each one independently to gain a better understanding of the model’s capabilities. In this regard, we focus on the extent to which the generated video adheres to the input text prompt and follows physical commonsense. Similar to \cite{bansal2024videophy,liu2024physgen,liu2024evalcrafter}, we prioritize rating-based feedback, as it quantifies the mode of failure or success for individual videos. In contrast, ranking-based feedback does not measure the magnitude of the difference between two videos but simply indicates which one is preferred.
Unlike prior work \cite{bansal2024videophy}, we collect dense rating feedback on a 5-point scale, allowing human annotators to express their judgments in a more fine-grained manner. Furthermore, we extend previous studies by evaluating whether generated videos adhere to diverse physical rules and their laws.

\paragraph{Semantic Adherence (SA):} Here, we aim to assess whether the input text prompt is semantically grounded in the generated video. Specifically, it studies whether the entities, actions, and relationships described in the prompt are accurately depicted in the video (e.g., a person visibly jumping into the water). To measure semantic adherence, annotators rate each video on a 5-point scale, selecting from the following options: \{$SA \in$ \textit{Very Unlikely (1), Unlikely (2), Neutral (3), Likely (4), Very Likely  (5)}\}. In this case, \textit{very unlikely} indicates that the video does not match the prompt at all, and \textit{very likely} highlights the video fully adheres to the prompt with no inconsistencies.

\paragraph{Physical Commonsense (PC):} Here, our goal is to assess whether the generated video follows the physical laws of the real-world intuitively (e.g., the football should start moving after impact in accordance with newton's first law). We note that the physical commonsense evaluation is independent of the underlying video generating text prompt. Since a video can follow (or violate) numerous laws, we are concerned with the holistic sense of the video's physical commonsense. In particular, the annotators rate each video on a 5-point scale, selecting from the following options: \{$PC \in$ \textit{Very Unlikely (1), Unlikely (2), Neutral (3), Likely (4), Very Likely  (5)}\}. Here, \textit{very unlikely} that the video contains numerous violations of fundamental physical laws, and \textit{very likely} indicates that the video demonstrates a strong understanding of physical commonsense with no violations. 

Similar to \cite{bansal2024videophy}, we compute \textbf{joint performance} as the main evaluation metric, which measures the fraction of videos that both adhere closely to the text prompt ($SA \geq 4$) and follow physical commonsense to a high degree ($PC \geq 4$). We do not report the posterior score ($PC>=4 | SA>=4$) since a bad model can game it.\footnote{For example, a model can adhere to the prompt for 1 out of 1000 prompts in the dataset. Now, assume that this video is also physically realistic. Then, the posterior performance of this model will be $100\%$ that is quite misleading for the model builders.}

\paragraph{Physical Rules (PR):} A key feature of the \name{} dataset is the collection of candidate physical rules (and their associated laws) that humans evaluate as being followed or violated in the generated video (e.g., `the ball should go down' is a physical rule associated with the law of gravity). These rules enable a fine-grained assessment of the video model's capabilities. Specifically, we determine whether a candidate physical rule is \textit{violated (0)}, \textit{followed (1)}, or \textit{cannot be determined (2)} in the generated video.\footnote{We include CBD category because LLM-generated physical rules may not always be visually grounded in the video.} Further, we ask human annotators to note more physical rule violations to ensure comprehensive coverage.

\subsection{Human Evaluation}
\label{sec:human_eval}

In practice, human evaluation serves as a gold standard for assessing the quality of generative foundation models \cite{ChatbotArena,yarom2023you}. 
In particular, we collect judgments using the Amazon Mechanical Turk (AMT) platform from a group of $12$ human annotators, which were selected after passing a qualification test. 
To promote high-quality annotations, we communicated with annotators remotely to provide detailed instructions and analyze annotation examples.\footnote{We pay a wage of \$18 per hour to our annotators.} Since physical commonsense is independent of the generated video-prompt alignment, we evaluate semantic adherence and physical commonsense (including rule-based judgment) as separate tasks for human annotators. This differs from prior work in \videophy{} \cite{bansal2024videophy}, which treats semantic adherence and physical commonsense assessment as a single task. It may introduce evaluation bias, as annotators have access to the prompt while conducting the physical commonsense evaluation, a scenario we explicitly avoid in this work.

We present the annotation UI for the semantic adherence task in Appendix Figure \ref{fig:interface_sa}, where the input consists of a text prompt and the corresponding generated video. Note that human annotators were shown the original prompt (not the upsampled prompts) to ensure a fair comparison between video models, regardless of their ability to handle short or long prompts. In the following task, human annotators are asked to evaluate only the generated video and with regard to adherence to specific physical rules (followed/violated/cannot be determined), overall physical commonsense (rated on a scale of 1-5), and observable behaviors that violate physical reality.\footnote{In our instructions to the annotators, we explicitly clarify that the overall physical commonsense judgments should extend beyond the predefined physical rules listed in the task.} The annotation interface for this task is shown in Appendix Figure \ref{fig:interface_pc}. 


\subsection{Automatic Evaluation}
\label{sec:automatic_eval}

While human judgments serve as the gold standard, automating the evaluation process is crucial for faster and more cost-effective model assessments. In this study, we evaluate several video-language foundation models (e.g., \flash{}, VideoScore \cite{he2024videoscore}) on two tasks: semantic adherence and physical commonsense scoring. Specifically, we prompt the models to score generated videos based on these two criteria and then normalize their predictions to a 5-point scale. We provide more details about score computation in Appendix \ref{app:baseline_score_computation}. Additionally, we introduce a classification task to determine whether a given physical rule is followed, violated, or indeterminate (CBD) in the generated video, leveraging video-language models such as VideoLLaVA \cite{lin2023video}. In this task, we prompt the model to classify each video-rule pair into one of three categories: followed, violated, or CBD. 

Our experiments reveal that existing video-language models struggle to achieve strong agreement with human annotators. This discrepancy primarily arises due to their limited understanding of physical commonsense and rules, as well as the complexity of the prompts. Hence, we supplement our benchmark with a video-language model \auto{} (7B parameters). Specifically, we aim to provide more accurate predictions for the generated videos along three axis -- semantic adherence score (1-5), physical commonsense score (1-5), and physical rule classification (0-2). We follow a data-driven approach to distill human knowledge into a foundation model for these tasks. Specifically, we fine-tune a video-language model VideoCon-Physics \cite{bansal2024videophy} on $50$K human annotations acquired for these tasks. We train a mult-task model to solve the three tasks using a shared backbone, to allow the inter-task knowledge transfer. We provide the templates and setup used for model finetuning in Appendix \ref{sec:mm_prompts} and Appendix \ref{sec:training_details}, respectively. 
\section{Setup}
\label{sec:setup}

\paragraph{Video generative models.} In this work, we evaluate a diverse range of state-of-the-art text-to-video generative models. Specifically, we assess five open models and two closed models, including \textit{CogVideoX-5B} \cite{yang2024cogvideox}, \textit{VideoCrafter2} \cite{chen2024videocrafter2}, \textit{HunyuanVideo-13B} \cite{kong2024hunyuanvideo}, \textit{Cosmos-Diffusion-7B} \cite{agarwal2025cosmos}, \textit{Wan2.1-14B} \cite{wan14b}, \textit{OpenAI Sora} \cite{videoworldsimulators2024}, and \textit{Luma Ray2} \cite{lumalabsLumaRay2}.\footnote{We exclude other closed models due to lack of API access (e.g., Veo2 \cite{deepmindveo2},  Kling \cite{klingai}).} We prompt these models with the upsampled captions, except for those that do not support long (dense) captions. Specifically, Hunyuan-13B and VideoCrafter2 are limited to 77 tokens due to their reliance on the CLIP \cite{radford2021learning} text encoder. Additionally, we generate short videos (less than 6s) as they are easier to evaluate and effectively highlight challenges on the \name{}. The model inference details are provided in Appendix \ref{app:inference_details}.

\paragraph{Dataset setup.} Similar to \cite{bansal2024videophy}, we take a data-driven approach and use human annotations across multiple tasks to train the automatic evaluator. We split the \name{} dataset into a test set for benchmarking and a training set for training the \auto{} model. Specifically, the training and testing prompts consist of $3350$ ($197$ actions × $17$ captions per action) and $590$ ($197$ actions × $3$ captions per action) prompts, respectively. 

\begin{table}[t]
\centering
\caption{\small{\textbf{Human evaluation results on \name{}.} We present the joint performance that focuses on high semantic adherence and high physical commonsense in the generated videos. Hard, PA, OI refer to the hard, physical activities, and object interactions subsets of the data, respectively. We mark the best performing models in each column by \best{blue} and second best by \second{yellow}.}}
\label{tab:main_results}
\begin{tabular}{lc|cc|cc}
\toprule
\textbf{Model}   & \textbf{Class}     & \textbf{All}  & \textbf{Hard} & \textbf{PA}   & \textbf{OI}   \\\hline
Wan2.1-14B  \cite{wan14b} &Open& \best{32.6} & \best{21.9}  & \best{31.5} & \best{36.2} \\
CogVideoX-5B  \cite{yang2024cogvideox} &Open& \second{25.0} & 0.0  & \second{24.6} & 26.1 \\
Cosmos-Diff-7B \cite{agarwal2025cosmos}   &Open    & 24.1 & \second{10.9} & 22.6 & \second{27.4} \\
Hunyuan-13B \cite{kong2024hunyuanvideo} &Open& 17.2 & 6.2  & 17.6 & 15.9 \\
VideoCrafter-2 \cite{chen2024videocrafter2} &Open& 10.5 & 2.9  & 10.1 & 13.1 \\
\hline
Ray2 \cite{lumalabsLumaRay2} &Closed        & 20.3 & 8.3  & 21.0 & 18.5 \\
Sora \cite{videoworldsimulators2024} &Closed        & 23.3 & 5.3  & 22.2 & 26.7 \\
\bottomrule
\end{tabular}
\end{table} 

\textbf{Benchmarking.} For every tested model, we generate one video per each test prompt, that is, $590$ videos per model. For Sora, however, we generate a subset of $60$ videos (randomly selected from $590$), manually, using Sora playground\footnote{\url{https://openai.com/sora/}} due to the lack of an official API, and $394$ videos ($2$ prompts per action) for Ray2 due to the limited API budget. 
After generating the videos, we ask three annotators to evaluate them based on semantic adherence, overall physical commonsense, and violations of various physical rules. Annotators can also suggest additional physical rules that may be missing from our list. We observe that annotator agreement ranges from $75\%$ to $80\%$ for these tasks, which is reasonable given the task subjectivity and comparable to the agreement scores in prior work \cite{bansal2024videophy}. For every generated video, we compute the SA and PC scores (1-5) by averaging the three annotators scores and rounding to the nearest integer. Following this, the joint score is computed to assess the quality of the generated video. We use the majority voting for determining whether the listed physical rule (and law) is followed, violated, or cannot be grounded in the generated video. Additional human-written violations are converted to a statement of a physical rule (and law) using \flash{}. With CogVideoX-5B as a strong reference model, we choose a \textit{hard} subset of $60$ actions for which it achieved a zero joint performance. In our experiments, we observe that this hard subset leads to big drop in performances in comparison to the entire data across diverse video models. In total, we collect $10.2$K, $10.2$K, and $30.6$K semantic adherence, physical commonsense, and physical rule annotations, which cost us \$$2600$ USD.


\paragraph{Training set for \auto{}.} In this case, we sample 1 video per caption from one of the three capable video models (HunyuanVideo-13B, Cosmos-Diffusion-7B, and CogVideoX-5B) from the training set, of size $3350$. Subsequently, we perform human annotations in the same way as the benchmarking process i.e., aggregating semantic adherence, physical commonsense and rule judgments across the three annotators. In total, we collect $\sim50K$ human annotations across the three tasks, and spend \$$3515$ USD on collecting the training data. Post-training, we compare the performance of \auto{} against several baselines on the semantic adherence and physical commonsense judgments using Pearson's correlation between the ground-truth and predicted scores. Further, we compare the joint score prediction accuracy and F1 score between our auto-rater and selected baselines. In addition, we compare the physical rule classification accuracy between the \auto{} and baselines.

\begin{figure}[t]
    \centering \includegraphics[width=0.7\linewidth]{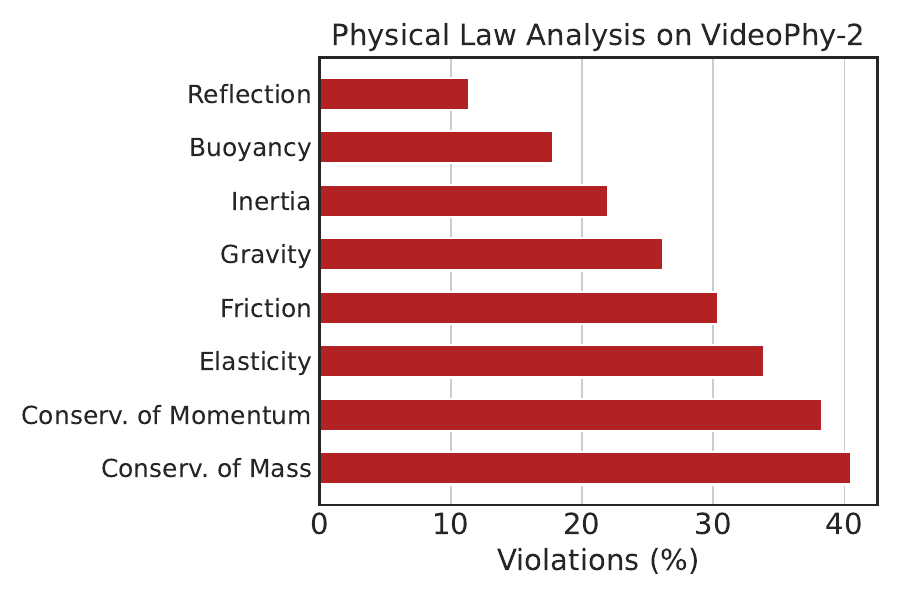}
    \caption{\textbf{Physical laws violation analysis.} We present the violation scores for diverse physical laws based on human annotations collected from various video generative models on \name{}.}
    \label{fig:physics_rule_violation_analysis}
\end{figure}

\begin{figure*}[t]
    \centering
    \includegraphics[width=\textwidth]{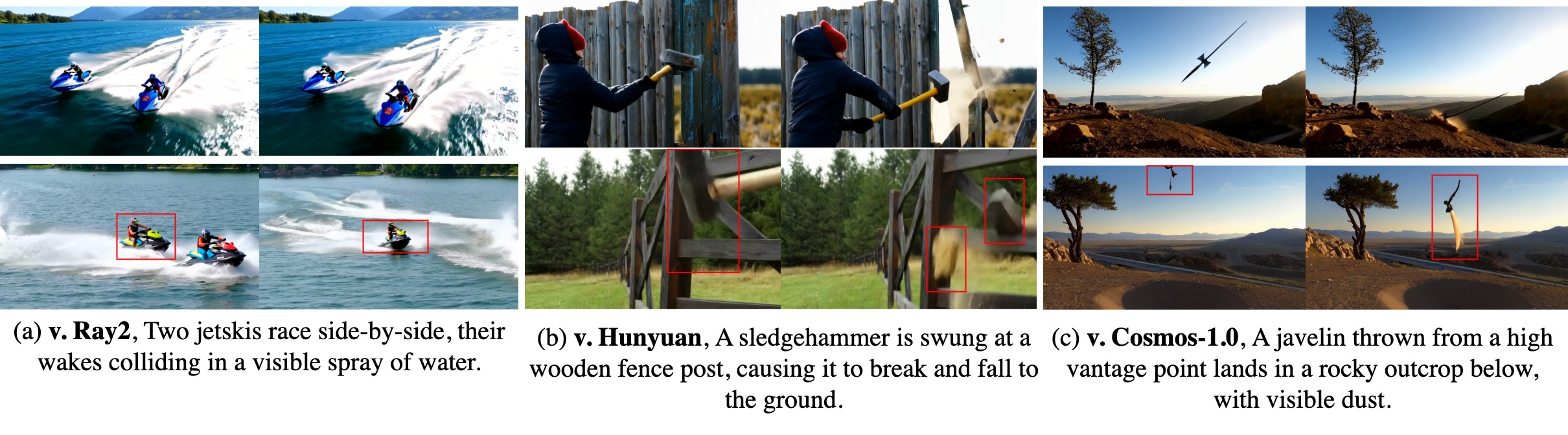}
    \caption{\small{\textbf{Comparison of Wan2.1 with other models}. The top row shows videos generated by Wan2.1: (a) For Ray2, the jetski on the left lags behind the other jetski and then starts moving backward. (b) For Hunyuan-13B, the sledgehammer deforms after the swing, and a broken wooden board appears out of nowhere. (c) For Cosmos-7B, the javelin expels sand before it even hits the ground.}}
    \label{fig:wan_comparisons_examples}
\end{figure*}

\begin{figure*}[t]
    \centering
    \includegraphics[width=\textwidth]{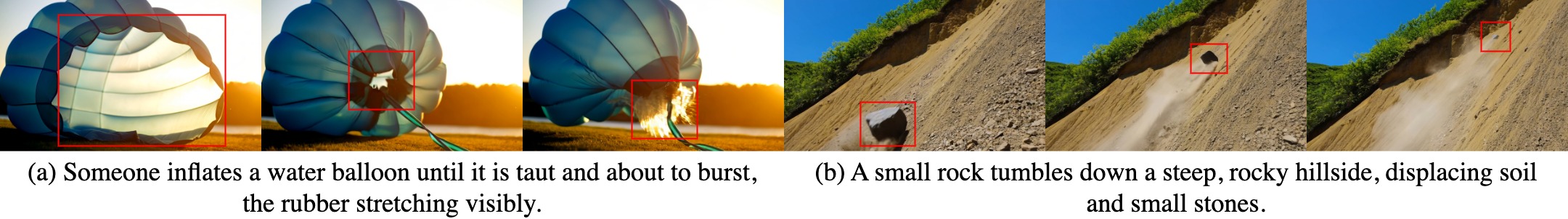}
    \caption{\small{\textbf{Illustration of Wan2.1's bad physical commonsense}. Even the best-performing model, Wan2.1, may struggle to correctly capture physical laws, leading to the generation of unnatural videos. Examples of such artifacts include: (a) A hot air balloon, which should be full of air, instead shrinking and expelling water. (b) A rock rolling and accelerating uphill instead of downhill.}}
    \label{fig:wan_limitations_examples}
\end{figure*}

\begin{table}[h]
\centering
\caption{\small{\textbf{Correlation analysis between semantic adherence and physical commonsense with other video metrics.}}}
\label{tab:correlation_sa_pc_motion_aesthetics}
\begin{tabular}{lccc}
\toprule
& \textbf{Aesthetics} & \textbf{Motion} & \textbf{SA}   \\
\hline
SA & 0.1        & 0.02   & 1    \\
PC & 0.09       & 0.002   & 0.14 \\
\bottomrule
\end{tabular}
\end{table}

\begin{table}[h]
\centering
\caption{\small{\textbf{Auto-rater evaluation results.} We present the pearson's correlation ($\times100$) between the predicted scores and ground-truth scores (1-5) on the unseen prompts and unseen video models.}}
\label{tab:autoeval_sa_pc_corr}
\begin{tabular}{lcccccc}
\toprule
                   & \multicolumn{3}{c}{\textbf{Unseen prompts}} & \multicolumn{3}{c}{\textbf{Unseen video models}} \\
           & \textbf{Avg.}    & \textbf{SA}    & \textbf{PC}    & \textbf{Avg.}     & \textbf{SA}    & \textbf{PC}    \\\hline
VideoCon-Physics \cite{bansal2024videophy}           & 28.5  & 32.0 & 25.0  & 26.5  & 27.0 & 26.0  \\
VideoCon \cite{bansal2024videocon}           & 12.5  & 23.0 & 2.0   & 8.9   & 17.0 & 0.8   \\
VideoLlava \cite{lin2023video}         & 16.0  & 30.0 & 2.0   & 19.0  & 33.0 & 5.0   \\
VideoScore  \cite{he2024videoscore}      & 13.5  & 17.0 & 10.0  & 9.0   & 5.0  & 13.0  \\
\flash{}       & 18.5  & 26.0 & 11.0  & 21.0  & 31.0 & 11.0  \\
\auto{}               & 42.0  & 47.0 & 37.0  & 41.0  & 45.0 & 37.0  \\\hline
\textit{Rel. to Best (\%)}   & \best{+47.4}  & \best{+46.9} & \best{+48.0}  & \best{+49.0}  & \best{+36.4} & \best{+61.5}  \\
\textit{Rel. to Gemini (\%)} & \best{+127.0} & \best{+80.8} & \best{+236.4} & \best{+107.1} & \best{+45.2} & \best{+281.8}\\
\bottomrule
\end{tabular}
\end{table}

\begin{table}[h]
\centering
\caption{\small{\textbf{Auto-rater evaluation on joint score judgments.} We present the joint accuracy and F1 score between the predicted scores and ground-truth scores (0-1) for our \auto{} and VideoCon-Physics.}}
\label{tab:autoeval_joint_acc}
\begin{tabular}{lcccccc}
\toprule
                     & \multicolumn{3}{c}{\textbf{Unseen prompts}} & \multicolumn{3}{c}{\textbf{Unseen video models}} \\
\textbf{Method}               & \textbf{Avg.}  & \textbf{Acc.} & \textbf{F1}   & \textbf{Avg.}   & \textbf{Acc.}   & \textbf{F1}    \\\hline
VideoCon-Physics \cite{bansal2024videophy}             & 39.1  & 75.6     & 2.6  & 39.6   & 75   & 4.2   \\
\auto{}                 & 65.1  & 79.1     & 51.1 & 62.8   & 76.3   & 49.3  \\\hline
\textit{Rel. to VideoCon-Physics (\%)} & \best{+66.4}  &          &      & \best{+49.1}   &        &    \\\bottomrule  
\end{tabular}
\end{table}

\begin{table}[h]
\centering
\caption{\textbf{Auto-rater evaluation on physical rule classification.} We present the accuracy results for \auto{} and other video-language models on the rule classification tasks.}
\label{tab:autoeval_rule_accuracy}
\begin{tabular}{lcc}
\toprule
     & \textbf{Unseen prompts}  & \textbf{Unseen video models}  \\\hline
Random & 34.5 & 31.2 \\
VideoLlava \cite{lin2023video} & 38.1 & 38.7 \\
\flash{}    & 59.2 & 57.1 \\
\auto{}       & 78.7 & 72.9\\\hline
\textit{Rel. to Best (\%)}	& \best{+32.9} &	\best{+27.7}\\
\bottomrule
\end{tabular}
\end{table}

\section{Experiments}
\label{sec:experiments}

Here, we present the benchmarking results and the fine-grained analysis (\S \ref{sec:benchmarking}). Then, we note the usefulness of our auto-rater against modern video-language models (\S \ref{sec:autoeval}).

\subsection{Main Results}
\label{sec:benchmarking}

\paragraph{Performance on the dataset.} We compare the joint performance of various open and closed text-to-video generative models on the \name{} dataset in Table \ref{tab:main_results}. Specifically, we present their performance on the entire dataset, the hard split, and subsets focused on physical activities/sports (PA) and object interactions (OI). Even the best-performing model, Wan2.1-14B, achieves only $32.6\%$ and $21.9\%$ on the full and hard splits of our dataset, respectively. Its relatively strong performance compared to other models can be attributed to the diversity of its multimodal training data, along with robust motion filtering that preserves high-quality videos across a wide range of actions.

Furthermore, we observe that closed models, such as Ray2, perform worse than open models like Wan2.1-14B and CogVideoX-5B. This suggests that closed models are not necessarily superior to open models in capturing physical commonsense. Notably, Cosmos-Diffusion-7B achieves the second-best score on the hard split, even outperforming the much larger HunyuanVideo-13B model. This may be due to the high representation of human actions in its training data, along with synthetically rendered simulations \citep{agarwal2025cosmos}.

Additionally, we find that performance on physical activities (sports) is generally lower than on object interactions across different video models. This suggests that future data curation efforts should focus on collecting high-quality sports activity videos (e.g., tennis, discus throw, baseball, cricket) to improve performance on the \name{} dataset. Finally, we present the correlation between SA and PC judgments and other video metrics, including aesthetics (measured using the LAION classifier \cite{LAIONAIaesthetic}) and motion quality (measured using optical flow from RAFT \cite{teed2020raft}), in Table \ref{tab:correlation_sa_pc_motion_aesthetics}. Our results reveal that physical commonsense is not well-correlated with any of these video metrics. This indicates that a model cannot achieve high performance on our dataset simply by optimizing for aesthetics and motion quality; rather, it requires dedicated efforts to incorporate physical commonsense into video generation. Overall, our findings suggest that \name{} presents a significant challenge for modern video models, with substantial room for improvement in future iterations.



\paragraph{Fine-grained Analysis.}

In our human annotations, we create a list of physical rules (and associated laws) that are violated in each video of the \name{} dataset. We then analyze the fraction of instances in which a physical law is violated to gain fine-grained insights into model behavior. For example, if 100 physical rules are associated with the law of gravity and 25 of them are violated, the violation score would be $25\%$. We present the results of physical law violations in Figure \ref{fig:physics_rule_violation_analysis}. We observe that the conservation of momentum (linear or angular) and the conservation of mass are among the most frequently violated physical laws, with violation scores of $40\%$, in the videos from the \name{} dataset. Conversely, we find that reflection and buoyancy are relatively mastered with violation scores less than $20\%$. 

\paragraph{Qualitative Analysis.}

We present qualitative analysis to provide visual insights into the model's mode of failures. Specifically, we cover model-specific poor physical commonsense instances along the caption and human-judged physical violations in Appendix \ref{app:poor_physical_commonsense_model}. For example, we show that the Sora-generated video violates the physical rule `The frisbee must contant the hand before any upward movement occurs' (Appendix Figure \ref{fig:sora_full_bad_examples}). We also provide several qualitative examples across diverse physical law violations across different models in Appendix \ref{app:poor_physical_commonsense_law}. For example, we highlight that the `golf ball does not move after being struck by the golf club' for Ray2 (Figure \ref{fig:conservationmomentum_full_bad_examples}). Furthermore, we present qualitative examples in Figure \ref{fig:wan_comparisons_examples} to compare the best-performing model, Wan2.1-14B, with other video models. Notably, we observe violations of physical commonsense, such as jetskis moving unnaturally in reverse and the deformation of a solid sledgehammer, defying the principles of elasticity. However, even Wan suffers from the lack of physical commonsense, as shown in Figure \ref{fig:wan_limitations_examples}. In this case, we highlight that a rock starts rolling and accelerating uphill, defying the physical law of gravity. 


\subsection{\auto{}}
\label{sec:autoeval}

To enable scalable judgments, we supplement the dataset with an automatic evaluator \auto{}. 
We assess the auto-rater performance for two settings: (a) unseen prompts: where the auto-rater is assessed on the same video models as used in training but on the videos generated for unseen (testing) captions, (b) unseen video models: where the auto-rater is assessed on the videos generated from the unseen video models for unseen (testing) captions.

We compare the correlation performance of \auto{} against several baselines in Table \ref{tab:autoeval_sa_pc_corr}. In particular, \auto{} achieves relative gains of $47.4\%$ and $49\%$ on unseen prompts and unseen video models, respectively, compared to the best-performing baselines. Further, our auto-rater outperforms the state-of-the-art multimodal model, \flash{}, with relative gains of $81\%$ in semantic adherence and $236\%$ in physical commonsense judgments. Further, we evaluate the accuracy and F1 performance of \auto{} against VideoCon-Physics for joint score judgments in Table \ref{tab:autoeval_joint_acc}. Our results show that \auto{} maintains a strong balance between joint accuracy and F1 scores. These findings highlight the need for further improvements in video-language models to enhance their physical commonsense understanding. Finally, we assess the physical rule classification accuracy of \auto{} against baselines in Table \ref{tab:autoeval_rule_accuracy}. Our model achieves relative gains of $32.9\%$ on unseen prompts and $27.7\%$ on unseen video models compared to \flash{}. This demonstrates that our unified auto-rater can reliably handle a variety of tasks, providing future model developers with a robust tool for testing on the \name{} dataset.

\section{Conclusion}
\label{sec:conclusion}

We introduce \name{}, a benchmark for evaluating physical commonsense in videos generated by modern models. We reveal a large gap in their ability to align with prompts and generate videos that follow physical commonsense. Further, we provide physical law violations and an auto-rater for scalable evaluation. Overall, this dataset advances our understanding of the current state of the video generative models as general-purpose world simulators.

\section{Acknowledgements}

Hritik Bansal is supported in part by AFOSR MURI grant FA9550-22-1-0380. We also thank Ashima Suvarna and Zongyu Lin for their helpful feedback on the draft.

\bibliographystyle{plain}
\bibliography{main}

\begin{thebibliography}{10}

\bibitem{agarwal2025cosmos}
Niket Agarwal, Arslan Ali, Maciej Bala, Yogesh Balaji, Erik Barker, Tiffany Cai, Prithvijit Chattopadhyay, Yongxin Chen, Yin Cui, Yifan Ding, et~al.
\newblock Cosmos world foundation model platform for physical ai.
\newblock {\em arXiv preprint arXiv:2501.03575}, 2025.

\bibitem{bakhtin2019phyrenewbenchmarkphysical}
Anton Bakhtin, Laurens van~der Maaten, Justin Johnson, Laura Gustafson, and Ross Girshick.
\newblock Phyre: A new benchmark for physical reasoning, 2019.

\bibitem{bansal2024videocon}
Hritik Bansal, Yonatan Bitton, Idan Szpektor, Kai-Wei Chang, and Aditya Grover.
\newblock Videocon: Robust video-language alignment via contrast captions.
\newblock In {\em Proceedings of the IEEE/CVF Conference on Computer Vision and Pattern Recognition}, pages 13927--13937, 2024.

\bibitem{bansal2024talc}
Hritik Bansal, Yonatan Bitton, Michal Yarom, Idan Szpektor, Aditya Grover, and Kai-Wei Chang.
\newblock Talc: Time-aligned captions for multi-scene text-to-video generation.
\newblock {\em arXiv preprint arXiv:2405.04682}, 2024.

\bibitem{bansal2024videophy}
Hritik Bansal, Zongyu Lin, Tianyi Xie, Zeshun Zong, Michal Yarom, Yonatan Bitton, Chenfanfu Jiang, Yizhou Sun, Kai-Wei Chang, and Aditya Grover.
\newblock Videophy: Evaluating physical commonsense for video generation.
\newblock {\em arXiv preprint arXiv:2406.03520}, 2024.

\bibitem{battaglia2013simulation}
Peter~W Battaglia, Jessica~B Hamrick, and Joshua~B Tenenbaum.
\newblock Simulation as an engine of physical scene understanding.
\newblock {\em Proceedings of the National Academy of Sciences}, 110(45):18327--18332, 2013.

\bibitem{bisk2019piqareasoningphysicalcommonsense}
Yonatan Bisk, Rowan Zellers, Ronan~Le Bras, Jianfeng Gao, and Yejin Choi.
\newblock Piqa: Reasoning about physical commonsense in natural language, 2019.

\bibitem{svd}
Andreas Blattmann, Tim Dockhorn, Sumith Kulal, Daniel Mendelevitch, Maciej Kilian, Dominik Lorenz, Yam Levi, Zion English, Vikram Voleti, Adam Letts, et~al.
\newblock Stable video diffusion: Scaling latent video diffusion models to large datasets.
\newblock {\em arXiv preprint arXiv:2311.15127}, 2023.

\bibitem{blattmann2023alignlatentshighresolutionvideo}
Andreas Blattmann, Robin Rombach, Huan Ling, Tim Dockhorn, Seung~Wook Kim, Sanja Fidler, and Karsten Kreis.
\newblock Align your latents: High-resolution video synthesis with latent diffusion models, 2023.

\bibitem{videoworldsimulators2024}
Tim Brooks, Bill Peebles, Connor Holmes, Will DePue, Yufei Guo, Li~Jing, David Schnurr, Joe Taylor, Troy Luhman, Eric Luhman, Clarence Ng, Ricky Wang, and Aditya Ramesh.
\newblock Video generation models as world simulators.
\newblock 2024.

\bibitem{bruce2024geniegenerativeinteractiveenvironments}
Jake Bruce, Michael Dennis, Ashley Edwards, Jack Parker-Holder, Yuge Shi, Edward Hughes, Matthew Lai, Aditi Mavalankar, Richie Steigerwald, Chris Apps, Yusuf Aytar, Sarah Bechtle, Feryal Behbahani, Stephanie Chan, Nicolas Heess, Lucy Gonzalez, Simon Osindero, Sherjil Ozair, Scott Reed, Jingwei Zhang, Konrad Zolna, Jeff Clune, Nando de~Freitas, Satinder Singh, and Tim Rocktäschel.
\newblock Genie: Generative interactive environments, 2024.

\bibitem{bruce2024genie}
Jake Bruce, Michael~D Dennis, Ashley Edwards, Jack Parker-Holder, Yuge Shi, Edward Hughes, Matthew Lai, Aditi Mavalankar, Richie Steigerwald, Chris Apps, et~al.
\newblock Genie: Generative interactive environments.
\newblock In {\em Forty-first International Conference on Machine Learning}, 2024.

\bibitem{carreira2018short}
Joao Carreira, Eric Noland, Andras Banki-Horvath, Chloe Hillier, and Andrew Zisserman.
\newblock A short note about kinetics-600.
\newblock {\em arXiv preprint arXiv:1808.01340}, 2018.

\bibitem{chen2024videocrafter2}
Haoxin Chen, Yong Zhang, Xiaodong Cun, Menghan Xia, Xintao Wang, Chao Weng, and Ying Shan.
\newblock Videocrafter2: Overcoming data limitations for high-quality video diffusion models, 2024.

\bibitem{GeminiFlash2p0}
Google DeepMind.
\newblock {I}ntroducing {G}emini 2.0: our new {A}{I} model for the agentic era --- blog.google.
\newblock \url{https://blog.google/technology/google-deepmind/google-gemini-ai-update-december-2024/}, 2025.

\bibitem{du2024learning}
Yilun Du, Sherry Yang, Bo~Dai, Hanjun Dai, Ofir Nachum, Josh Tenenbaum, Dale Schuurmans, and Pieter Abbeel.
\newblock Learning universal policies via text-guided video generation.
\newblock {\em Advances in Neural Information Processing Systems}, 36, 2024.

\bibitem{duan2022survey}
Jiafei Duan, Arijit Dasgupta, Jason Fischer, and Cheston Tan.
\newblock A survey on machine learning approaches for modelling intuitive physics.
\newblock {\em arXiv preprint arXiv:2202.06481}, 2022.

\bibitem{deepmindGenie2}
GoogleDeepMind Genie2.
\newblock {G}enie 2: {A} large-scale foundation world model --- deepmind.google.
\newblock \url{https://deepmind.google/discover/blog/genie-2-a-large-scale-foundation-world-model/}, 2024.

\bibitem{goyal2017something}
Raghav Goyal, Samira Ebrahimi~Kahou, Vincent Michalski, Joanna Materzynska, Susanne Westphal, Heuna Kim, Valentin Haenel, Ingo Fruend, Peter Yianilos, Moritz Mueller-Freitag, et~al.
\newblock The" something something" video database for learning and evaluating visual common sense.
\newblock In {\em Proceedings of the IEEE international conference on computer vision}, pages 5842--5850, 2017.

\bibitem{he2024videoscore}
Xuan He, Dongfu Jiang, Ge~Zhang, Max Ku, Achint Soni, Sherman Siu, Haonan Chen, Abhranil Chandra, Ziyan Jiang, Aaran Arulraj, et~al.
\newblock Videoscore: Building automatic metrics to simulate fine-grained human feedback for video generation.
\newblock {\em arXiv preprint arXiv:2406.15252}, 2024.

\bibitem{ho2022imagenvideohighdefinition}
Jonathan Ho, William Chan, Chitwan Saharia, Jay Whang, Ruiqi Gao, Alexey Gritsenko, Diederik~P. Kingma, Ben Poole, Mohammad Norouzi, David~J. Fleet, and Tim Salimans.
\newblock Imagen video: High definition video generation with diffusion models, 2022.

\bibitem{ho2022videodiffusionmodels}
Jonathan Ho, Tim Salimans, Alexey Gritsenko, William Chan, Mohammad Norouzi, and David~J. Fleet.
\newblock Video diffusion models, 2022.

\bibitem{hong2022cogvideo}
Wenyi Hong, Ming Ding, Wendi Zheng, Xinghan Liu, and Jie Tang.
\newblock Cogvideo: Large-scale pretraining for text-to-video generation via transformers.
\newblock {\em arXiv preprint arXiv:2205.15868}, 2022.

\bibitem{hu2022lora}
Edward~J Hu, Yelong Shen, Phillip Wallis, Zeyuan Allen-Zhu, Yuanzhi Li, Shean Wang, Lu~Wang, Weizhu Chen, et~al.
\newblock Lora: Low-rank adaptation of large language models.
\newblock {\em ICLR}, 1(2):3, 2022.

\bibitem{huang2024vbench}
Ziqi Huang, Yinan He, Jiashuo Yu, Fan Zhang, Chenyang Si, Yuming Jiang, Yuanhan Zhang, Tianxing Wu, Qingyang Jin, Nattapol Chanpaisit, et~al.
\newblock Vbench: Comprehensive benchmark suite for video generative models.
\newblock In {\em Proceedings of the IEEE/CVF Conference on Computer Vision and Pattern Recognition}, pages 21807--21818, 2024.

\bibitem{khachatryan2023text2videozerotexttoimagediffusionmodels}
Levon Khachatryan, Andranik Movsisyan, Vahram Tadevosyan, Roberto Henschel, Zhangyang Wang, Shant Navasardyan, and Humphrey Shi.
\newblock Text2video-zero: Text-to-image diffusion models are zero-shot video generators, 2023.

\bibitem{kingma2014adam}
Diederik~P Kingma and Jimmy Ba.
\newblock Adam: A method for stochastic optimization.
\newblock {\em arXiv preprint arXiv:1412.6980}, 2014.

\bibitem{klar2016drucker}
Gergely Kl{\'a}r, Theodore Gast, Andre Pradhana, Chuyuan Fu, Craig Schroeder, Chenfanfu Jiang, and Joseph Teran.
\newblock Drucker-prager elastoplasticity for sand animation.
\newblock {\em ACM Transactions on Graphics (TOG)}, 35(4):1--12, 2016.

\bibitem{klingai}
KlingAI.
\newblock {K}{L}{I}{N}{G} {A}{I} --- klingai.com.
\newblock \url{https://www.klingai.com/}, 2024.

\bibitem{kondratyuk2024videopoetlargelanguagemodel}
Dan Kondratyuk, Lijun Yu, Xiuye Gu, José Lezama, Jonathan Huang, Grant Schindler, Rachel Hornung, Vighnesh Birodkar, Jimmy Yan, Ming-Chang Chiu, Krishna Somandepalli, Hassan Akbari, Yair Alon, Yong Cheng, Josh Dillon, Agrim Gupta, Meera Hahn, Anja Hauth, David Hendon, Alonso Martinez, David Minnen, Mikhail Sirotenko, Kihyuk Sohn, Xuan Yang, Hartwig Adam, Ming-Hsuan Yang, Irfan Essa, Huisheng Wang, David~A. Ross, Bryan Seybold, and Lu~Jiang.
\newblock Videopoet: A large language model for zero-shot video generation, 2024.

\bibitem{kong2024hunyuanvideo}
Weijie Kong, Qi~Tian, Zijian Zhang, Rox Min, Zuozhuo Dai, Jin Zhou, Jiangfeng Xiong, Xin Li, Bo~Wu, Jianwei Zhang, et~al.
\newblock Hunyuanvideo: A systematic framework for large video generative models.
\newblock {\em arXiv preprint arXiv:2412.03603}, 2024.

\bibitem{LAIONAIaesthetic}
Laion.
\newblock {G}it{H}ub - {L}{A}{I}{O}{N}-{A}{I}/aesthetic-predictor: {A} linear estimator on top of clip to predict the aesthetic quality of pictures --- github.com.
\newblock \url{https://github.com/LAION-AI/aesthetic-predictor}.

\bibitem{liang2024dreamitate}
Junbang Liang, Ruoshi Liu, Ege Ozguroglu, Sruthi Sudhakar, Achal Dave, Pavel Tokmakov, Shuran Song, and Carl Vondrick.
\newblock Dreamitate: Real-world visuomotor policy learning via video generation.
\newblock {\em arXiv preprint arXiv:2406.16862}, 2024.

\bibitem{lin2023video}
Bin Lin, Yang Ye, Bin Zhu, Jiaxi Cui, Munan Ning, Peng Jin, and Li~Yuan.
\newblock Video-llava: Learning united visual representation by alignment before projection.
\newblock {\em arXiv preprint arXiv:2311.10122}, 2023.

\bibitem{lin2024evaluatingtexttovisualgenerationimagetotext}
Zhiqiu Lin, Deepak Pathak, Baiqi Li, Jiayao Li, Xide Xia, Graham Neubig, Pengchuan Zhang, and Deva Ramanan.
\newblock Evaluating text-to-visual generation with image-to-text generation, 2024.

\bibitem{liu2024physgen}
Shaowei Liu, Zhongzheng Ren, Saurabh Gupta, and Shenlong Wang.
\newblock Physgen: Rigid-body physics-grounded image-to-video generation.
\newblock In {\em European Conference on Computer Vision}, pages 360--378. Springer, 2024.

\bibitem{liu2024evalcrafter}
Yaofang Liu, Xiaodong Cun, Xuebo Liu, Xintao Wang, Yong Zhang, Haoxin Chen, Yang Liu, Tieyong Zeng, Raymond Chan, and Ying Shan.
\newblock Evalcrafter: Benchmarking and evaluating large video generation models.
\newblock In {\em Proceedings of the IEEE/CVF Conference on Computer Vision and Pattern Recognition}, pages 22139--22149, 2024.

\bibitem{liu2024sora}
Yixin Liu, Kai Zhang, Yuan Li, Zhiling Yan, Chujie Gao, Ruoxi Chen, Zhengqing Yuan, Yue Huang, Hanchi Sun, Jianfeng Gao, et~al.
\newblock Sora: A review on background, technology, limitations, and opportunities of large vision models.
\newblock {\em arXiv preprint arXiv:2402.17177}, 2024.

\bibitem{ChatbotArena}
lmarena.
\newblock {C}hatbot {A}rena {L}eaderboard - a {H}ugging {F}ace {S}pace by lmarena-ai --- huggingface.co.
\newblock \url{https://huggingface.co/spaces/lmarena-ai/chatbot-arena-leaderboard}.

\bibitem{lumalabsLumaRay2}
LumaAI.
\newblock {L}uma {R}ay2 --- lumalabs.ai.
\newblock \url{https://lumalabs.ai/ray}, 2025.

\bibitem{luo2025openmagvit2opensourceprojectdemocratizing}
Zhuoyan Luo, Fengyuan Shi, Yixiao Ge, Yujiu Yang, Limin Wang, and Ying Shan.
\newblock Open-magvit2: An open-source project toward democratizing auto-regressive visual generation, 2025.

\bibitem{mccloskey1983intuitive}
Michael McCloskey, Allyson Washburn, and Linda Felch.
\newblock Intuitive physics: the straight-down belief and its origin.
\newblock {\em Journal of Experimental Psychology: Learning, Memory, and Cognition}, 9(4):636, 1983.

\bibitem{meng2024towards}
Fanqing Meng, Jiaqi Liao, Xinyu Tan, Wenqi Shao, Quanfeng Lu, Kaipeng Zhang, Yu~Cheng, Dianqi Li, Yu~Qiao, and Ping Luo.
\newblock Towards world simulator: Crafting physical commonsense-based benchmark for video generation.
\newblock {\em arXiv preprint arXiv:2410.05363}, 2024.

\bibitem{motamed2025generative}
Saman Motamed, Laura Culp, Kevin Swersky, Priyank Jaini, and Robert Geirhos.
\newblock Do generative video models learn physical principles from watching videos?
\newblock {\em arXiv preprint arXiv:2501.09038}, 2025.

\bibitem{o2002graphical}
James~F O'brien, Adam~W Bargteil, and Jessica~K Hodgins.
\newblock Graphical modeling and animation of ductile fracture.
\newblock In {\em Proceedings of the 29th annual conference on Computer graphics and interactive techniques}, pages 291--294, 2002.

\bibitem{phan2025humanity}
Long Phan, Alice Gatti, Ziwen Han, Nathaniel Li, Josephina Hu, Hugh Zhang, Sean Shi, Michael Choi, Anish Agrawal, Arnav Chopra, et~al.
\newblock Humanity's last exam.
\newblock {\em arXiv preprint arXiv:2501.14249}, 2025.

\bibitem{qin2024worldsimbench}
Yiran Qin, Zhelun Shi, Jiwen Yu, Xijun Wang, Enshen Zhou, Lijun Li, Zhenfei Yin, Xihui Liu, Lu~Sheng, Jing Shao, et~al.
\newblock Worldsimbench: Towards video generation models as world simulators.
\newblock {\em arXiv preprint arXiv:2410.18072}, 2024.

\bibitem{radford2021learning}
Alec Radford, Jong~Wook Kim, Chris Hallacy, Aditya Ramesh, Gabriel Goh, Sandhini Agarwal, Girish Sastry, Amanda Askell, Pamela Mishkin, Jack Clark, et~al.
\newblock Learning transferable visual models from natural language supervision.
\newblock In {\em International conference on machine learning}, pages 8748--8763. PMLR, 2021.

\bibitem{roberts2025zerobench}
Jonathan Roberts, Mohammad~Reza Taesiri, Ansh Sharma, Akash Gupta, Samuel Roberts, Ioana Croitoru, Simion-Vlad Bogolin, Jialu Tang, Florian Langer, Vyas Raina, et~al.
\newblock Zerobench: An impossible visual benchmark for contemporary large multimodal models.
\newblock {\em arXiv preprint arXiv:2502.09696}, 2025.

\bibitem{soomro2012ucf101}
Khurram Soomro, Amir~Roshan Zamir, and Mubarak Shah.
\newblock Ucf101: A dataset of 101 human actions classes from videos in the wild.
\newblock {\em arXiv preprint arXiv:1212.0402}, 2012.

\bibitem{teed2020raft}
Zachary Teed and Jia Deng.
\newblock Raft: Recurrent all-pairs field transforms for optical flow.
\newblock In {\em Computer Vision--ECCV 2020: 16th European Conference, Glasgow, UK, August 23--28, 2020, Proceedings, Part II 16}, pages 402--419. Springer, 2020.

\bibitem{unterthiner2019accurategenerativemodelsvideo}
Thomas Unterthiner, Sjoerd van Steenkiste, Karol Kurach, Raphael Marinier, Marcin Michalski, and Sylvain Gelly.
\newblock Towards accurate generative models of video: A new metric and challenges, 2019.

\bibitem{deepmindveo2}
GoogleDeepMind Veo2.
\newblock {V}eo 2 --- deepmind.google.
\newblock \url{https://deepmind.google/technologies/veo/veo-2/}, 2024.

\bibitem{villegas2022phenakivariablelengthvideo}
Ruben Villegas, Mohammad Babaeizadeh, Pieter-Jan Kindermans, Hernan Moraldo, Han Zhang, Mohammad~Taghi Saffar, Santiago Castro, Julius Kunze, and Dumitru Erhan.
\newblock Phenaki: Variable length video generation from open domain textual description, 2022.

\bibitem{wan14b}
Wan.
\newblock {W}an-{A}{I}/{W}an2.1-{T}2{V}-14{B} · {H}ugging {F}ace --- huggingface.co.
\newblock \url{https://huggingface.co/Wan-AI/Wan2.1-T2V-14B}, 2025.

\bibitem{wang2023modelscopetexttovideotechnicalreport}
Jiuniu Wang, Hangjie Yuan, Dayou Chen, Yingya Zhang, Xiang Wang, and Shiwei Zhang.
\newblock Modelscope text-to-video technical report, 2023.

\bibitem{wang2023laviehighqualityvideogeneration}
Yaohui Wang, Xinyuan Chen, Xin Ma, Shangchen Zhou, Ziqi Huang, Yi~Wang, Ceyuan Yang, Yinan He, Jiashuo Yu, Peiqing Yang, Yuwei Guo, Tianxing Wu, Chenyang Si, Yuming Jiang, Cunjian Chen, Chen~Change Loy, Bo~Dai, Dahua Lin, Yu~Qiao, and Ziwei Liu.
\newblock Lavie: High-quality video generation with cascaded latent diffusion models, 2023.

\bibitem{yang2024cogvideox}
Zhuoyi Yang, Jiayan Teng, Wendi Zheng, Ming Ding, Shiyu Huang, Jiazheng Xu, Yuanming Yang, Wenyi Hong, Xiaohan Zhang, Guanyu Feng, et~al.
\newblock Cogvideox: Text-to-video diffusion models with an expert transformer.
\newblock {\em arXiv preprint arXiv:2408.06072}, 2024.

\bibitem{yarom2023you}
Michal Yarom, Yonatan Bitton, Soravit Changpinyo, Roee Aharoni, Jonathan Herzig, Oran Lang, Eran Ofek, and Idan Szpektor.
\newblock What you see is what you read? improving text-image alignment evaluation.
\newblock {\em Advances in Neural Information Processing Systems}, 36:1601--1619, 2023.

\end{thebibliography}

\clearpage
\appendix

\begin{table*}[t]
\centering
\caption{\small{\textbf{Comparison between \name{} and several prior work.} We highlight the salient features of the \name{} and show that its unique contributions. For instance, it is one of the largest datasets for physical commonsense evaluation, along with violated physical rule (and law) annotations. Further, we will release all the data, and videos openly for public-use.}}
\label{tab:related_work_comparison}
\resizebox{\textwidth}{!}{%
\begin{tabular}{lcccccc}
\toprule
\textbf{Feature} & \textbf{VBench}\cite{huang2024vbench}  & \textbf{PhyGenBench} \cite{meng2024towards} & \textbf{PhysicsIQ} \cite{motamed2025generative} & \textbf{EvalCrafter} \cite{liu2024evalcrafter}  & \textbf{\videophy{}} \cite{bansal2024videophy} & \textbf{\name{}} (Ours)    \\\hline
Num of captions                        & 1746    & 160                                      & 396                                    & 700          & 688                                    & 3940         \\
Gold human evaluation                          & \cmark      & \cmark                                      & \xmark                                     & \cmark          & \cmark                                    & \cmark          \\

Physical commonsense eval.                      & \xmark      & \cmark                                      & \cmark                                    & \xmark           & \cmark                                    & \cmark          \\
Physical rules and laws annotations            & \xmark      & \cmark                                      & \xmark                                     & \xmark           & \xmark                                     & \cmark          \\
Real-world action-centric   & \xmark      & \xmark                                       & \xmark                                     & \xmark           & \xmark                                     & \cmark          \\
Long (dense) captions                            & \xmark     & \cmark                                      & \cmark                                    & \xmark           & \xmark                                     & \cmark          \\
Hard subset                       & \xmark      & \xmark                                       & \xmark                                     & \xmark           & \xmark                                     & \cmark          \\
Automatic evaluator                            & \cmark     & \cmark                                      & \cmark                                    & \cmark          & \cmark                                    & \cmark          \\
Release videos and annotations  & \cmark      & \xmark                                       & \xmark                                     & \cmark           & \cmark                                    & \cmark          \\
Human feedback type                            & Pairwise & Rating                                   & -                                     & Rating (1-5) & Rating (0-1)                           & Rating (1-5)\\
\bottomrule
\end{tabular}%
}
\end{table*}

\section{Related Work}
\label{sec:related_work}




The rapid advancement of video generation models necessitates robust benchmarks and evaluation methodologies, particularly for assessing \textit{physical commonsense} understanding. This is a crucial step towards realizing the vision of video generation models as `world simulators' \cite{videoworldsimulators2024}. Our work, \name{}, extends prior work by addressing key limitations in existing benchmarks and evaluation methods. We structure our discussion of related work around the following key areas:

\subsection{Video Generation Models}
Recent progress in text-to-video (T2V) generation has been driven primarily by two architectural paradigms: diffusion models \cite{ho2022imagenvideohighdefinition, ho2022videodiffusionmodels, blattmann2023alignlatentshighresolutionvideo, wang2023laviehighqualityvideogeneration, wang2023modelscopetexttovideotechnicalreport, khachatryan2023text2videozerotexttoimagediffusionmodels} and autoregressive models \cite{luo2025openmagvit2opensourceprojectdemocratizing, kondratyuk2024videopoetlargelanguagemodel, hong2022cogvideo, villegas2022phenakivariablelengthvideo}. Diffusion models, such as Stable Video Diffusion (SVD) \cite{svd} and Sora \cite{liu2024sora}, involve a multi-stage training process and operate in a latent space. Autoregressive models, including VideoPoet \cite{kondratyuk2024videopoetlargelanguagemodel} and CogVideo \cite{hong2022cogvideo}, predict future frames based on past frames. While these models demonstrate impressive visual fidelity, their ability to capture underlying physical principles remains an open question. This highlights the need for rigorous benchmarks like \name{}. Additionally, recent works, such as Genie \cite{bruce2024geniegenerativeinteractiveenvironments}, explore interactive video generation, further expanding the potential applications of these models beyond static content creation.

\subsection{General Video Generation Benchmarks}
Several benchmarks focus on evaluating general aspects of video quality but do not specifically assess physical reasoning. \textbf{VBench} \cite{huang2024vbench} introduces a hierarchical evaluation approach, covering motion smoothness, background consistency, and overall visual fidelity. \textbf{EvalCrafter} \cite{liu2024evalcrafter} proposes 17 objective metrics focused on different aspects of video quality. While these benchmarks contribute to understanding model performance, they do not isolate or systematically evaluate physical commonsense in generated videos.

\subsection{Benchmarks for Physical Understanding in Video Generation}
Several works have introduced benchmarks to assess the physical plausibility of generated videos. \textbf{\videophy{}} \cite{bansal2024videophy} was an early effort in this direction, evaluating semantic adherence and physical plausibility across 688 prompts. \name{} extends this work by introducing a more extensive dataset of 3940 prompts, shifting the focus from material interactions to \textit{real-world actions}, and providing a \textit{5-point Likert scale} for human evaluation. Furthermore, \name{} explicitly annotates physical rules in each video, allowing for a more fine-grained understanding of model behavior.

\textbf{Physics-IQ} \cite{motamed2025generative} conditions models on initial frames of real videos and measures the similarity between predicted and ground-truth video continuations. While this approach offers valuable insights, it is primarily designed for video prediction rather than open-ended text-to-video generation. Additionally, extending it to longer or more complex multi-event scenarios presents challenges.

\textbf{PhyGenBench} \cite{meng2024towards} proposes a benchmark of 160 prompts and an automated evaluation framework, PhyGenEval. This work introduces structured evaluations of physical reasoning; however, its relatively small scale makes generalization difficult. Additionally, it adopts a strict one-to-one mapping between prompts and physical laws, whereas real-world physics often involves multiple interacting principles. Its evaluation relies on sequential vision-language model queries, which can introduce inconsistencies and increase computational complexity. \name{} builds on these efforts by expanding dataset size, allowing for prompts that reflect multiple physical laws, and providing a more interpretable evaluation framework.

\textbf{WorldSimBench} \cite{qin2024worldsimbench} assesses video models' ability to act as \textit{world simulators} by aligning their outputs with numerical solvers. While valuable, this approach faces challenges in bridging the \textit{sim-to-real gap}, as physical simulations may not fully capture the complexity and variability of real-world interactions. \name{} addresses this by exclusively using real-world videos to ensure that evaluations remain closely aligned with practical physical dynamics.

\subsection{Automatic Evaluation Methods}
Traditional video quality metrics, such as Fréchet Video Distance (FVD) \cite{unterthiner2019accurategenerativemodelsvideo}, were not designed to assess physical plausibility. More recent approaches incorporate vision-language models (VLMs) for automatic evaluation \cite{he2024videoscore, bansal2024videophy, lin2024evaluatingtexttovisualgenerationimagetotext}. \videophy{} introduced a fine-tuned VLM-based evaluator, but existing VLMs still face challenges in reliably assessing physical commonsense. \name{} proposes \name{}-AutoEval, an enhanced automatic evaluator trained on a larger and more diverse dataset with explicit physical rule annotations. This leads to improved alignment with human assessments while maintaining interpretability.

\subsection{Physical Reasoning in AI}
The study of physical commonsense in AI builds upon insights from both artificial intelligence and cognitive science. Research on intuitive physics has explored how humans reason about object interactions and causal physical events \cite{mccloskey1983intuitive,battaglia2013simulation,duan2022survey}. In AI, approaches such as PIQA \cite{bisk2019piqareasoningphysicalcommonsense} assess physical reasoning in language, while others focus on synthetic environments for learning physical interactions \cite{bakhtin2019phyrenewbenchmarkphysical, du2024learning, bruce2024genie, agarwal2025cosmos}. \name{} contributes to this area by providing a large-scale, real-world benchmark that bridges the gap between theoretical physical reasoning and practical video generation.

In summary, \name{} advances the evaluation of physical commonsense in video generation by introducing a large-scale, action-centric benchmark, a fine-grained evaluation framework with explicit rule annotations, and a robust, human-aligned automatic evaluator. These contributions address key limitations of prior work and facilitate a more comprehensive assessment of video models' ability to simulate the physical world.

\section{List of Actions}
\label{app:actions}

Our benchmark includes a broad set of actions spanning simple motions (e.g., \textit{walking through snow, dribbling basketball}), object interactions (e.g., \textit{peeling fruit, using a paint roller}), and complex physical activities (e.g., \textit{pole vault, tightrope walking}). These actions ensure models are tested across diverse movement dynamics, real-world interactions, and varying levels of physical complexity.

Table~\ref{app_table:all_actions_list} presents the full set of 197 actions.

\section{List of Hard Actions}
\label{app:hard_actions}

Certain actions pose significant challenges for generative models due to rapid motion (e.g., \textit{throwing discus, gymnastics tumbling}), intricate interactions (e.g., \textit{popping balloons, pouring until overflow}), and structural deformations (e.g., \textit{ripping paper, bending until breaking}).Actions were deemed hard using CogVideoX-5B as a baseline.

Table~\ref{app_table:hard_actions_list} presents a subset of 60 hard actions, ensuring our benchmark tests model capabilities beyond simple motions, including stability, object manipulation, and real-world physics.

\section{Diversity of prompts for diverse actions}
\label{app:diversity_actions}

In this work, our objective was to curate a set of actions from which we could generate a wide range of diverse prompts that, while focusing on motion, still encompass various real-world settings. By ensuring diversity in prompt selection within each action, our dataset becomes more robust and capable of testing generative models across a broad spectrum of physical interactions.

\begin{figure*}[t]
     \centering
     \begin{subfigure}[h]{0.3\textwidth}
         \centering
         \includegraphics[width=\textwidth]{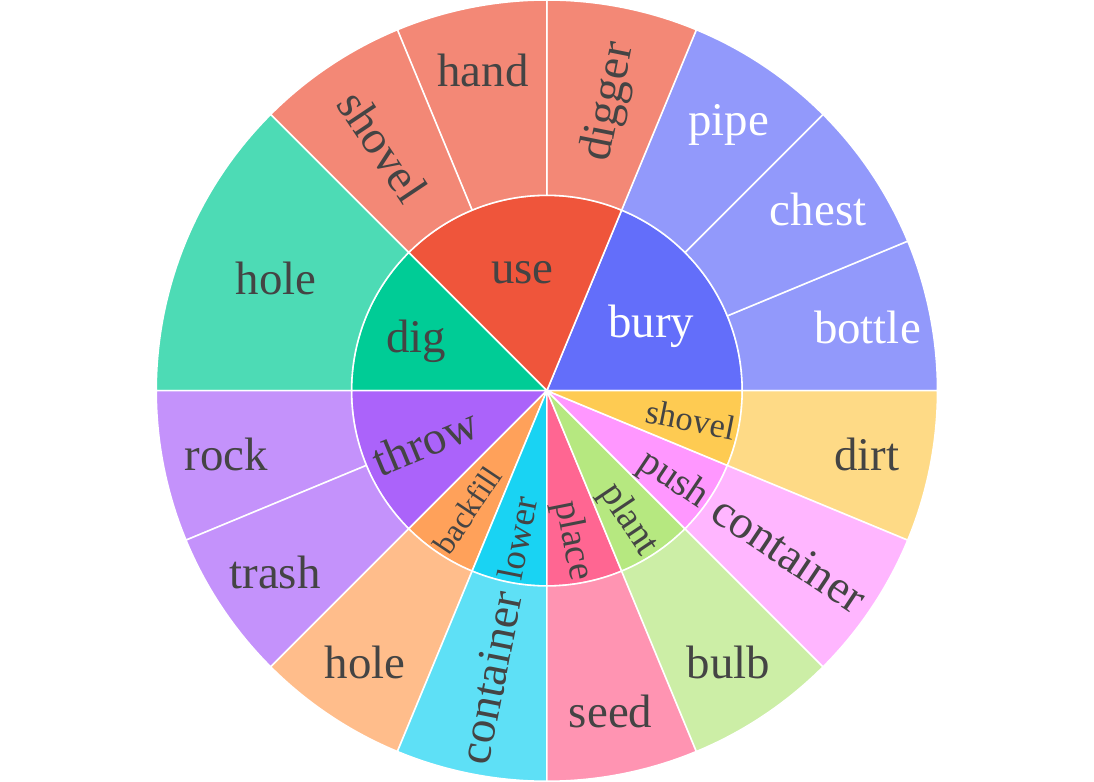}
         \caption{Burying something in something.}
         \label{fig:vn_1}
     \end{subfigure}
     \hfill
     \begin{subfigure}[h]{0.3\textwidth}
         \centering
         \includegraphics[width=\textwidth]{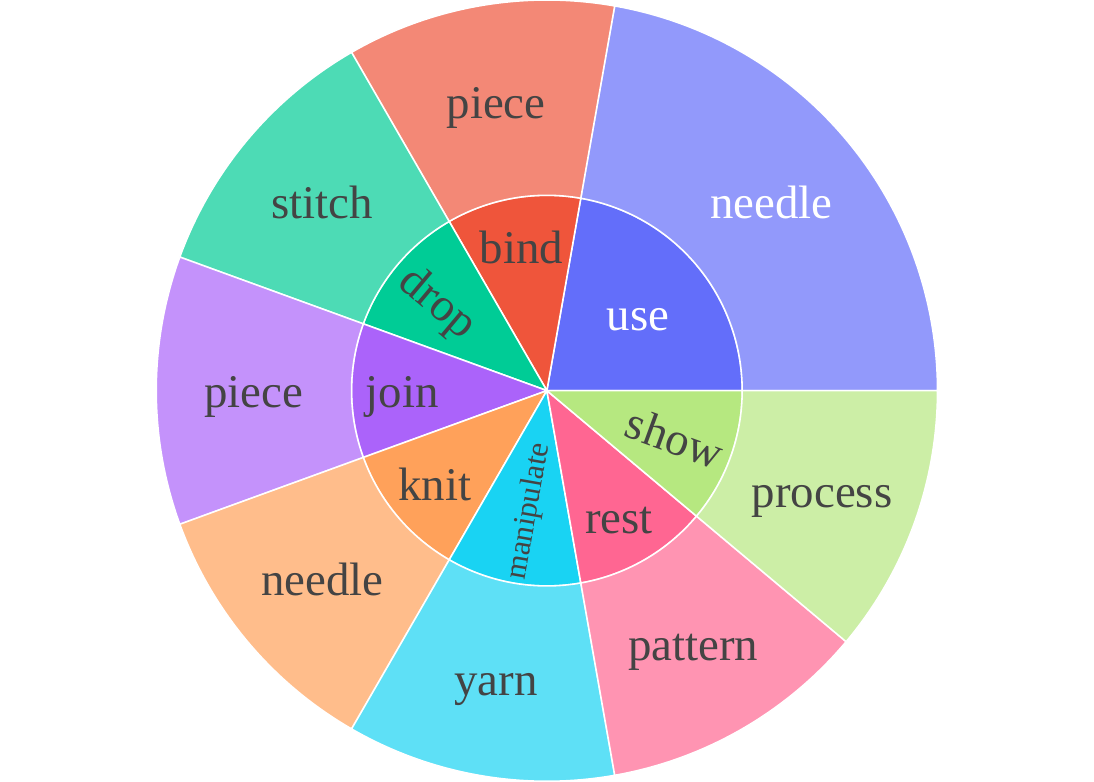}
         \caption{Knitting.}
         \label{fig:vn_2}
     \end{subfigure}
     \hfill
     \begin{subfigure}[h]{0.3\textwidth}
         \centering
         \includegraphics[width=\textwidth]{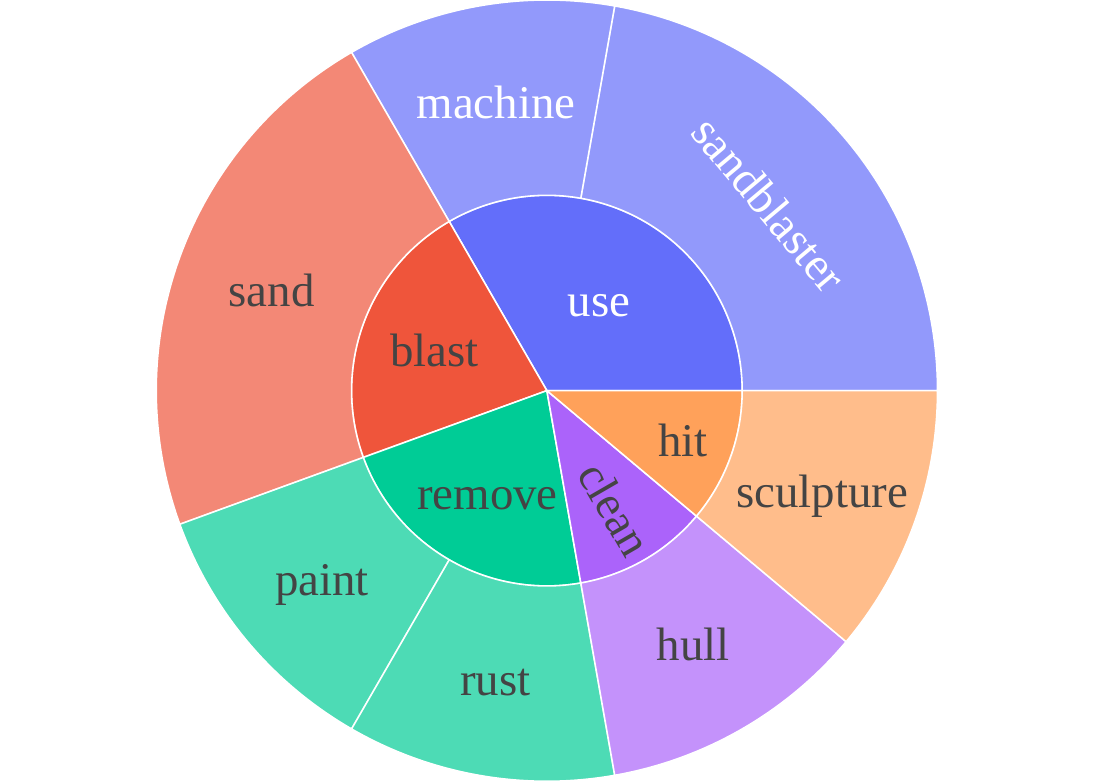}
         \caption{Blasting Sand.}
         \label{fig:vn_3}
     \end{subfigure}
        \qquad
     \begin{subfigure}[h]{0.3\textwidth}
         \centering
         \includegraphics[width=\textwidth]{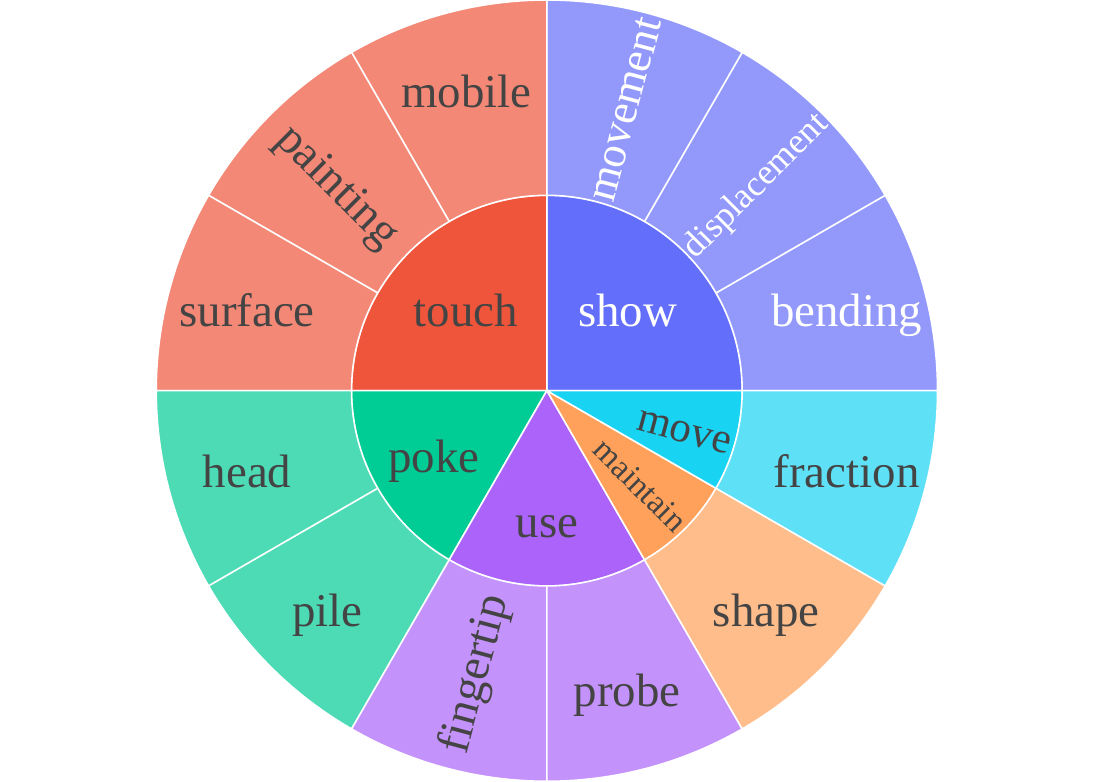}
\caption{Poking something.}
         \label{fig:vn_4}
     \end{subfigure}
     \hfill
     \begin{subfigure}[h]{0.3\textwidth}
         \centering
         \includegraphics[width=\textwidth]{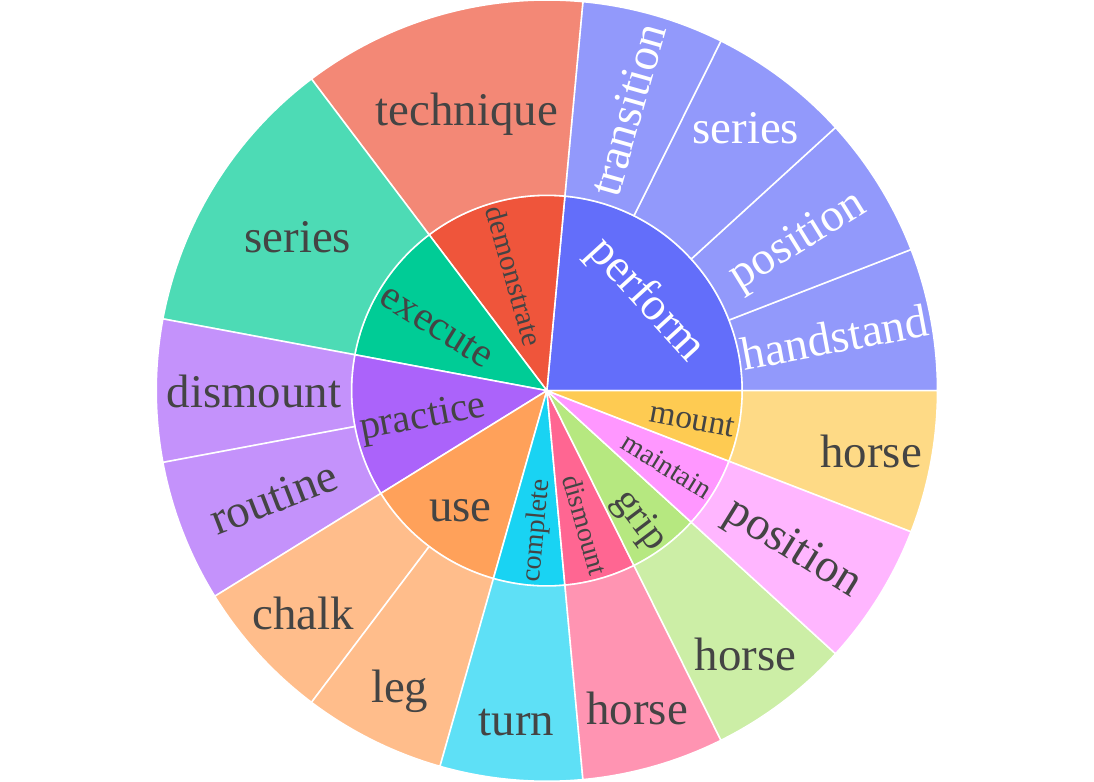}
         \caption{Pommel Horse.}
         \label{fig:vn_5}
     \end{subfigure}
     \hfill
     \begin{subfigure}[h]{0.3\textwidth}
         \centering
         \includegraphics[width=\textwidth]{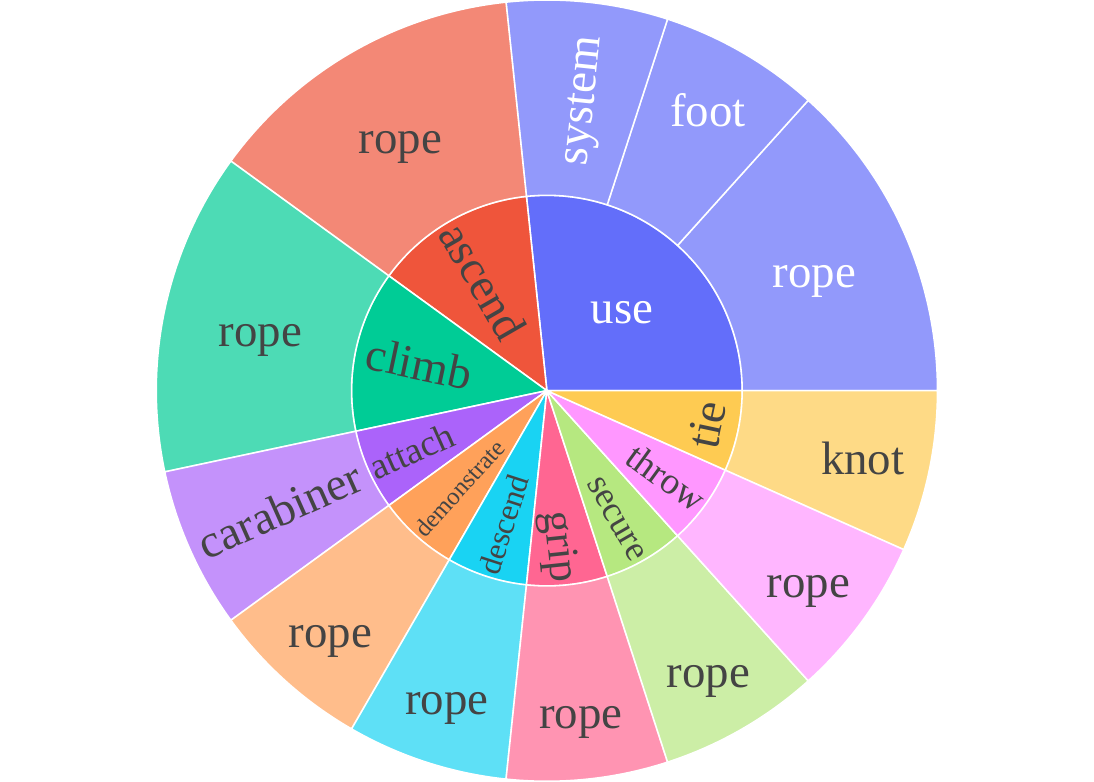}
         \caption{Rope Climbing.}
         \label{fig:vn_6}
     \end{subfigure}
     \qquad
 \begin{subfigure}[h]{0.3\textwidth}
         \centering
         \includegraphics[width=\textwidth]{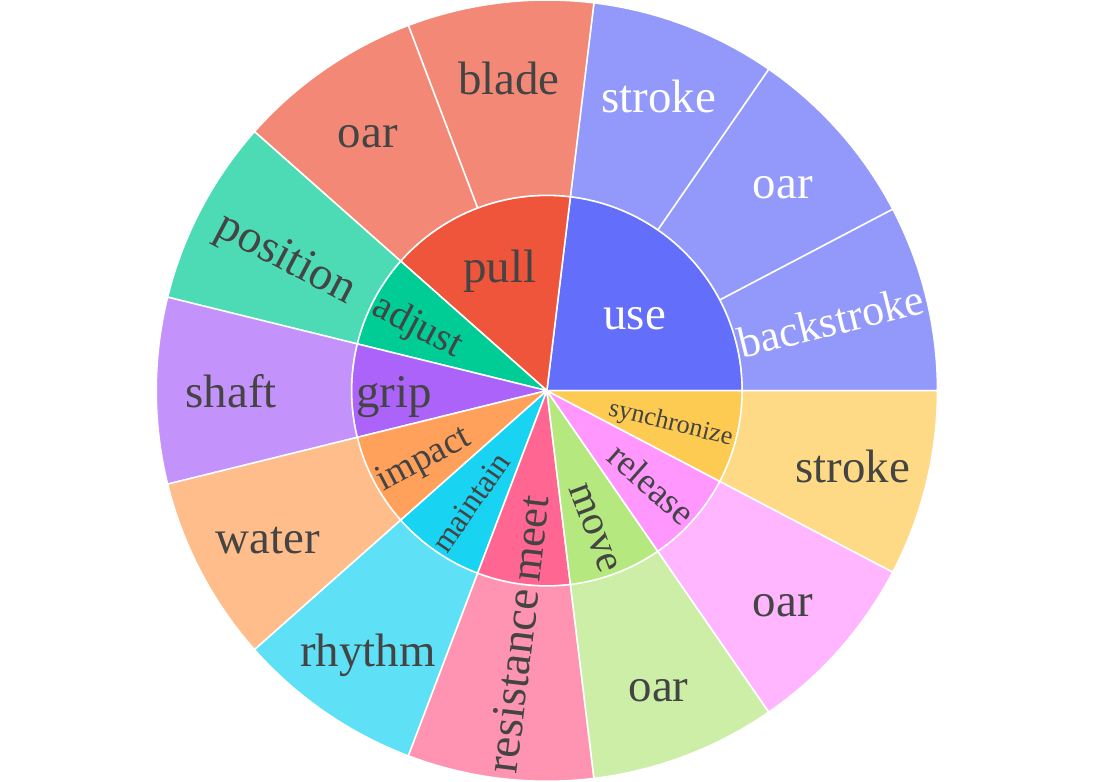}
\caption{Rowing.}
         \label{fig:vn_7}
     \end{subfigure}
     \hfill
     \begin{subfigure}[h]{0.3\textwidth}
         \centering
         \includegraphics[width=\textwidth]{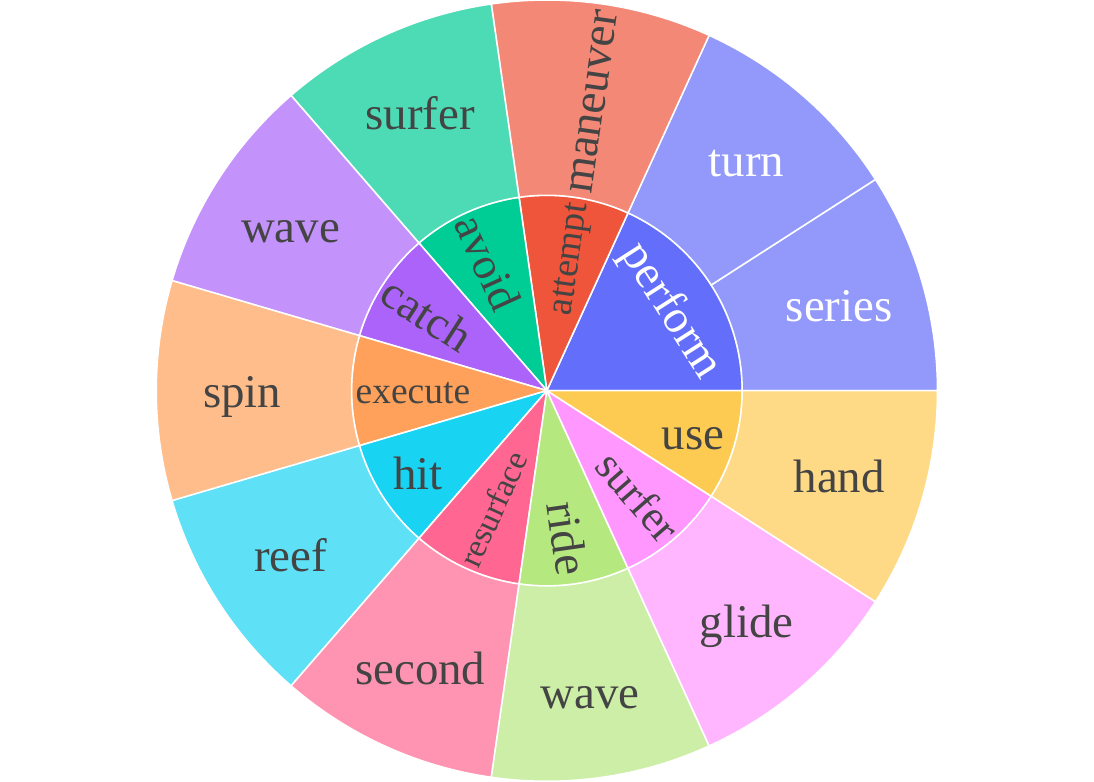}
         \caption{Surfing.}
         \label{fig:vn_8}
     \end{subfigure}
     \hfill
     \begin{subfigure}[h]{0.3\textwidth}
         \centering
         \includegraphics[width=\textwidth]{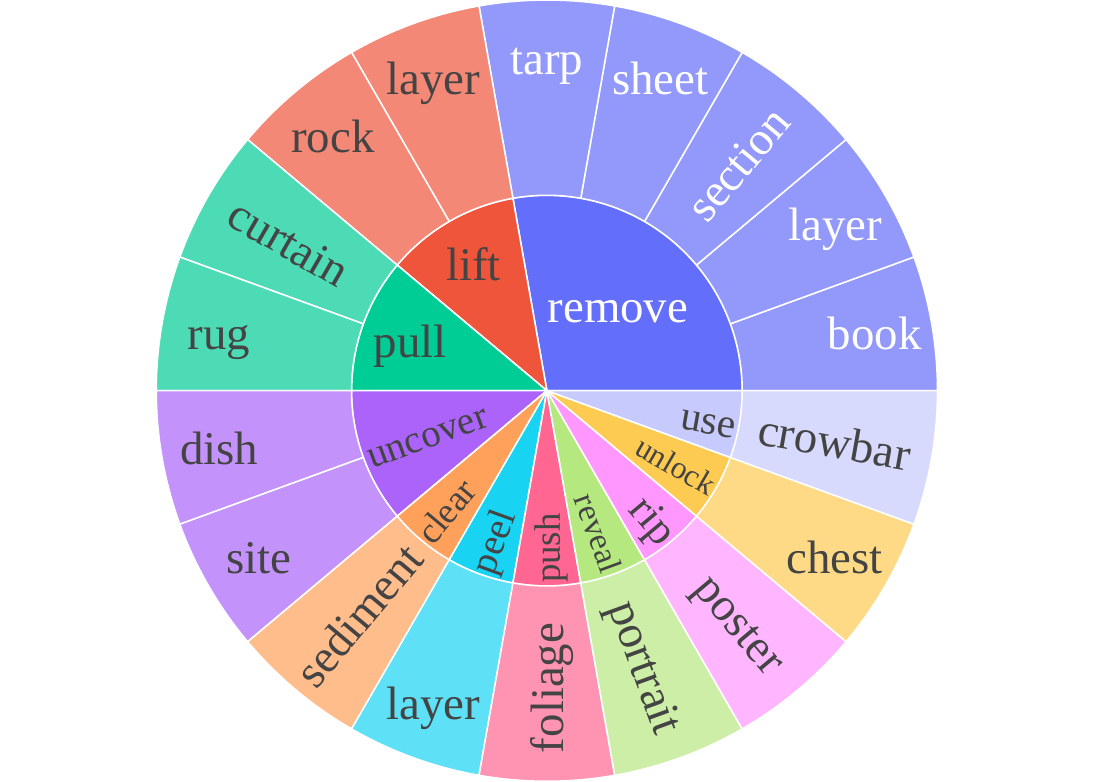}
         \caption{Uncovering something.}
         \label{fig:vn_9}
     \end{subfigure}
     \qquad
 \begin{subfigure}[h]{0.3\textwidth}
         \centering
         \includegraphics[width=\textwidth]{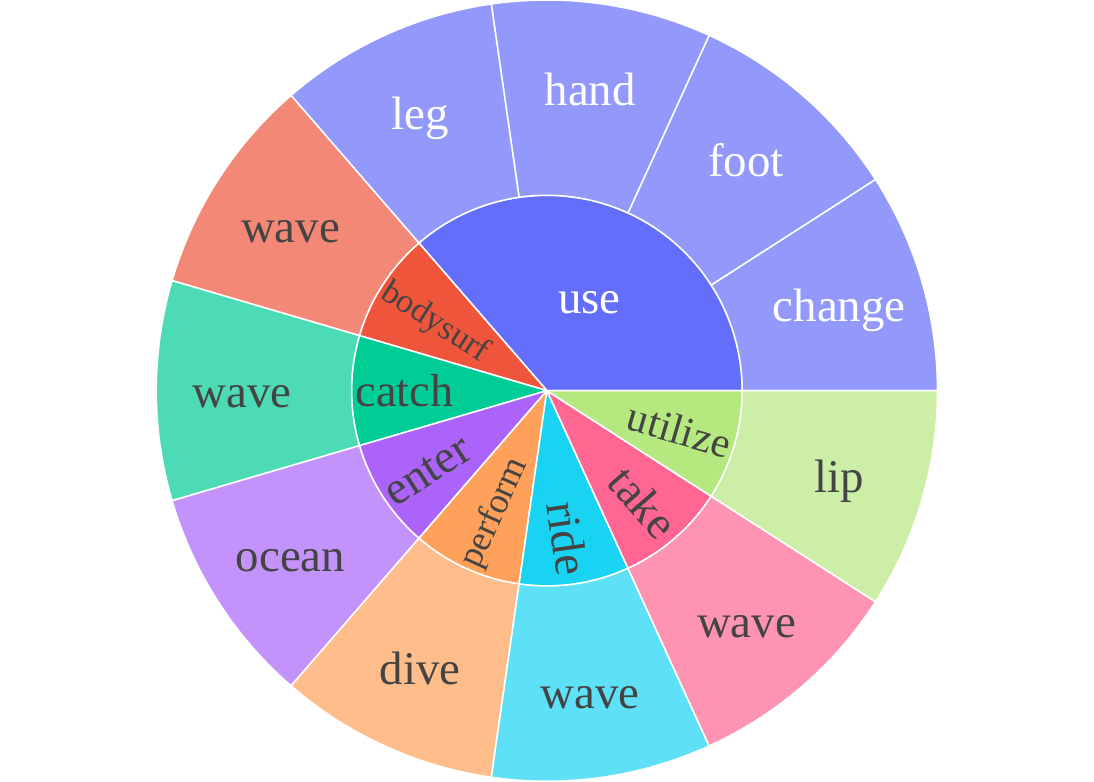}
\caption{Body surfing.}
         \label{fig:vn_10}
     \end{subfigure}
     \hfill
     \begin{subfigure}[h]{0.3\textwidth}
         \centering
         \includegraphics[width=\textwidth]{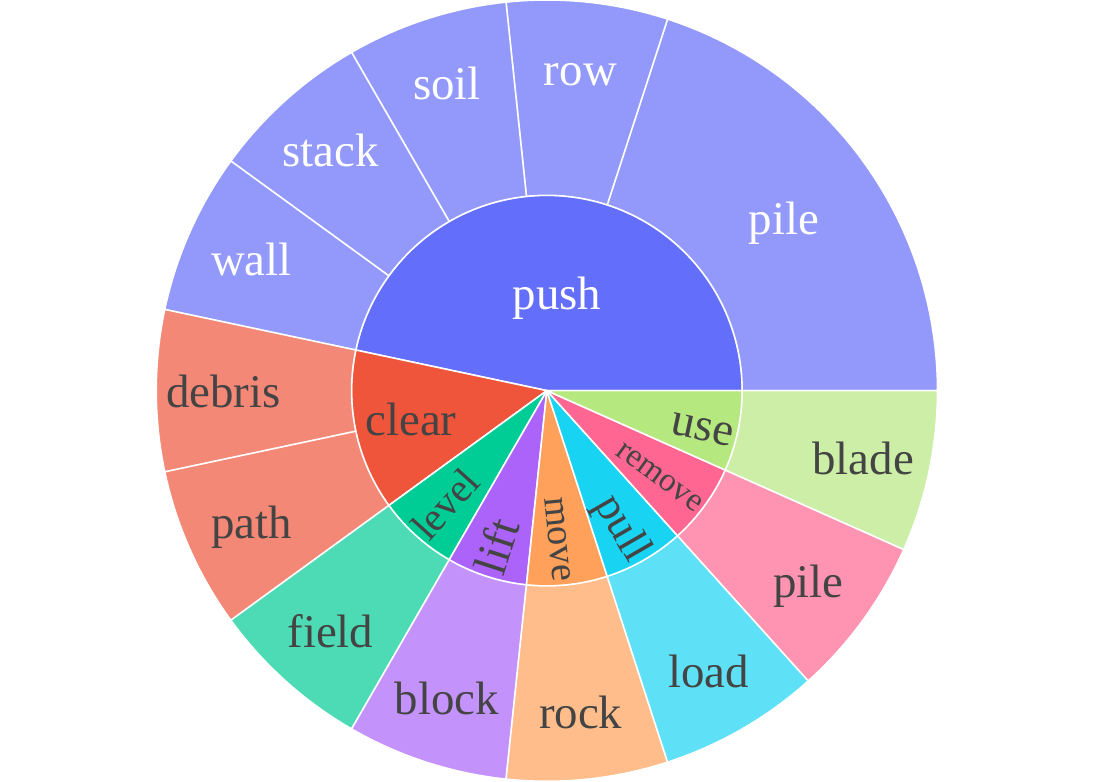}
         \caption{Bulldozing.}
         \label{fig:vn_11}
     \end{subfigure}
     \hfill
     \begin{subfigure}[h]{0.3\textwidth}
         \centering
         \includegraphics[width=\textwidth]{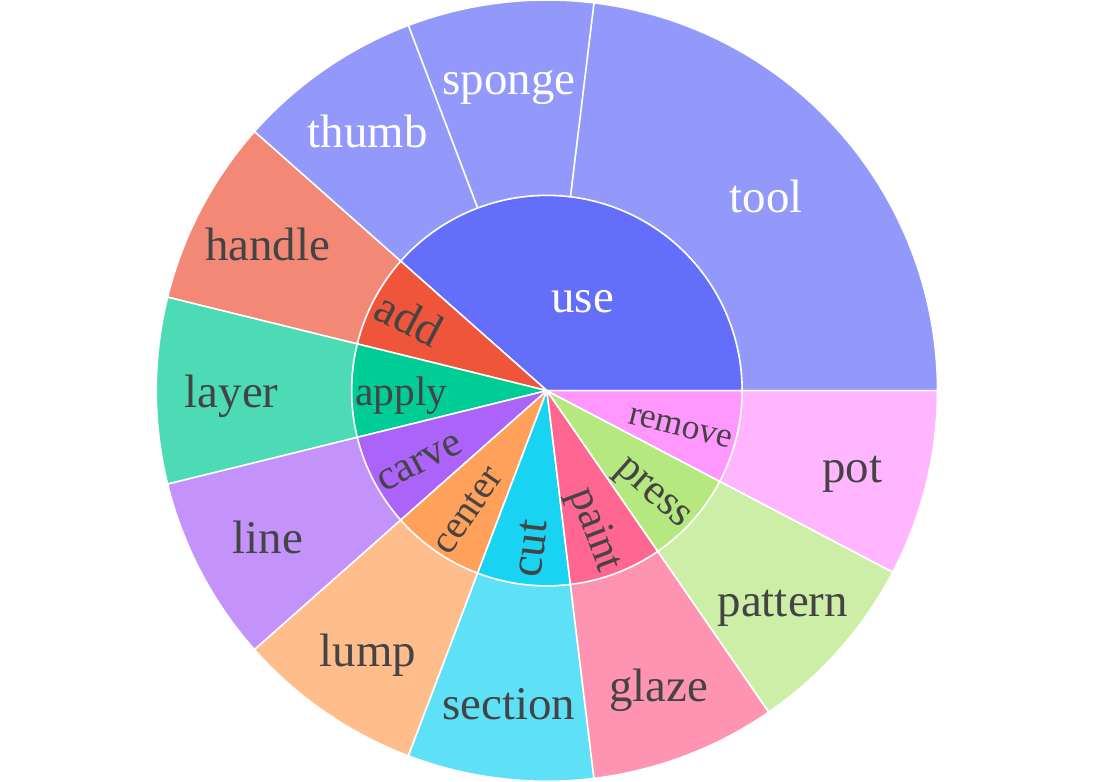}
         \caption{Clay pottery making.}
         \label{fig:vn_12}
     \end{subfigure}
        \caption{A subset of actions, and the verb-noun pairs in their respective generated prompts.}
        \label{app_fig:verb_noun_specific_actions}
\end{figure*}

Figure~\ref{app_fig:verb_noun_specific_actions} illustrates a subset of actions used in our study, along with the verb-noun pairs appearing in their respective generated prompts. The diversity of nouns and verbs in these prompts ensures that our dataset covers a wide range of contextual variations. For instance, the action *Rowing* can be associated with nouns such as *boat*, *paddle*, or *water*, leading to prompts that span different environments and interactions. Similarly, *Knitting* may involve *yarn*, *needles*, or *fabric*, allowing for a richer set of generated scenarios that helps to better evaluate video-generation models.

\section{LLM-generated captions prompt}
\label{app:llm_generated_caption_prompt}

In this section, we discuss the prompt we gave Gemini-2.0-Flash-Exp to generate a list of diverse prompts for each action, displayed in Table ~\ref{tab:video_prompt_generation}.

\section{Video-specific physical rule generation prompt}
\label{app:video_physical_rule_prompt}
In this section we discuss the prompt we gave Gemini-2.0-Flash-Exp to generate a list of 3 rules for each prompt, displayed in Table ~\ref{tab:video_physical_rule_prompt}.

\section{Upsampled Caption Examples}
\label{app:upsampled_captions}

To enhance video generation quality, we upsample captions by upsampling them to make them more specific with finer details. This process helps models generate better videos by providing additional cues about motion, environment, and object interactions.

Table~\ref{tab:upsampled_prompts} presents examples where simple captions (e.g., \textit{“A person uses nunchucks to break a stack of wooden blocks”}) are expanded into more vivid descriptions (e.g., including details about lighting, camera angles, and material properties). These enriched captions guide models toward generating more coherent and contextually accurate videos.

\section{Training Details for \auto{}}
\label{sec:training_details}

We finetune VideoCon-Physics \cite{bansal2024videophy} (7B) that acts a strong base model for semantic adherence and physical commonsense evaluation. Specifically, we use low rank adaptation \cite{hu2022lora} applied to all the transformer blocks including QKVO, gate, up, and down weight matrices. In our experiments, we set $r, \alpha=32$ and dropout=$0.05$. We finetune the base model for $3$ epochs and pick the best checkpoint using the performance on the validation set. Further, we use Adam optimizer \cite{kingma2014adam} with a linear warmup steps of $50$ steps followed by linear decay. We perform a hyperparameter search over several peak learning rates $\{5e-5, 1e-4, 5e-4, 1e-3\}$ and found $5e-4$ worked the best. We use $4$ Nvidia A6000 GPUs with a global batch size of $64$.

\section{Multimodal Prompts for Automatic Evaluation}
\label{sec:mm_prompts}

We prompt our model to generate a text response conditioned on the multimodal template $\mathcal{T}_{t}(x)$ for semantic adherence, physical commonsense, and rule tasks. Formally, 
\begin{equation}
    \mathcal{T}_{t}(x) = \begin{cases} 
      \mathcal{T}_{SA}(V, C), & t = SA \\
     \mathcal{T}_{PC}(V), & t = PC \\
     \mathcal{T}_{R}(V, R), & t = RS \\
   \end{cases}
\end{equation}
where $t$ is either semantic adherence to the caption, physical commonsense, or the rule score, $C$ is the conditioning caption, $V$ is the generated video for the caption $C$, and $R$ is the rule candidate. We provide the multimodal templates ($\mathcal{T}_{SA}(V, C)$, $ \mathcal{T}_{PC}(V)$, and $ \mathcal{T}_{R}(V, R)$). We compute the score from the model using simple autoregressive generation.

We present the prompts used for the this multimodal evaluation for semantic adherence evaluation in Figure~\ref{tikz:vta}, physical commonsense alignment in Figure~\ref{tikz:physics}, and rule scoring in Figure~\ref{tikz:rules}.

\section{Baselines Judgments}
\label{app:baseline_score_computation}

\subsection{VideoPhysics and VideoCon}
For our baseline comparisons, we obtained raw scores from VideoPhysics\cite{bansal2024videophy} and VideoCon\cite{bansal2024videocon} frameworks, which originally produced scores in the range of 0-1. To maintain consistency with our 1-5 rating scale, we normalized these scores using linear scaling. Specifically, we applied the transformation $score_{normalized} = score_{raw} \times 4 + 1$, followed by rounding to the nearest integer. This transformation maps the minimum possible score (0) to 1 and the maximum possible score (1) to 5, preserving the relative performance differences between models while making the scores directly comparable to our human evaluation ratings.

\subsection{VideoScore Regression}
For our implementation of VideoScore \cite{he2024videoscore}, we selected specific component metrics that align with our evaluation objectives. For SA, we utilized the Text-to-Video Alignment score, as it effectively measures the correspondence between video content and textual descriptions. For PC, we employed the Factual Consistency Score, which assesses the physical plausibility of events depicted in the videos.

\subsection{VideoLlava}
For VideoLlava \cite{lin2023video}, we used the same prompts shown in Appendix \ref{sec:mm_prompts}. The model produced scores directly on our 1-5 scale following the evaluation criteria provided in the prompts. 

\subsection{Gemini}
For Gemini-2.0-Flash-Exp, we leveraged its larger context window to provide more detailed and structured evaluation prompts, allowing us to specify evaluation criteria with greater granularity, and adding few-shot examples. This enabled more comprehensive assessments for semantic adherence (Figure~\ref{app_tab:gemini_sa}) and physical commonsense (Figure~\ref{app_tab:gemini_pc}).

\section{Human Annotation Interface}
\label{app:human_annotation_interface}

\begin{figure*}[h]
    \centering
\includegraphics[scale=0.6]{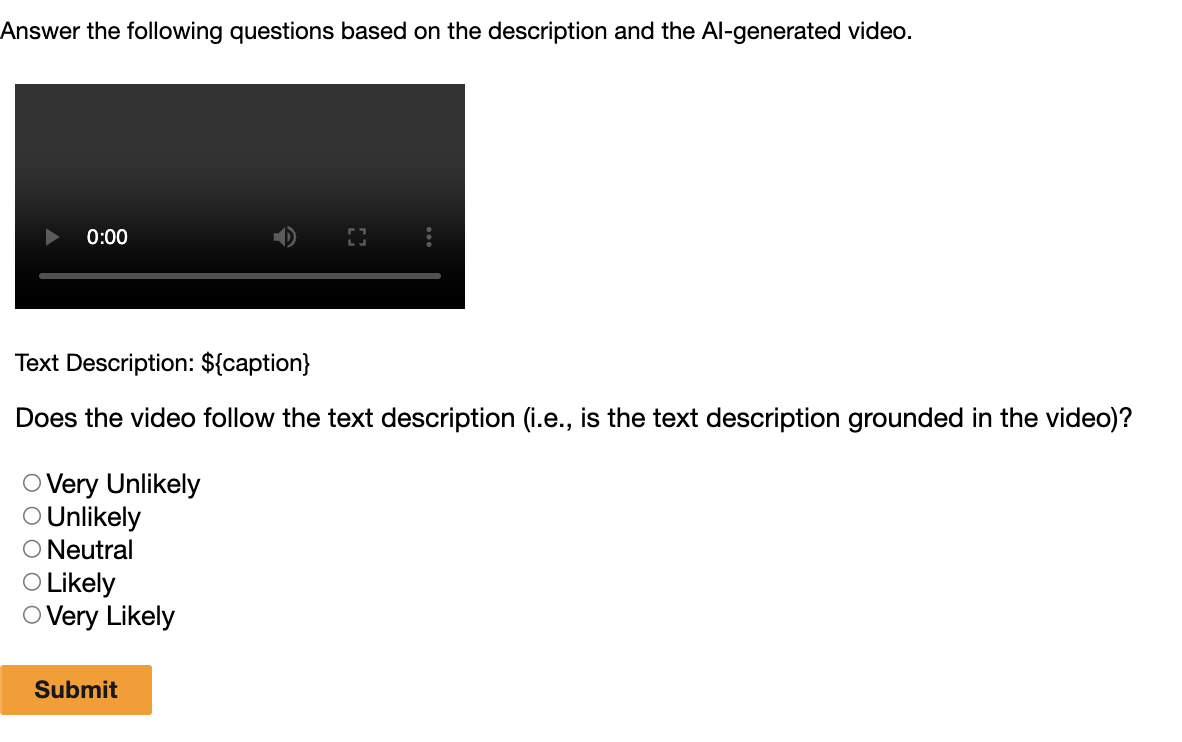}
    \caption{The screenshot of the human annotation interface for semantic adherence task.}
    \label{fig:interface_sa}
\end{figure*}

\begin{figure*}[h]
    \centering
\includegraphics[scale=0.6]{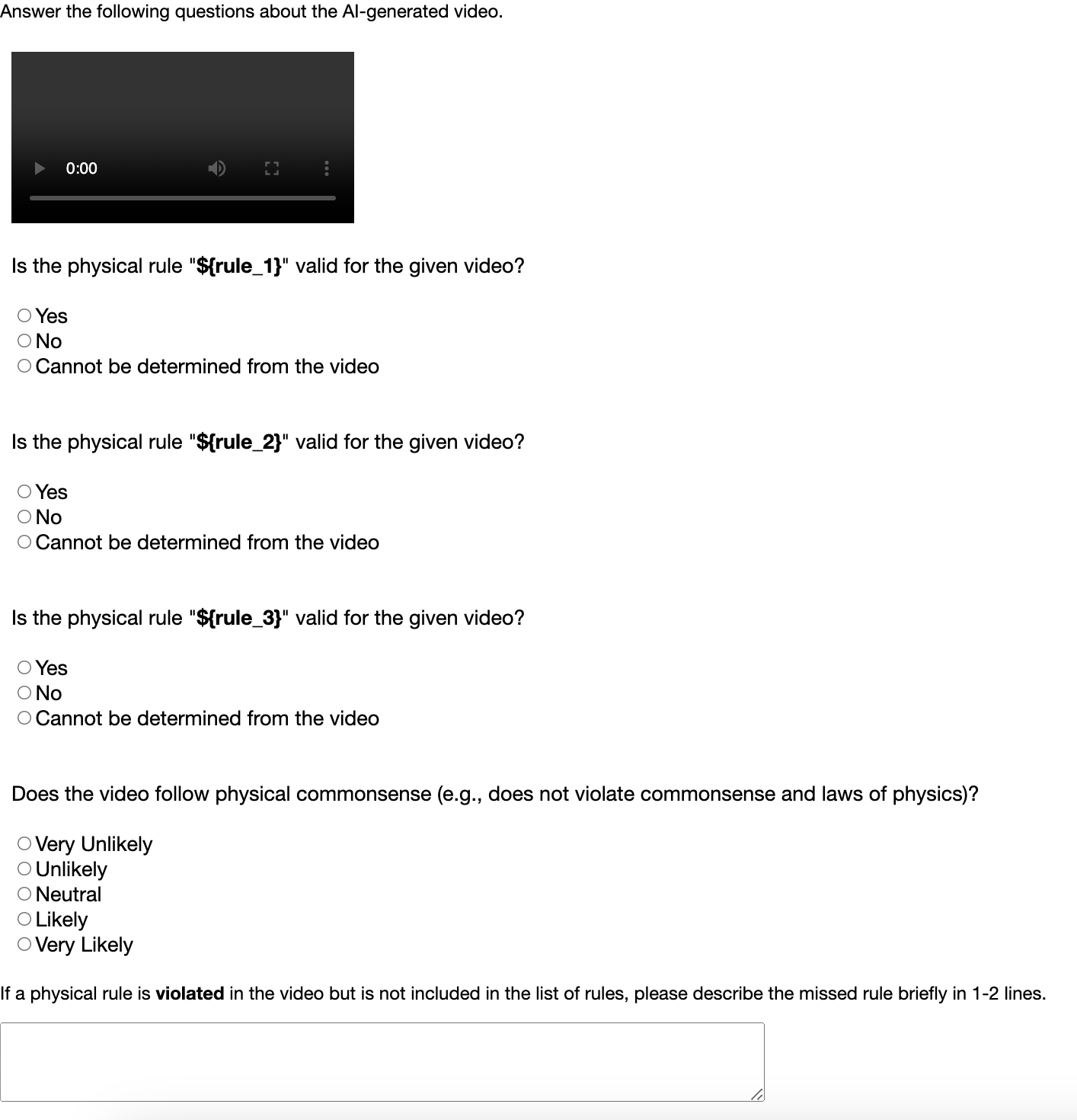}
    \caption{The screenshot of the human annotation interface for physical rule and commonsense judgment tasks.}
    \label{fig:interface_pc}
\end{figure*}

Figure~\ref{fig:interface_sa} and Figure~\ref{fig:interface_pc} showcase our human annotation interface. The interface is designed to facilitate both semantic adherence and physical commonsense evaluation. We combine rule scoring and physical commonsense assessment into a single task, allowing annotators to provide additional rules they believe are violated.

\section{Model Inference Details}
\label{app:inference_details}

Table~\ref{app_table:inference_details} summarizes the inference settings for the video generation models used in our study. The table highlights key parameters such as resolution, frame rate, guidance scale, sampling steps, and precision.

In our experiments, all models except those with a token limit of 77 (due to CLIP \cite{radford2021learning} embeddings) used upsampled captions. Models like Wan2.1-14B and CogVideoX-5B particularly benefit from these richer descriptions, enhancing the quality of their generated videos.

We evaluated the models using original and upsampled captions. Different models use different schedulers, and use different default precision levels (bf16, fp16, fp32).

For closed models, such as Ray2 and Sora, we used fewer captions, which influences their evaluation outcomes.

\begin{table*}[t]
\centering
\caption{\textbf{Inference settings for different video generation models.}}
\label{app_table:inference_details}
\resizebox{\textwidth}{!}{%
\begin{tabular}{lcccccccc}
\toprule
\textbf{Models} & \textbf{Caption} & \textbf{FPS/Video} & \textbf{Resolution} & \textbf{Frames} & \textbf{Scale} & \textbf{Steps} & \textbf{Scheduler} & \textbf{Precision} \\
& \textbf{Type} & \textbf{Length (s)} & & & \textbf{Guidance} & \textbf{Sampling} & \textbf{Noise} & \\
\midrule
Wan2.1-14B & Upsampled & 16 (4s) & 832×480 & 61 & 5 & 50 & FlowUniPCMultistepScheduler & bf16 \\
CogVideoX-5B & Upsampled & 8 (6s) & 480×720 & 49 & 6 & 50 & CogVideoXDPMScheduler & bf16 \\
Cosmos-Diffusion-7B & Upsampled & 24 (5s) & 576×576 & 120 & 7 & 60 & - & bf16 \\
HunyuanVideo-13B & Original & 15 (4s) & 320×512 & 61 & 6 & 50 & FlowMatchEulerDiscreteScheduler & fp16 \\
VideoCrafter2-1.4B & Original & 10 (3s) & 320×512 & 32 & 12 & 50 & DDIM & fp32 \\
\bottomrule
\end{tabular}%
}
\end{table*}


\section{Distribution of semantic adherence and physical commonsense scores}
\label{app:distribution_score}

\begin{figure*}[t]
     \centering
     \begin{subfigure}[h]{0.3\textwidth}
         \centering
         \includegraphics[width=\textwidth]{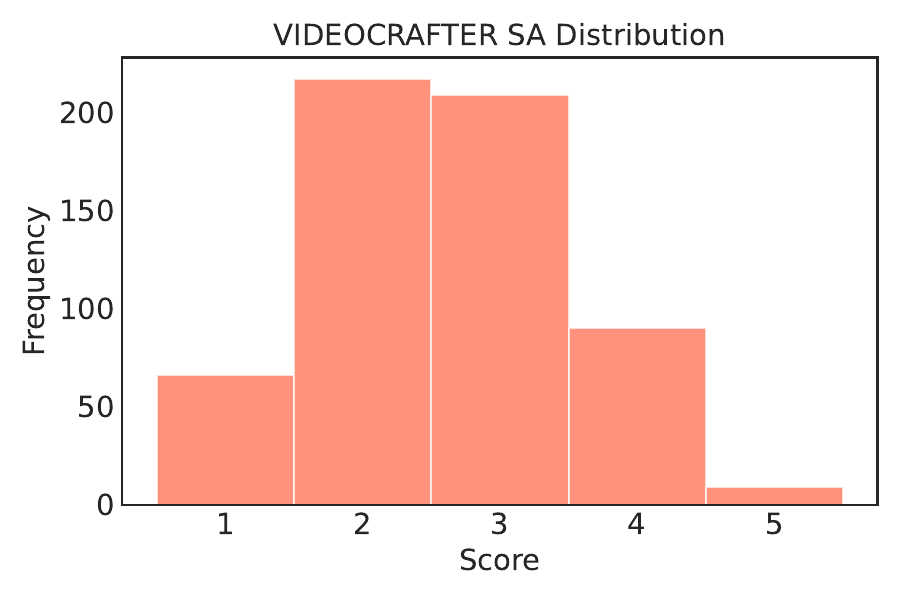}
         
         \label{fig:sa_score_1}
     \end{subfigure}
     \hfill
     \begin{subfigure}[h]{0.3\textwidth}
         \centering
         \includegraphics[width=\textwidth]{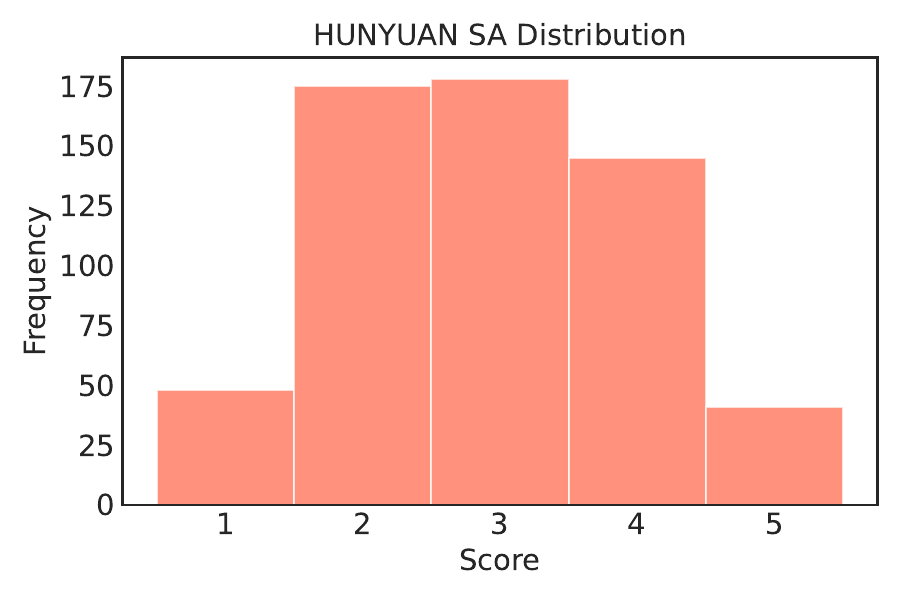}
         
         \label{fig:sa_score_2}
     \end{subfigure}
     \hfill
     \begin{subfigure}[h]{0.3\textwidth}
         \centering
         \includegraphics[width=\textwidth]{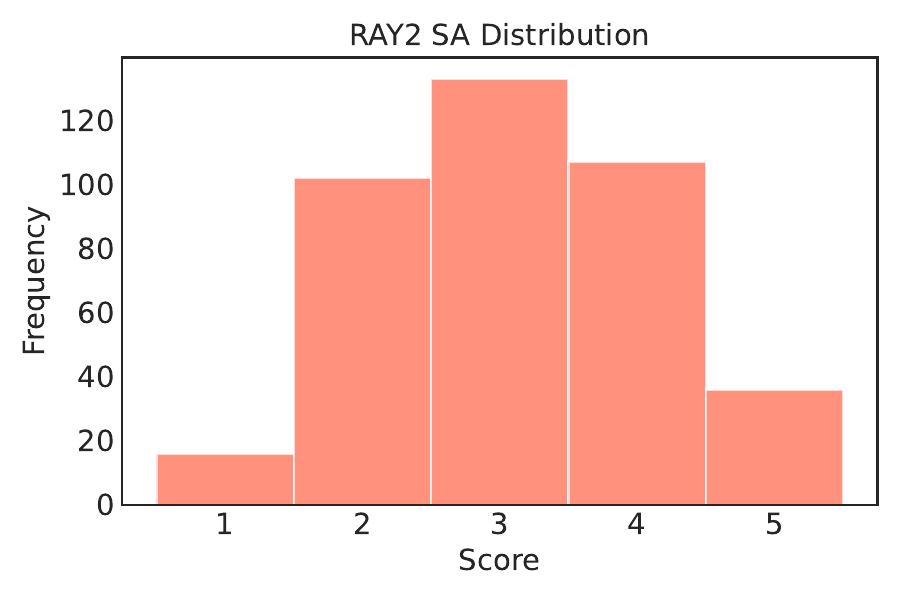}
         
         \label{fig:sa_score_3}
     \end{subfigure}
        \qquad
     \begin{subfigure}[h]{0.3\textwidth}
         \centering
         \includegraphics[width=\textwidth]{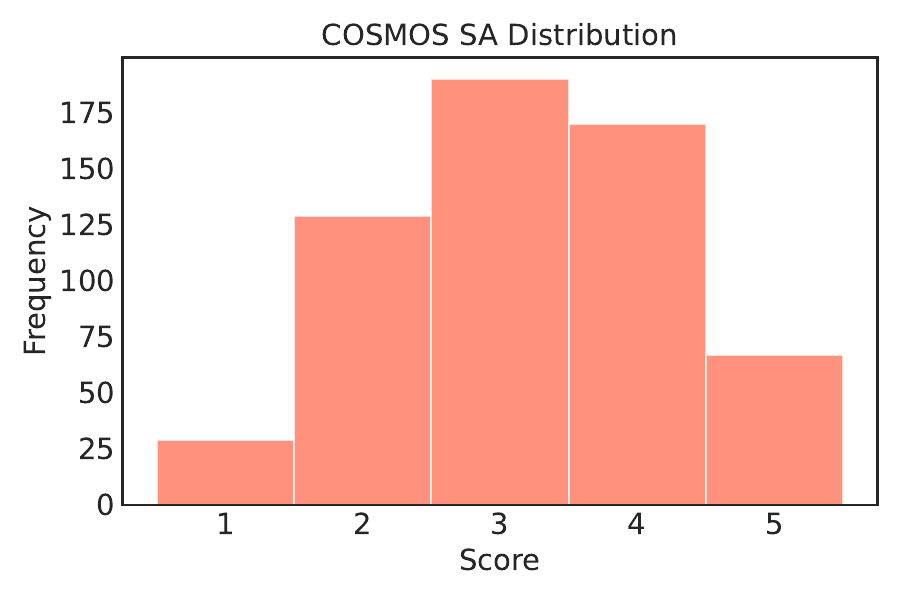}

         \label{fig:sa_score_4}
     \end{subfigure}
     \hfill
     \begin{subfigure}[h]{0.3\textwidth}
         \centering
         \includegraphics[width=\textwidth]{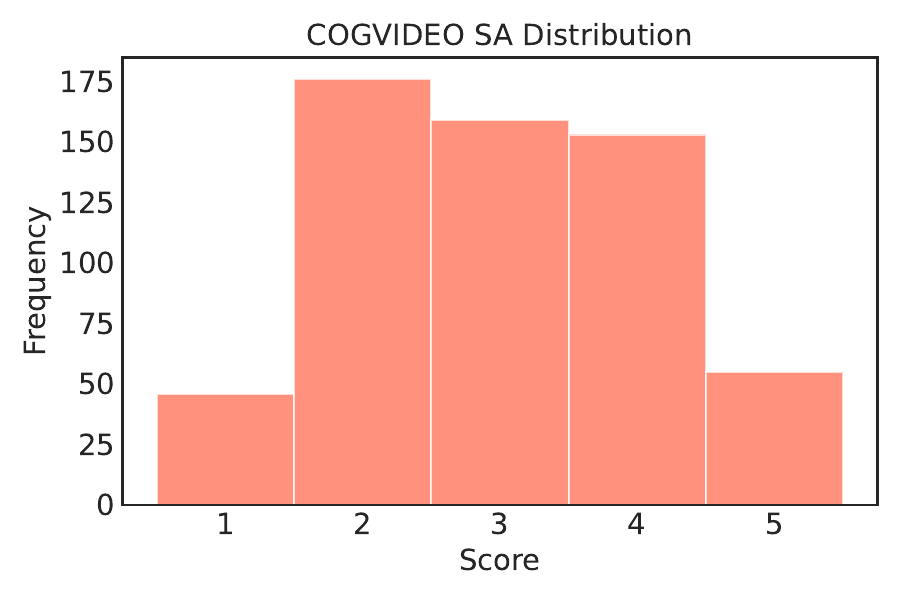}
         
         \label{fig:sa_score_5}
     \end{subfigure}
     \hfill
     \begin{subfigure}[h]{0.3\textwidth}
         \centering
         \includegraphics[width=\textwidth]{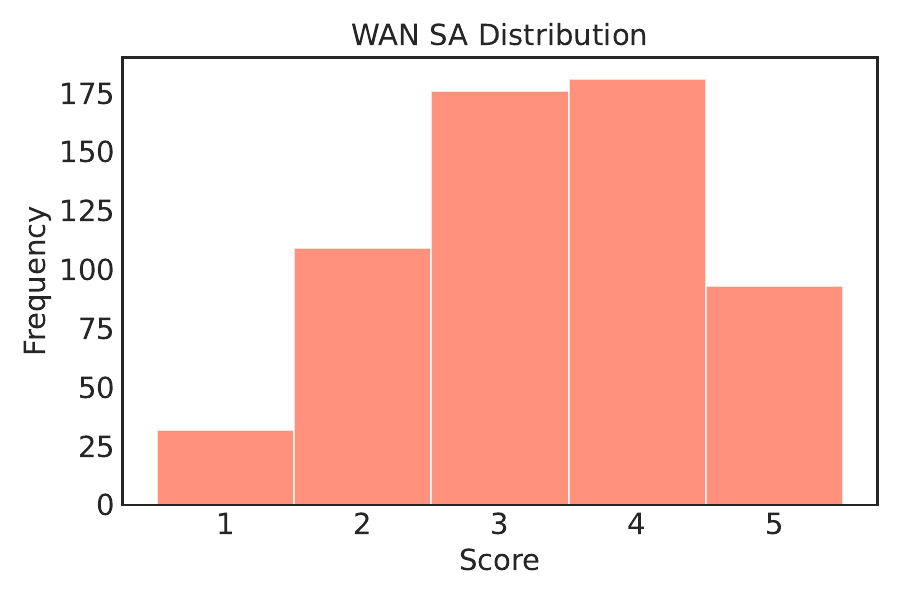}
         
         \label{fig:sa_score_6}
     \end{subfigure}
        \caption{Distribution of semantic adherence scores for various models in \name{}.}
        \label{app_fig:sa_score_distribution}
\end{figure*}

\begin{figure*}[t]
     \centering
     \begin{subfigure}[h]{0.3\textwidth}
         \centering
         \includegraphics[width=\textwidth]{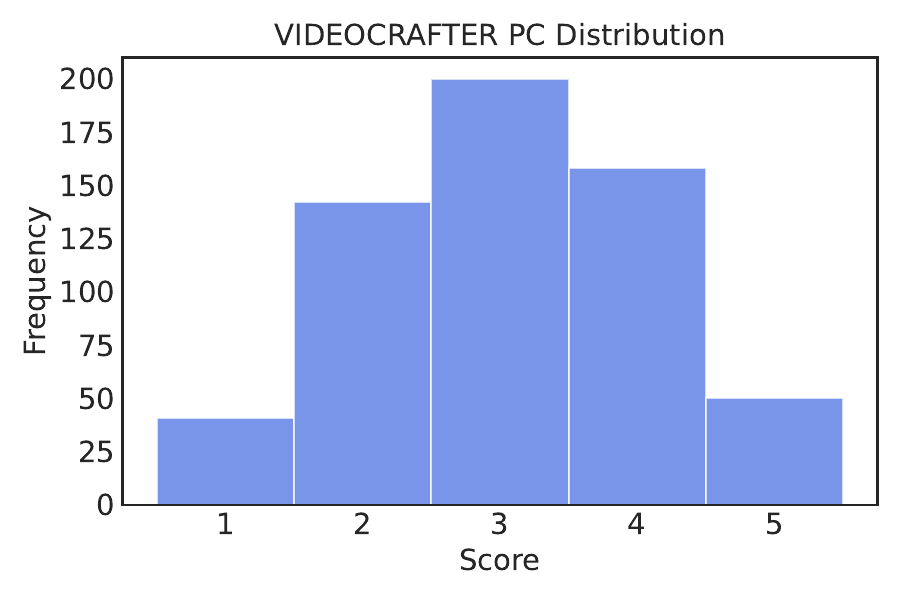}
         
         \label{fig:pc_score_1}
     \end{subfigure}
     \hfill
     \begin{subfigure}[h]{0.3\textwidth}
         \centering
         \includegraphics[width=\textwidth]{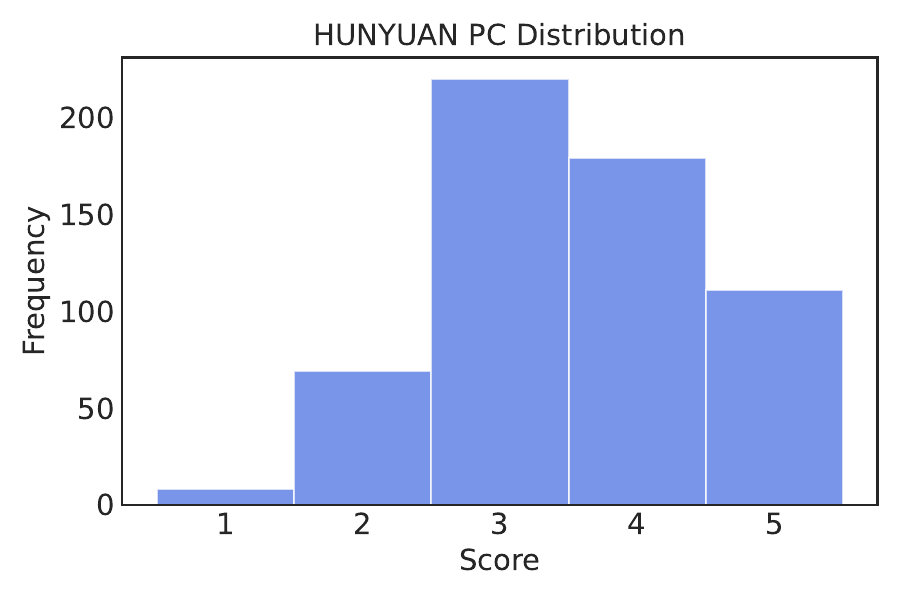}
         
         \label{fig:pc_score_2}
     \end{subfigure}
     \hfill
     \begin{subfigure}[h]{0.3\textwidth}
         \centering
         \includegraphics[width=\textwidth]{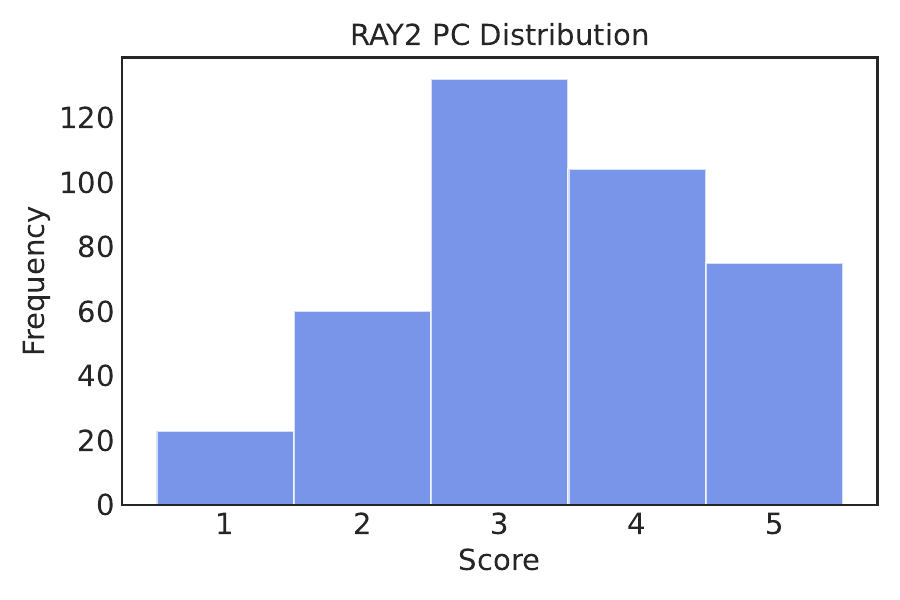}
         
         \label{fig:pc_score_3}
     \end{subfigure}
        \qquad
     \begin{subfigure}[h]{0.3\textwidth}
         \centering
         \includegraphics[width=\textwidth]{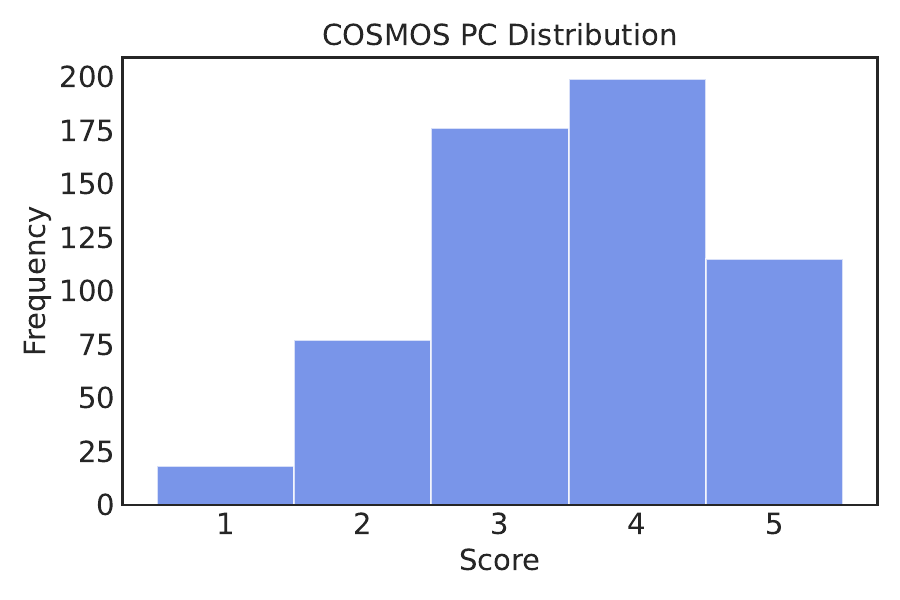}

         \label{fig:pc_score_4}
     \end{subfigure}
     \hfill
     \begin{subfigure}[h]{0.3\textwidth}
         \centering
         \includegraphics[width=\textwidth]{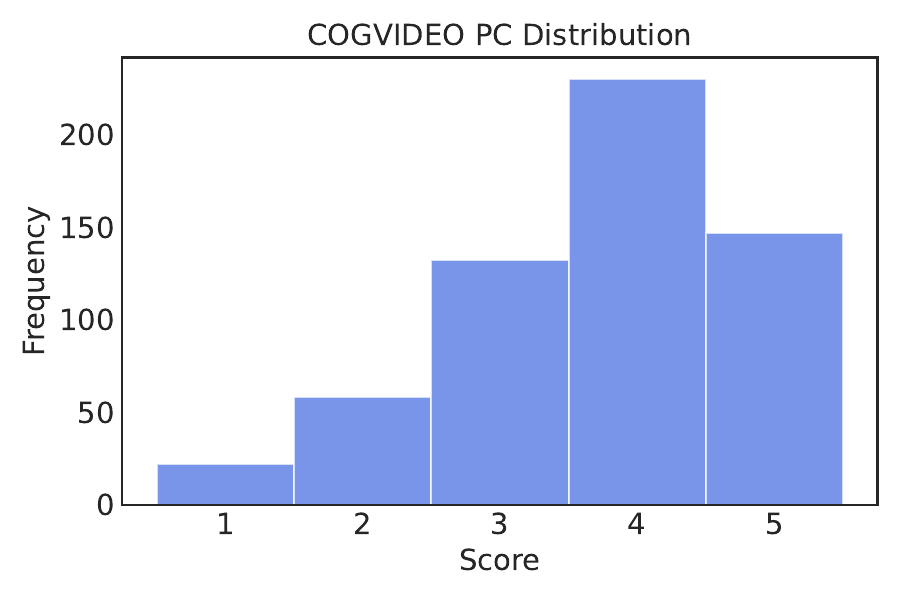}
         
         \label{fig:pc_score_5}
     \end{subfigure}
     \hfill
     \begin{subfigure}[h]{0.3\textwidth}
         \centering
         \includegraphics[width=\textwidth]{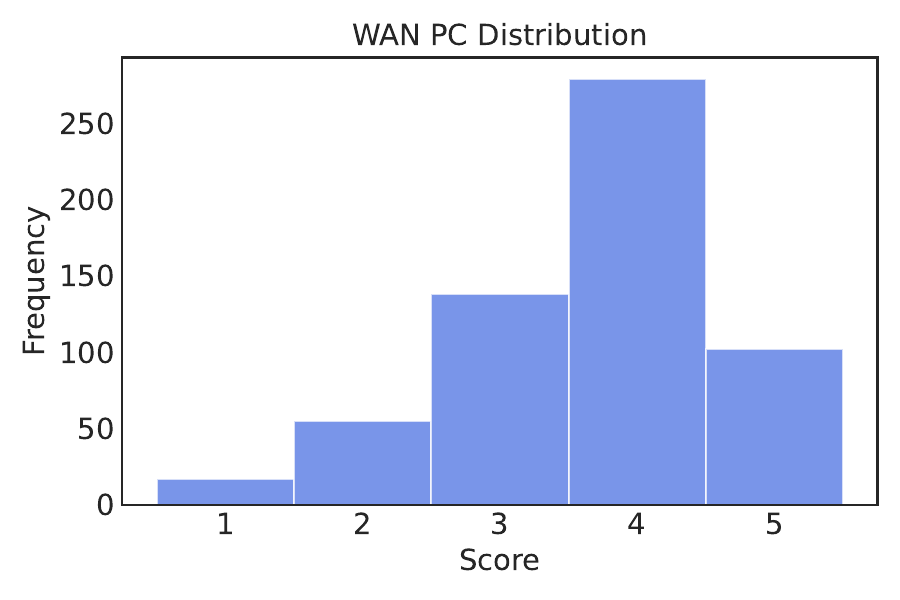}
         \label{fig:pc_score_6}
     \end{subfigure}
        \caption{Distribution of physical commonsense scores for various models in \name{}.}
        \label{app_fig:pc_score_distribution}
\end{figure*}

We present the distribution of semantic adherence (SA) and physical commonsense (PC) scores in Figure \ref{app_fig:sa_score_distribution} and \ref{app_fig:pc_score_distribution}.




\begin{table*}
    \centering
    \begin{tabular}{p{\textwidth}}
         \toprule
         \small{\texttt{mopping floor, blowing out candles, throwing water balloon, passing american football, billiards, lifting a surface with something on it until it starts sliding down, using a sledge hammer, swing, hitting baseball, hammerthrow, playing tennis, longjump, sewing, roller skating, wading through water, riding mechanical bull, pole vault, blowdryhair, tying necktie, paragliding, something falling like a rock, playing ice hockey, sailing, gymnastics tumbling, pushing something so that it slightly moves, folding clothes, poking something so lightly that it doesn't or almost doesn't move, javelinthrow, surfing, snatch weight lifting, chiseling stone, spinning something that quickly stops spinning, poking a stack of something without the stack collapsing, carving ice, throwing something, twisting (wringing) something wet until water comes out, putting something onto something else that cannot support it so it falls down, wiping something off of something, playing field hockey, juggling balls, wading through mud, shooting basketball, welding, hoverboarding, javelin throw, catching or throwing softball, hammering, knitting, putting something that can't roll onto a slanted surface so it slides down, playing polo, pouring something onto something, pulling two ends of something so that it gets stretched, using circular saw, flint knapping, backflip (human), long jump, uncovering something, pouring something into something until it overflows, dribbling basketball, poking a hole into some substance, bulldozing, peeling fruit, parallelbars, playing darts, spinning something so it continues spinning, luge, pushing something so it spins, curling (sport), riding unicycle, throwing discus, folding paper, ripping paper, trapezing, playing pinball, burying something in something, throwing axe, wrapping present, yarn spinning, tying shoe laces, flying kite, tightrope walking, using a paint roller, using a wrench, sharpening pencil, pizzatossing, catching or throwing baseball, catching or throwing frisbee, playing kickball, golf, nunchucks, pouring something out of something, opening bottle (not wine), unfolding something, stuffing something into something, canoeing or kayaking, rolling something on a flat surface, playing ping pong, punching bag, picking fruit, poking something so that it spins around, balancebeam, parasailing, jumprope, bungee jumping, drop kicking, hammer throw, using segway, biking through snow, swimming, making snowman, rowing, attaching something onto something, something colliding with something and both come to a halt, blasting sand, throwdiscus, tying knot (not on a tie), digging something out of something, trimming shrubs, inflating balloons, bouncing on trampoline, spinning poi, shot put, pushing something onto something, something falling like a feather or paper, letting something roll down a slanted surface, cutting, chopping wood, breaststroke, hurdling, ice climbing, popping balloons, throwing knife, bending something so that it deforms, playing cricket, shaping bread dough, bobsledding, smashing, blowing bubble gum, something colliding with something and both are being deflected, breaking boards, somersaulting, skateboarding, squeezing something, ropeclimbing, bending something until it breaks, yoyo, bouncing on bouncy castle, hula hooping, letting something roll along a flat surface, pushing something off of something, lifting a surface with something on it but not enough for it to slide down, throwing snowballs, shoveling snow, rolling pastry, tossing coin, threading needle, skijet, clay pottery making, pommelhorse, playing squash or racquetball, pushing something so that it falls off the table, bodysurfing, twisting something, poking a hole into something soft, tearing something into two pieces, walking through snow, putting something that cannot actually stand upright upright on the table so it falls on its side, poking something so that it falls over, poking a stack of something so the stack collapses, pushing something so that it almost falls off but doesn't, kicking soccer ball, tying bow tie, wood burning (art), putting something that can't roll onto a slanted surface so it stays where it is, extinguishing fire, ice skating, playing badminton, archery, folding napkins, soccerjuggling, blowing leaves, bowling, mountain climber (exercise), jetskiing, pouring something into something, pouring beer, pulling two ends of something so that it separates into two pieces, volleyballspiking, poking something so it slightly moves, smoking, tearing something just a little bit, rock climbing, letting something roll up a slanted surface so it rolls back down, blowdrying hair, folding something, riding scooter, playing volleyball}}\\\bottomrule
    \end{tabular}
    \caption{Actions List: A comprehensive collection of all physical actions that were used test video models' ability to produce videos that align with physical commonsense}
      \label{app_table:all_actions_list}
\end{table*}

\begin{table*}
    \centering
    \begin{tabular}{p{\textwidth}}
    \toprule
        \texttt{blowing out candles, playing squash or racquetball, passing american football, billiards, backflip (human), tearing something into two pieces, pouring something into something until it overflows, lifting a surface with something on it until it starts sliding down, using a sledge hammer, swing, hitting baseball, peeling fruit, parallelbars, putting something that cannot actually stand upright upright on the table so it falls on its side, playing darts, poking something so that it falls over, chopping wood, throwing discus, poking a stack of something so the stack collapses, pushing something so that it almost falls off but doesn't, kicking soccer ball, pole vault, popping balloons, ripping paper, blowdryhair, throwing knife, playing ice hockey, throwing axe, bending something so that it deforms, yarn spinning, playing cricket, tightrope walking, gymnastics tumbling, playing badminton, archery, pizzatossing, catching or throwing baseball, catching or throwing frisbee, chiseling stone, spinning something that quickly stops spinning, folding napkins, golf, throwing something, nunchucks, opening bottle (not wine), bending something until it breaks, wiping something off of something, playing field hockey, balancebeam, pushing something off of something, throwing snowballs, pouring beer, bungee jumping, drop kicking, pulling two ends of something so that it separates into two pieces, catching or throwing softball, rowing, hammering, letting something roll up a slanted surface so it rolls back down, playing polo}\\\bottomrule
    \end{tabular}%
    \caption{Hard Actions List: A comprehensive collection of challenging physical actions in the "Hard" category used to test video models' ability to produce videos that align with physical commonsense}
    \label{app_table:hard_actions_list}
\end{table*}

\begin{figure*}[htbp]
\centering
\resizebox{\linewidth}{!}{
\begin{tabular}{p{1.3\linewidth}}
\toprule
Task: Generate 20 realistic, detailed, and diverse video prompts based on a given action. The prompts must focus solely on clear, observable physical actions and interactions between characters, objects, and their environments that can be easily represented visually in a video. \\

\textbf{Requirements:} \\

1. \textbf{Physical Actions with Direct Outcomes:} Each prompt must clearly describe observable physical actions involving objects (e.g., bows, arrows, targets, or archers) and their direct, visible outcomes (e.g., an arrow striking a target, string tension when pulled).\\ 

2. \textbf{Exclude Non-Visual Details:} Avoid including sensory or inferred details like sounds, smells, emotions, or mental states (e.g., "focus," "determination," or "excitement").\\

3. \textbf{Avoid Subtle Movements:} Do not include descriptions of small-scale or subtle motions that are hard to detect in standard video playback (e.g., trembling hands, tiny vibrations, or imperceptible shifts).\\

4. \textbf{Concrete, Not Abstract:} Steer clear of poetic, artistic, or abstract descriptions (e.g., “the shimmering flight of an arrow” or “a poised stance”) and instead focus on tangible, visible actions.\\

5. \textbf{Diverse Scenarios:} Use a variety of settings, character types, objects, and equipment to make the prompts diverse and adaptable to different video generation contexts.\\

6. \textbf{Specific Visual Actions:} Center prompts on observable, specific actions such as pulling a bowstring, releasing an arrow, an arrow’s flight path, or its interaction with targets. Highlight the physical interactions between objects (e.g., wood, metal, or other materials) and environments.\\

\textbf{Examples:}\\

\textbf{Good Prompts for the Action “Archery”}\\
"An archer draws the bowstring back to full tension, then releases the arrow, which flies straight and strikes a bullseye on a paper target."\\
"A compound bow is fired, and the arrow pierces through a stack of hay bales, stopping halfway through."\\
"An archer adjusts their stance, takes aim, and releases an arrow, which embeds itself into a foam target with visible force."\\
"A crossbow is loaded with a bolt, cocked, and fired, hitting a glass bottle, which shatters on impact."\\
"An arrow is shot toward a wooden target, splinters flying as it embeds deep into the surface."\\

\textbf{Bad Prompts for the Action “Archery”}\\
"The archer feels confident as they aim their arrow at the target." \#\# (Describes inferred mental state.)\\
"The arrow flies silently and gracefully toward the target." \#\# (Includes non-visual elements and artistic descriptions.)\\
"The string vibrates slightly after the arrow is released." \#\# (Focuses on subtle, hard-to-detect motion.)\\
"The archer holds their bow with a poised and elegant stance." \#\# (Focuses on posture instead of action.)\\

Now, Generate up to 20 prompts relevant to the given action: "{}" while adhering to the above criteria. Use various objects, environments, and physical interactions to ensure diversity and realism. Output in a python-parsable list of strings, stored in the variable ```prompts```.\\
\bottomrule
\end{tabular} }

\caption{LLM Prompt for Generating Realistic Video Captions.}
\label{tab:video_prompt_generation}
\end{figure*}

\begin{figure*}[htbp]
\centering
\resizebox{\linewidth}{!}{
\begin{tabular}{p{1.3\linewidth}}
\toprule
\textbf{Task Description:} Generate simple, clear behaviors/rules for describing visible, real-world physical interactions in a given video scene. These behaviors will be used to determine whether a video aligns with realistic physical interactions. \\

\textbf{Key Requirements:} \\

1. \textbf{Observation-Centric:} Focus strictly on what is visually observable in the video (e.g., motion, deformation, changes in shape or position). Avoid abstract concepts or invisible factors (e.g., forces, emotions, intentions).\\

2. \textbf{Action-Oriented:} The behaviors should directly describe interactions between materials or objects, not actions or intentions of individuals. The focus should be on the materials' behavior (e.g., how the gauze behaves when wrapped, how it stretches, or conforms), not the motions of the hands or other actors.\\

3. \textbf{Simple and Testable:} The behaviors must be concise, clear, and directly testable from the video. Avoid jargon or unnecessary technical details.\\

4. \textbf{Associated Physical Laws:} Each rule should include the physical law(s) it exemplifies. Use applicable laws from the following list: \\
\hspace{1em} \textit{"Gravity", "Buoyancy", "Elasticity", "Friction", "Conservation of Mass",}\\
\hspace{1em} \textit{"Reflection", "Refraction", "Interference and Diffraction", "Tyndall Effect"}\\
\hspace{1em} \textit{"Sublimation", "Melting", "Boiling", "Liquefaction"}\\
\hspace{1em} \textit{"Hardness", "Solubility", "Dehydration Properties", "Flame Reaction"}\\
Other laws are acceptable (e.g., Archimedes' Principle), but they must be well-known physical laws. Rules should only be related to visible, observable physical properties of materials, such as shape, deformation, or material properties, and not vague statements that are not strictly physical phenomena.\\

5. \textbf{Guaranteed:} The behaviors must occur in the video as specified from the prompt description. Events that are not guaranteed to occur from the prompt description should not appear in the rules list. Do not add information beyond what is present in the caption.\\

\textbf{Example Prompt:} \textit{A basketball bounces up and down}\\

\textbf{Good Examples of Behaviors:}\\
- The ball is faster at the bottom of the bounce. (Gravity)\\
- The ball moves up after bouncing off the floor. (Elasticity)\\

\textbf{Bad Examples of Behaviors:}\\
- The gravity acts on the ball. (\#\# Not specific enough to test.)\\
- The ball deforms when it collides with the floor. (\#\# Difficult to test.)\\
- The ball is caught if it is being dribbled. (\#\# Not guaranteed from the prompt.)\\
- The ball stops bouncing after a while. (\#\# Not guaranteed; the prompt does not specify video duration.)\\

\textbf{Example Prompt:} \textit{A nurse applies a bandage around a patient's arm.}\\

\textbf{Good Examples of Behaviors:}\\
- The bandage is a flexible and solid material. (Elasticity)\\
- The bandage roll becomes smaller as it is unrolled. (Conservation of Mass)\\

\textbf{Bad Examples of Behaviors:}\\
- The bandage stops the bleeding. (\#\# No bleeding mentioned in the prompt.)\\
- The bandage secures the wound tightly. (\#\# Tightness is subjective and not visually testable.)\\
- The bandage is elastic and stretches when pulled. (\#\# Elasticity applies, but not directly related to the video's action.)\\

Now, generate a suitable list of 3 behaviors for this prompt: \\
\textbf{A person shoots an explosive arrow at a metal target.}\\

Output in a Python-comprehensible format with:\\
```[("rule1", ["Melting","Gravity"]), ("rule2", ["Conservation of Momentum"]), ..., etc.]```\\
Stored in the variable: ```behaviors```.\\
\bottomrule
\end{tabular}}
\caption{LLM Prompt for Generating Physical Rules.}
\label{tab:video_physical_rule_prompt}
\end{figure*}

\begin{table*}[t]
\caption{\small{\textbf{Sample prompts in the \name{}.} We present the action and prompt, along with their category—either focusing on physical activities or object interactions. Additionally, we highlight the potential physical principle associated with each prompt.}}
\label{tab:action_category_examples}
\centering
\label{tab:actions_prompts}
\resizebox{\textwidth}{!}{%
\begin{tabular}{lp{6cm}cc}
\toprule
\textbf{Action} & \textbf{Prompt} & \textbf{Physical Principle} & \textbf{Category} \\ \midrule
Canoeing & A person uses a kayak paddle to push their kayak up the bank of a river. & Buoyancy & Physical Activity  \\ \midrule
Riding Unicycle & A person on a unicycle stops abruptly, putting a foot down to regain balance. & Inertia & Physical Activity \\\midrule
Pushing something so it spins & A chef pushes a pizza spinning on a tray with their hand. & Conversation of Momentum & Object Interaction \\
\bottomrule
\end{tabular}%
}
\end{table*}
 
\begin{table*}[ht]
\centering
\resizebox{\textwidth}{!}{%
\begin{tabular}{p{0.3\textwidth}|p{0.67\textwidth}}
\hline
\textbf{Original Caption}  & \textbf{Upsampled Caption} \\ \hline
A person uses nunchucks to break a stack of wooden blocks. & In a dynamic display of martial arts prowess, a skilled practitioner wields a pair of nunchucks, their hands clad in black gloves that enhance grip and safety. The scene unfolds in a dimly lit, industrial setting, where a stack of wooden blocks, meticulously arranged in a pyramid, awaits the test of strength. The camera captures the moment with a static focus, highlighting the nunchucks' fluid motion as they swing through the air, their polished wooden surfaces glinting under the soft, ambient light. \\ \hline
A player throws a softball sidearm, and the ball spins as it travels through the air. & In a sun-drenched outdoor sports arena, a dynamic softball game unfolds, captured with cinematic precision. The camera, positioned at a low angle, focuses on a player in a sleek black jersey, their arm poised in a powerful sidearm throwing motion. The softball, a vibrant white against the lush green field, spins rapidly as it leaves their hand, tracing a graceful arc through the air. \\ \hline
A small rock tumbles down a steep, rocky hillside. & In a breathtaking display of nature's raw power, a small, dark rock careens down a steep, rugged hillside, its descent punctuated by a cascade of displaced soil and smaller stones. The camera, positioned at a static angle, captures the dynamic interplay of gravity and friction as the rock bounces and rolls, kicking up a cloud of dust that dances in the sunlight. \\ \hline
An athlete throws a hammer, with the hammer head clearly visible in rotation. & In a breathtaking outdoor setting, a skilled athlete stands poised, clad in a sleek black and white athletic uniform, their focus unwavering as they prepare to hurl a gleaming hammer. The hammer rotates gracefully, its head spinning in a mesmerizing blur against the backdrop of a clear blue sky. \\ \hline
A person on a unicycle stops abruptly, putting a foot down to regain balance. & In a serene outdoor setting, a solitary figure clad in a vibrant red jacket and black pants gracefully navigates a narrow, winding path on a sleek black unicycle. The camera captures the scene with a static shot, allowing the viewer to fully appreciate the rider's skill and balance as they glide effortlessly forward. \\ \hline
\end{tabular}%
}
\caption{Example of upsampled captions given to video generation models}
\label{tab:upsampled_prompts}
\end{table*}

\begin{figure*}[htbp]
\centering
\resizebox{\linewidth}{!}{
\begin{tabular}{p{1.3\linewidth}}
\toprule
\textbf{Task Description:} Evaluate the alignment between a video and its accompanying caption:  
\textit{"A badminton shuttlecock is served underhand, traveling across the net and landing within the service box."} \\

\textbf{Evaluation Criteria:} \\
1. **Entities and Objects:** Do the objects, entities, and subjects mentioned in the caption appear in the video?\\
2. **Actions and Events:** Are the actions, events, or interactions described in the caption clearly depicted in the video?\\
3. **Temporal Consistency:** If the caption describes a sequence or progression of events, does the video follow the same temporal order?\\
4. **Scene and Context:** Does the overall scene (e.g., background, setting) match the description in the caption?\\

\textbf{Instructions for Scoring:} \\
- **1:** No alignment. The video does not match the caption at all (e.g., different objects, events, or scene).\\
- **2:** Poor alignment. Only a few elements of the caption are depicted, but key objects or events are missing or incorrect.\\
- **3:** Moderate alignment. The video matches the caption partially, but there are inconsistencies or omissions.\\
- **4:** Good alignment. Most elements of the caption are depicted correctly in the video, with minor issues.\\
- **5:** Perfect alignment. The video fully adheres to the caption with no inconsistencies.\\

\textbf{Example Prompt:}  
\textit{"A badminton shuttlecock is served underhand, traveling across the net and landing within the service box."} \\

\textbf{Example Responses:} \\
\textbf{Score:} 3  
\textbf{Explanation:} The shuttlecock is present, and an underhand serve is performed. However, the landing position is unclear, and the trajectory is partially obstructed. \\

\bottomrule
\end{tabular}
}
\caption{Gemini Prompt for Evaluating Semantic Adherence.}
\label{app_tab:gemini_sa}
\end{figure*}

\begin{figure*}[htbp]
\centering
\resizebox{\linewidth}{!}{
\begin{tabular}{p{1.3\linewidth}}
\toprule
\textbf{Task Description:} Evaluate whether the video follows physical commonsense. This judgment is based solely on the video itself and does not depend on the caption. \\

\textbf{Evaluation Criteria:} \\
1. **Object Behavior:** Do objects behave according to their expected physical properties (e.g., rigid objects do not deform unnaturally, fluids flow naturally)?\\
2. **Motion and Forces:** Are motions and forces depicted in the video consistent with real-world physics (e.g., gravity, inertia, conservation of momentum)?\\
3. **Interactions:** Do objects interact with each other and their environment in a plausible manner (e.g., no unnatural penetration, appropriate reactions on impact)?\\
4. **Consistency Over Time:** Does the video maintain consistency across frames without abrupt, unexplainable changes in object behavior or motion?\\

\textbf{Instructions for Scoring:} \\
- **1:** No adherence to physical commonsense. The video contains numerous violations of fundamental physical laws.\\
- **2:** Poor adherence. Some elements follow physics, but major violations are present.\\
- **3:** Moderate adherence. The video follows physics for the most part but contains noticeable inconsistencies.\\
- **4:** Good adherence. Most elements in the video follow physical laws, with only minor issues.\\
- **5:** Perfect adherence. The video demonstrates a strong understanding of physical commonsense with no violations.\\

\textbf{Example Responses:} \\
\textbf{Score:} 2  
\textbf{Explanation:} The ball's motion is inconsistent with gravity; it hovers momentarily before falling. Additionally, object interactions lack expected momentum transfer, suggesting physics inconsistencies.\\

\bottomrule
\end{tabular}
}
\caption{Gemini Prompt for Evaluating Physical Commonsense.}
\label{app_tab:gemini_pc}
\end{figure*}

\begin{figure*}[h!]
\centering
\resizebox{0.8\linewidth}{!}{ 
\begin{tikzpicture}
    \node[draw, rounded corners, fill=blue!10, align=left, text width=0.8\linewidth] { 
    \underline{\textbf{Semantic Adherence}}: \\[0.5em]
    \textbf{Given:} \textbf{V} (Video), \textbf{T} (Caption) \\[0.5em]
    \textbf{Instruction (I):} \textit{[\textbf{V}] Does this video match the description: "[\textbf{T}]"?} \\ 
    \textit{Please rate the video on a scale from 1 to 5, where 5 indicates a perfect match and 1 indicates no relevance.} \\[0.5em]
    \textbf{Response (R):} \textit{1, 2, 3, 4, or 5}
    };
\end{tikzpicture}
}
\caption{Template for assessing semantic adherence using a multi-modal VLM.}
\label{tikz:vta}
\end{figure*}

\begin{figure*}[h!]
\centering
\resizebox{0.8\linewidth}{!}{ 
\begin{tikzpicture}
    \node[draw, rounded corners, fill=blue!10, align=left, text width=0.8\linewidth] { 
    \underline{\textbf{Physical Commonsense}}: \\[0.5em]
    \textbf{Given:} \textbf{V} (Video) \\[0.5em]
    \textbf{Instruction (I):} \textit{[\textbf{V}] Does this video adhere to physical laws?} \\ 
    \textit{Rate the video on a scale from 1 to 5, where 5 means full compliance and 1 means significant violations.} \\[0.5em]
    \textbf{Response (R):} \textit{1, 2, 3, 4, or 5}
    };
\end{tikzpicture}
}
\caption{Template for assessing physical commonsense in video-based evaluation.}
\label{tikz:physics}
\end{figure*}

\begin{figure*}[h!]
\centering
\resizebox{0.8\linewidth}{!}{ 
\begin{tikzpicture}
    \node[draw, rounded corners, fill=blue!10, align=left, text width=0.8\linewidth] { 
    \underline{\textbf{Rule Validation}}: \\[0.5em]
    \textbf{Given:} \textbf{V} (Video), \textbf{R} (Physical Rule) \\ [0.5em]
    \textbf{Instruction (I):} \textit{[\textbf{V}] Does the video follow the physical rule: "[\textbf{R}]"?} \\
    \textit{Choose 0 if not, 1 if valid, or 2 if indeterminate.} \\[0.5em]
    \textbf{Response (R):} \textit{0, 1, or 2}
    };
\end{tikzpicture}
}
\caption{Template for validating specific physical rules in a video.}
\label{tikz:rules}
\end{figure*}

\section{Poor physical commonsense qualitative examples by model}
\label{app:poor_physical_commonsense_model}

\begin{figure*}[htbp]
    \centering
    \includegraphics[width=\textwidth]{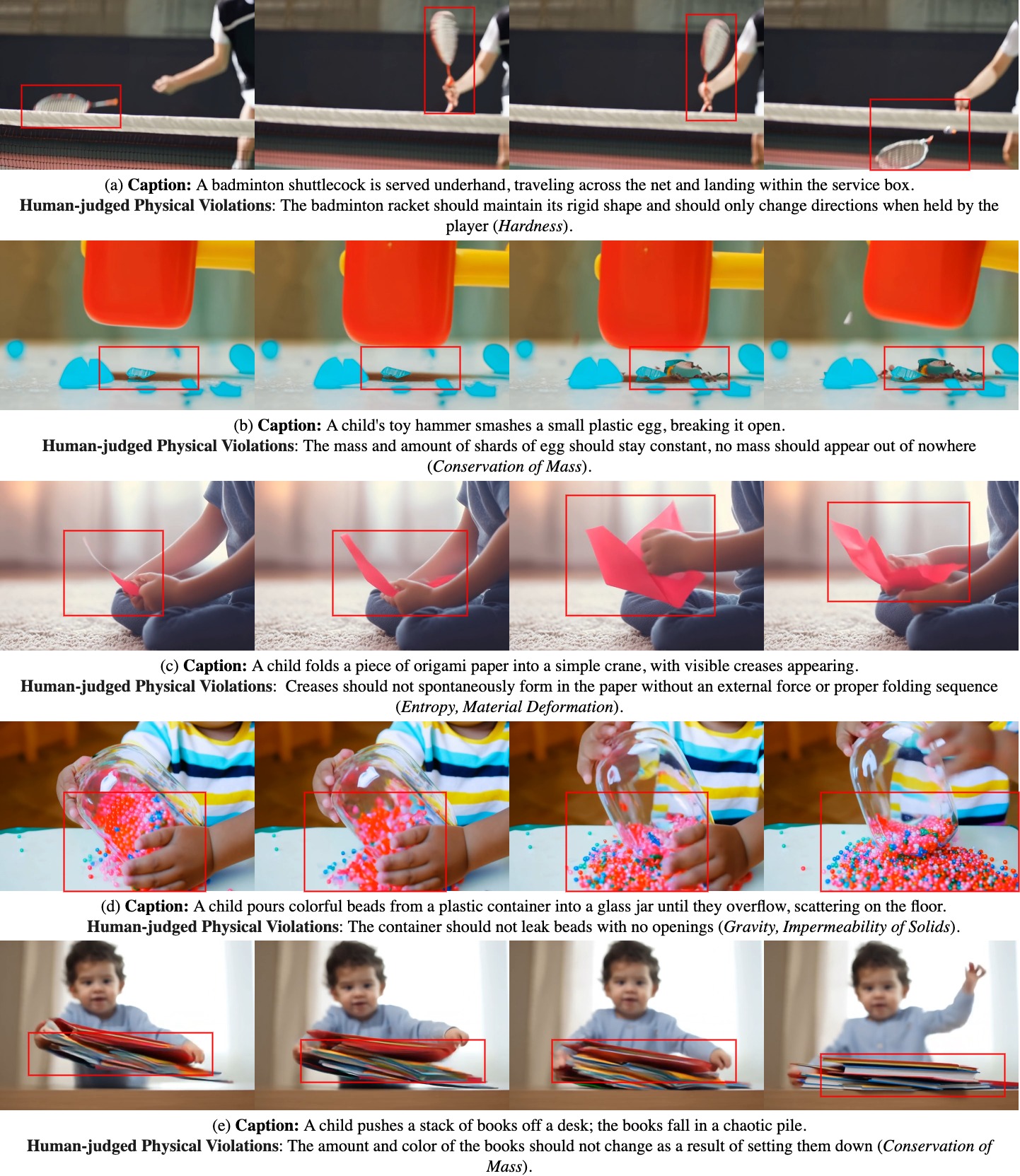}
    \caption{Examples of physically unlikely video generations from CogVideoX-5B. Each case demonstrates violations of fundamental physical laws.}
    \label{fig:cogvideo_full_bad_examples}
\end{figure*}

\begin{figure*}[htbp]
    \centering
    \includegraphics[width=\textwidth]{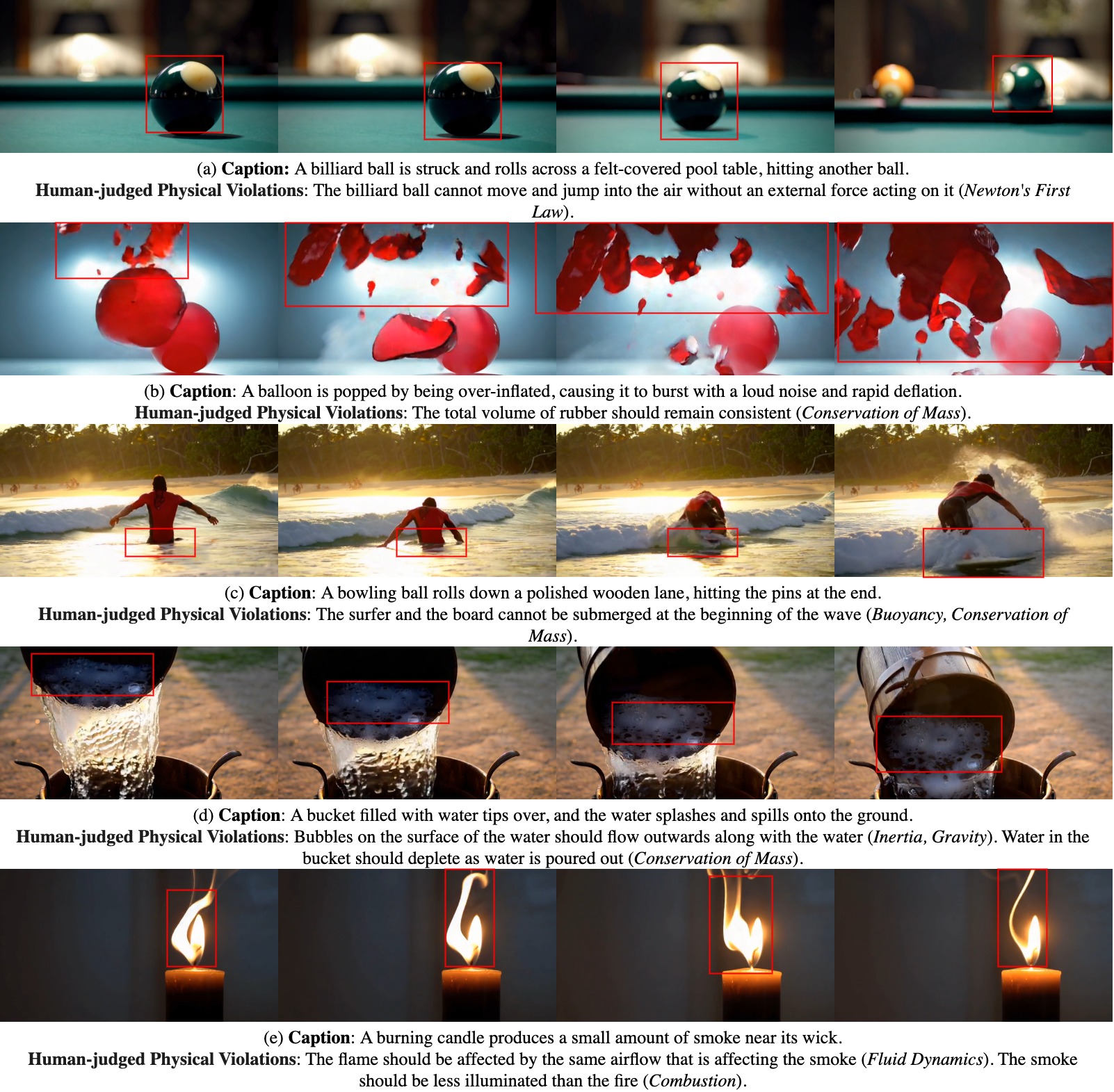}
    \caption{Examples of physically unlikely video generations from Cosmos. Each case demonstrates violations of fundamental physical laws.}
    \label{fig:cosmos_full_bad_examples}
\end{figure*}

\begin{figure*}[htbp]
    \centering
    \includegraphics[width=\textwidth]{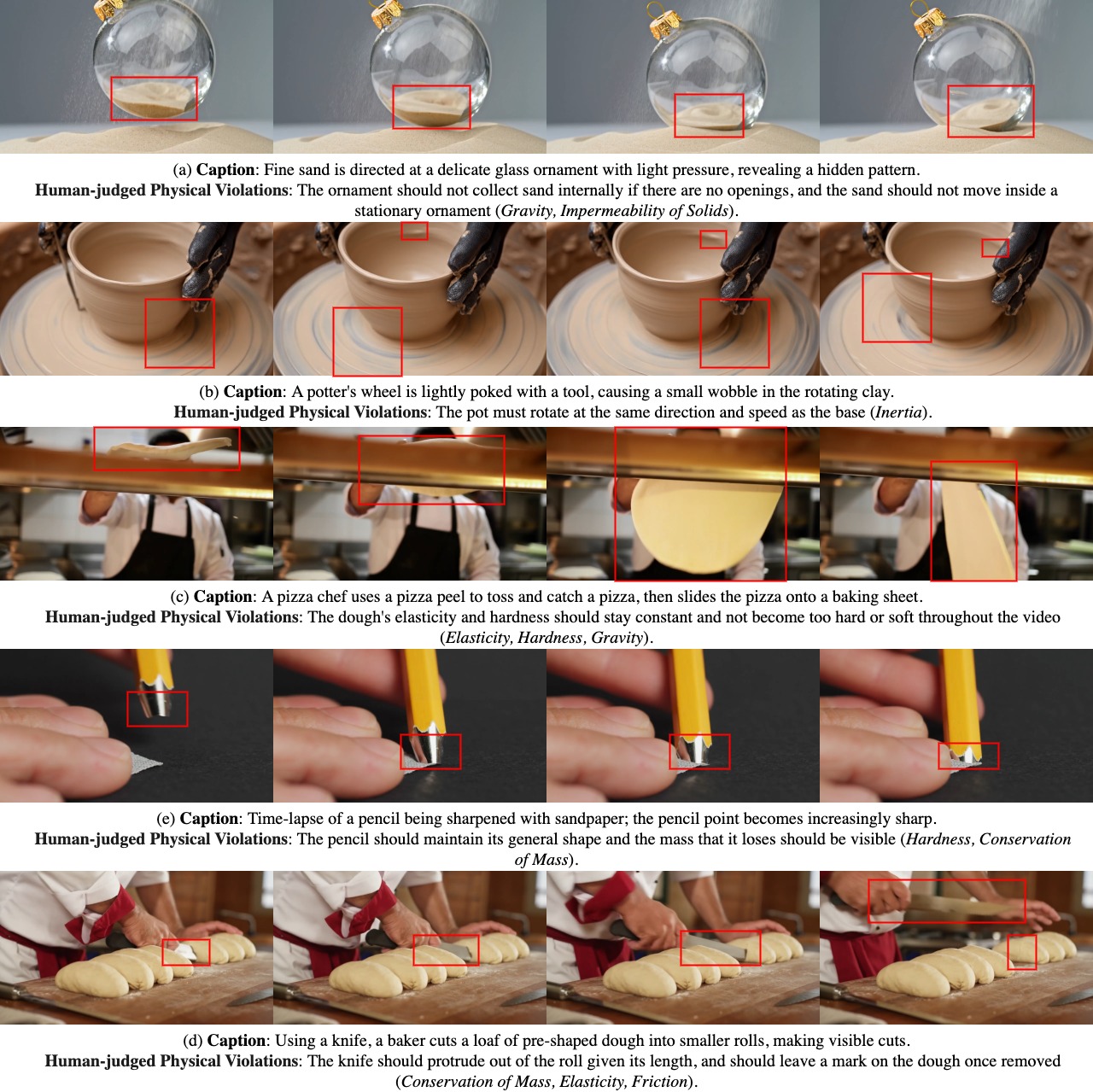}
    \caption{Examples of physically unlikely video generations from Ray2. Each case demonstrates violations of fundamental physical laws.}
    \label{fig:ray2_full_bad_examples}
\end{figure*}

\begin{figure*}[htbp]
    \centering
    \includegraphics[width=\textwidth]{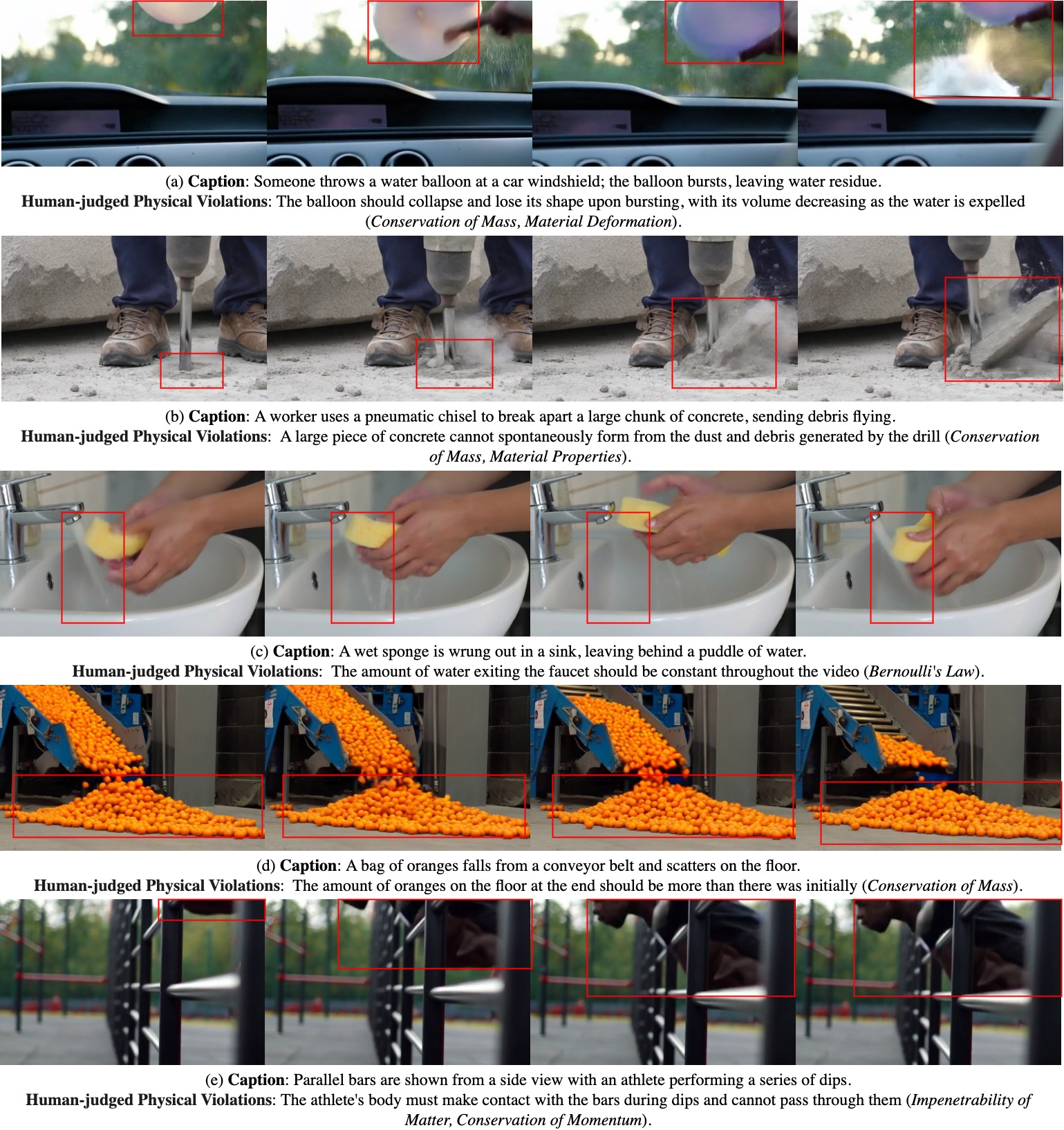}
    \caption{Examples of physically unlikely video generations from Hunyuan. Each case demonstrates violations of fundamental physical laws.}
    \label{fig:hunyuan_full_bad_examples}
\end{figure*}

\begin{figure*}[htbp]
    \centering
    \includegraphics[width=\textwidth]{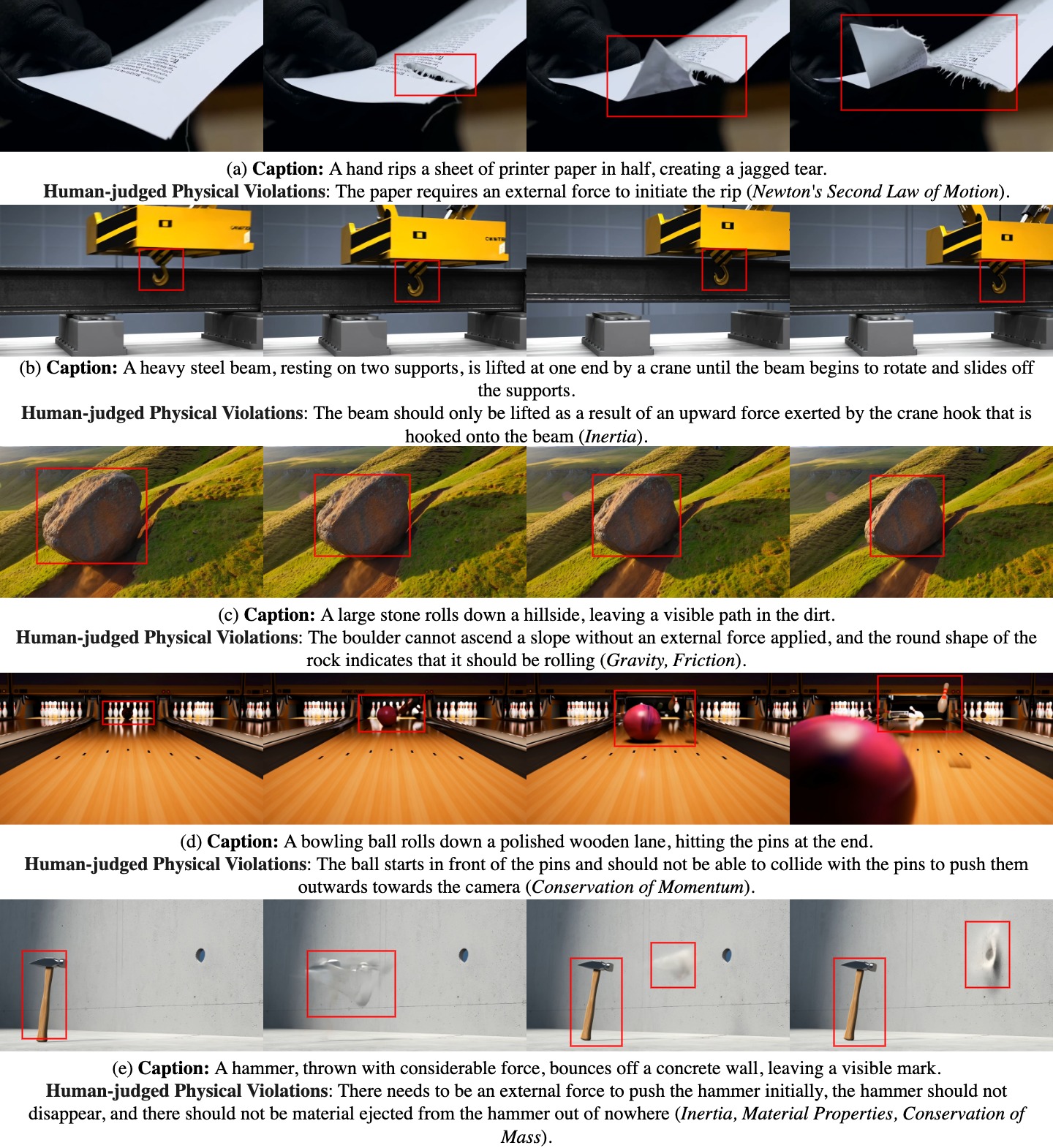}
    \caption{Examples of physically unlikely video generations from Wan2.1. Each case demonstrates violations of fundamental physical laws.}
    \label{fig:wan_full_bad_examples}
\end{figure*}

\begin{figure*}[htbp]
    \centering
    \includegraphics[width=\textwidth]{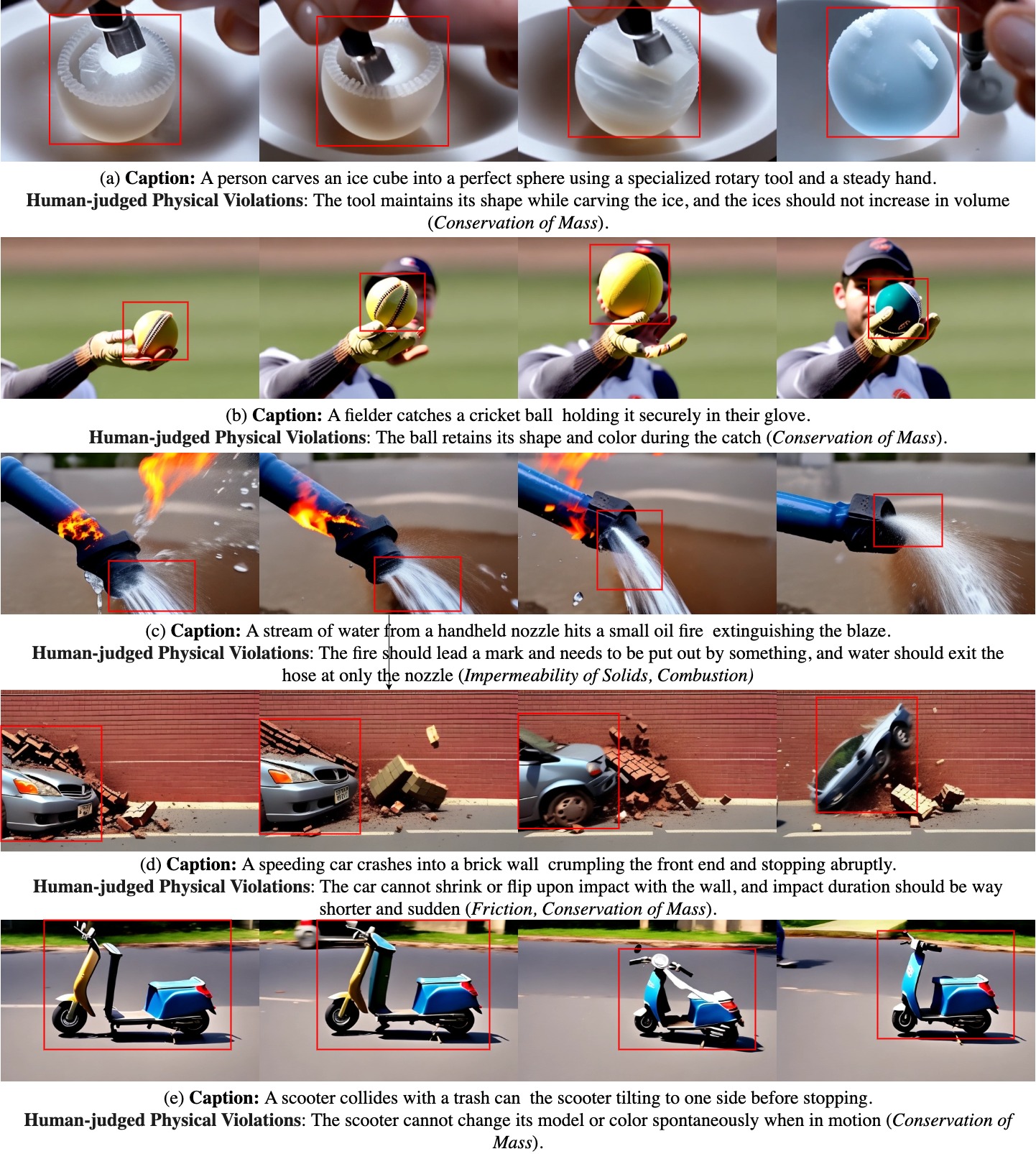}
    \caption{Examples of physically unlikely video generations from VideoCrafter2. Each case demonstrates violations of fundamental physical laws.}
    \label{fig:videocrafter_full_bad_examples}
\end{figure*}

We present more examples from each generative model where one or more physical laws are violated in Figure \ref{fig:cogvideo_full_bad_examples} - Figure \ref{fig:videocrafter_full_bad_examples}.

\section{Poor physical commonsense qualitative examples by law}
\label{app:poor_physical_commonsense_law}

\begin{figure*}[htbp]
    \centering
    \includegraphics[width=\textwidth]{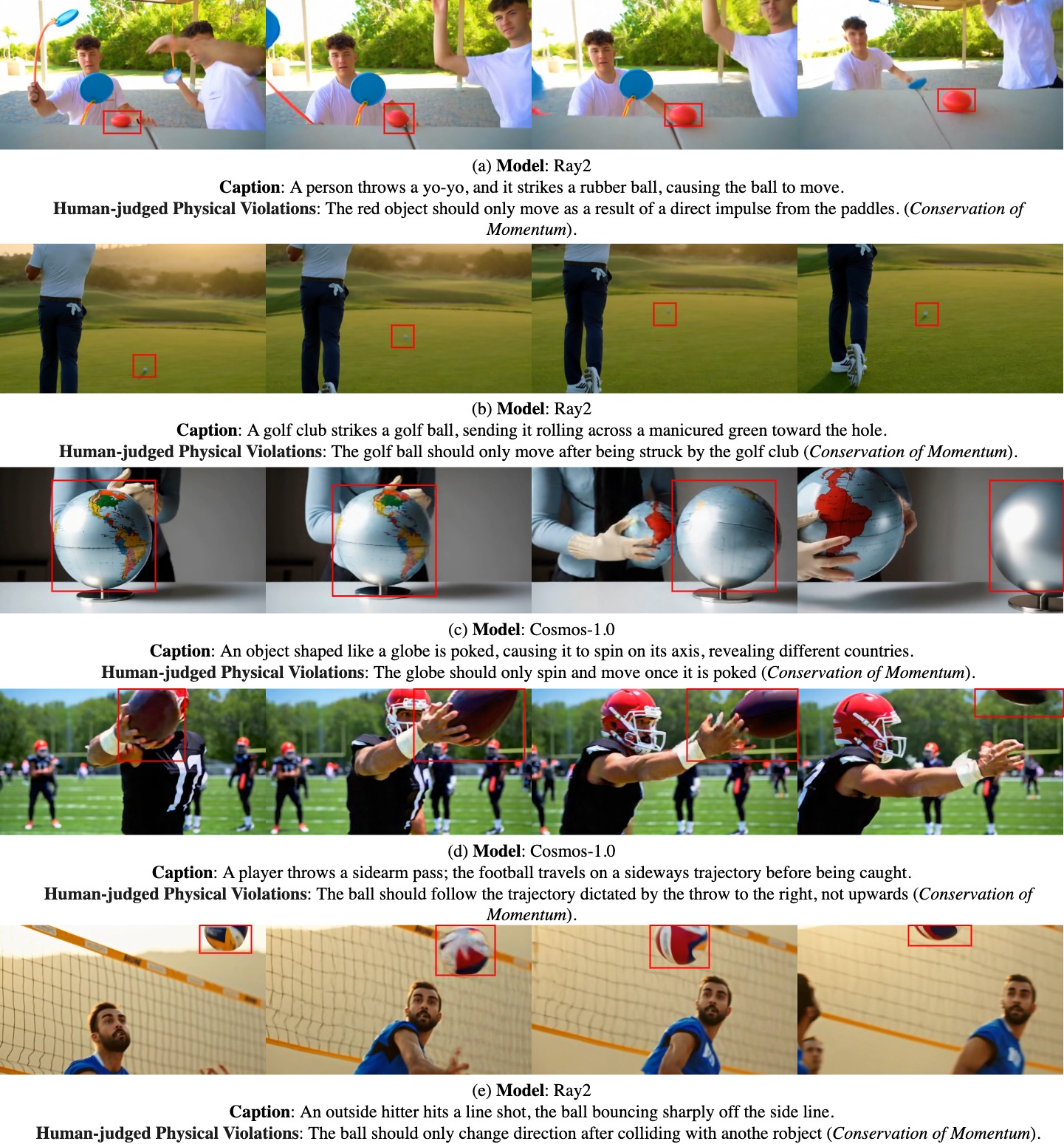}
    \caption{Examples of physically unlikely video generations where the physical law of conservation of momentum is violated}
    \label{fig:conservationmomentum_full_bad_examples}
\end{figure*}

\begin{figure*}[htbp]
    \centering
    \includegraphics[width=\textwidth]{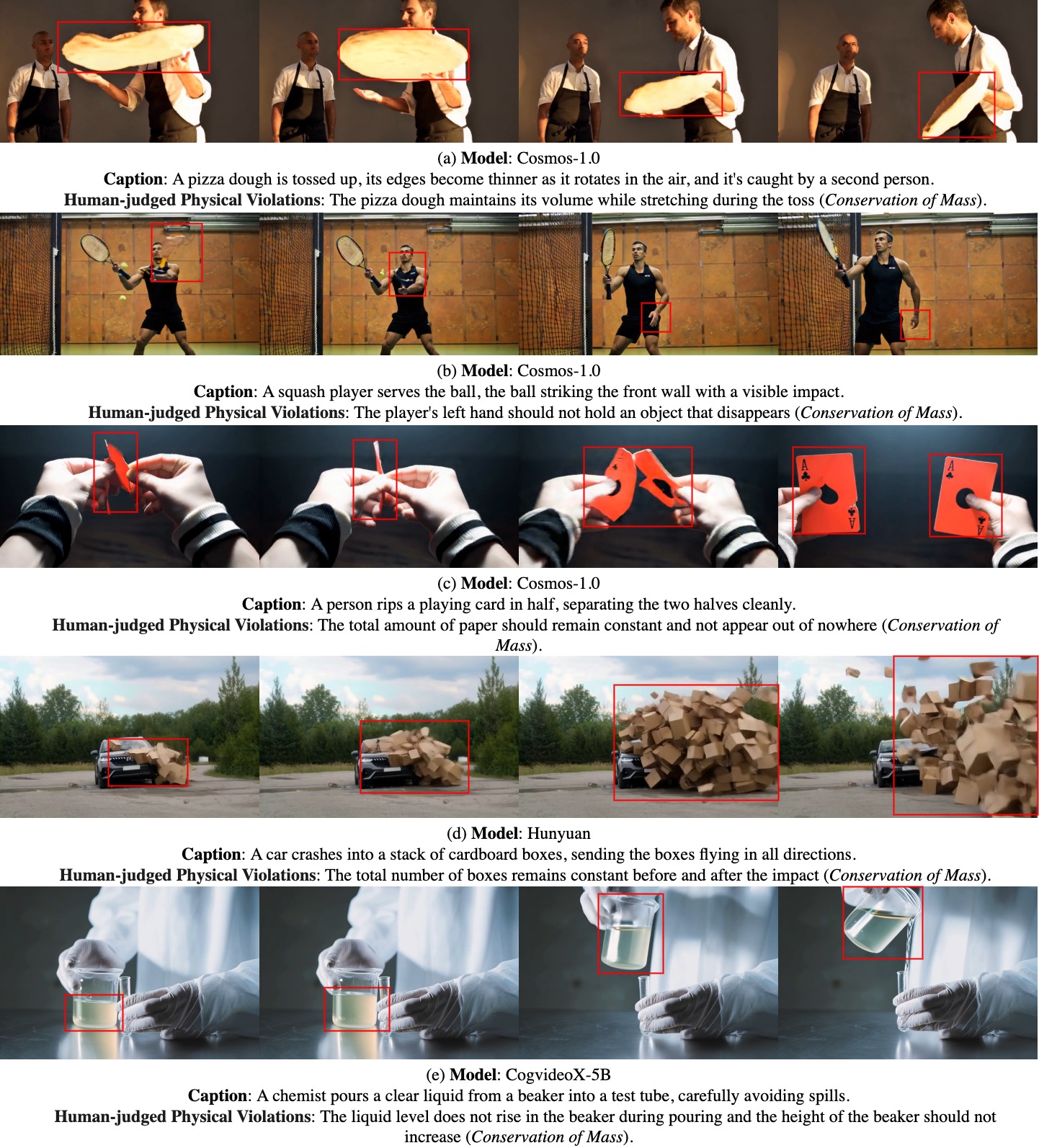}
    \caption{Examples of physically unlikely video generations where the physical law of conservation of mass is violated}
    \label{fig:conservationmass_full_bad_examples}
\end{figure*}

\begin{figure*}[htbp]
    \centering
    \includegraphics[width=\textwidth]{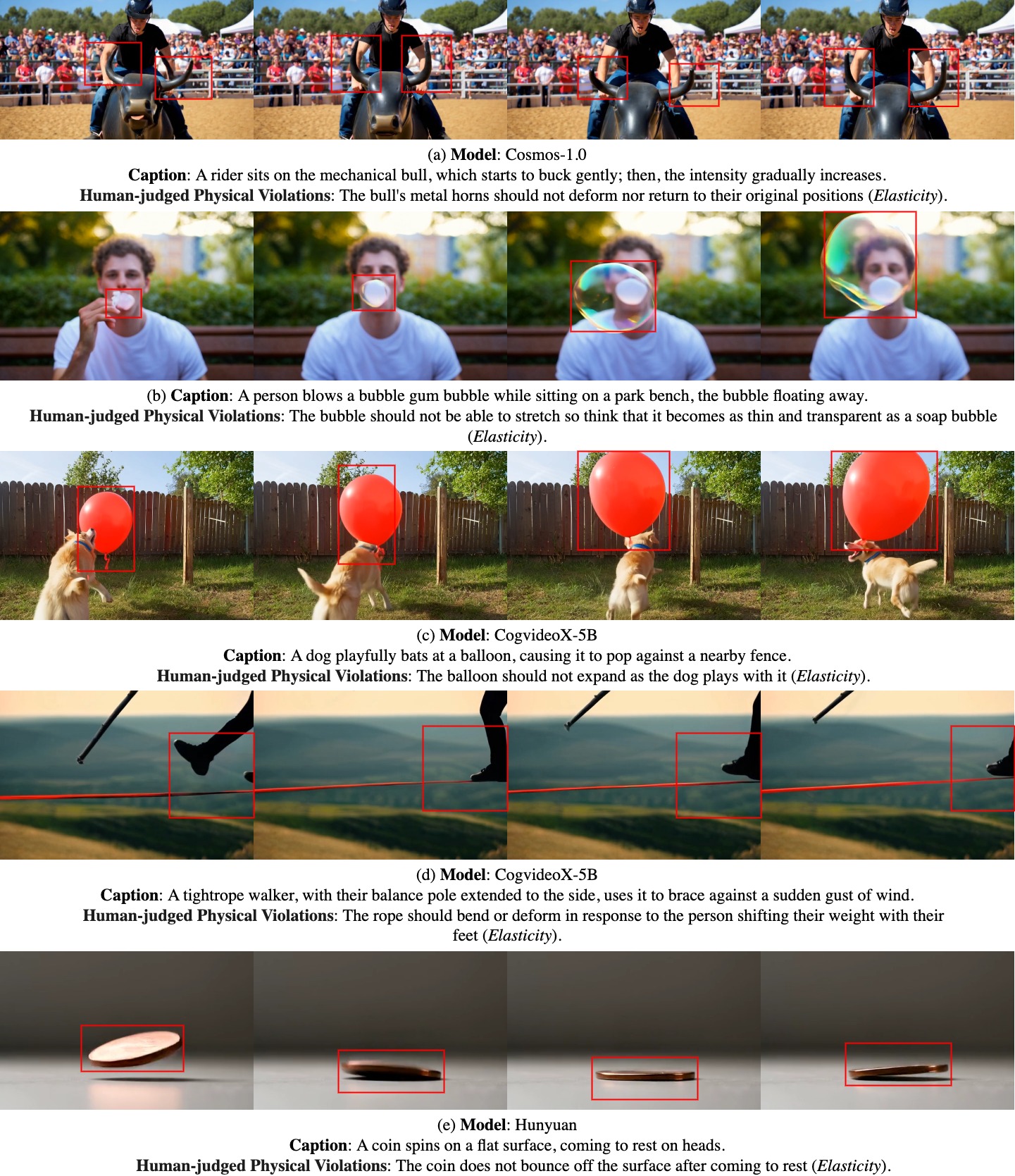}
    \caption{Examples of physically unlikely video generations where the physical law of elasticity is violated}
    \label{fig:elasticity_full_bad_examples}
\end{figure*}

\begin{figure*}[htbp]
    \centering
    \includegraphics[width=\textwidth]{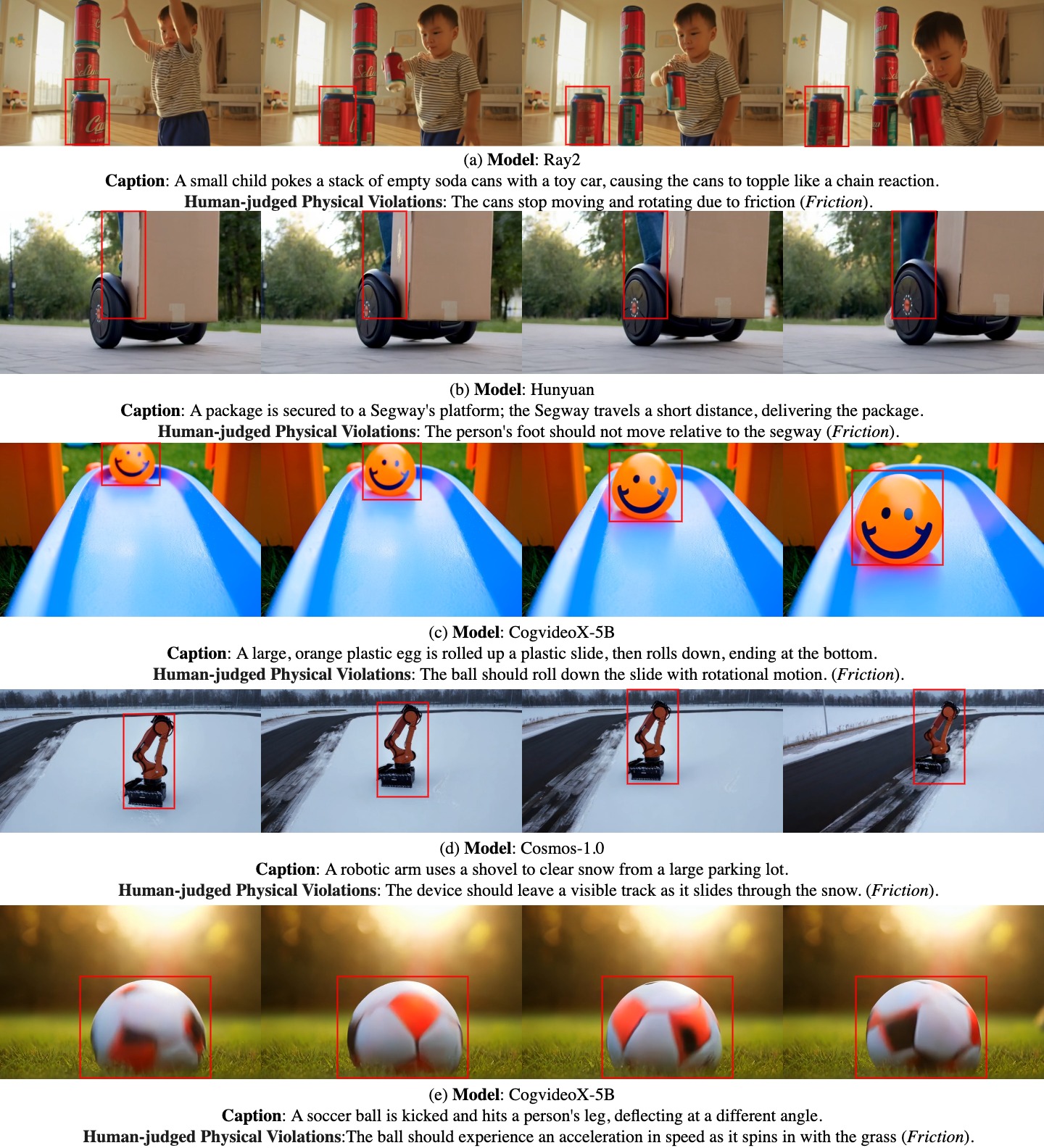}
    \caption{Examples of physically unlikely video generations where the physical law of friction is violated}
    \label{fig:friction_full_bad_examples}
\end{figure*}

\begin{figure*}[htbp]
    \centering
    \includegraphics[width=\textwidth]{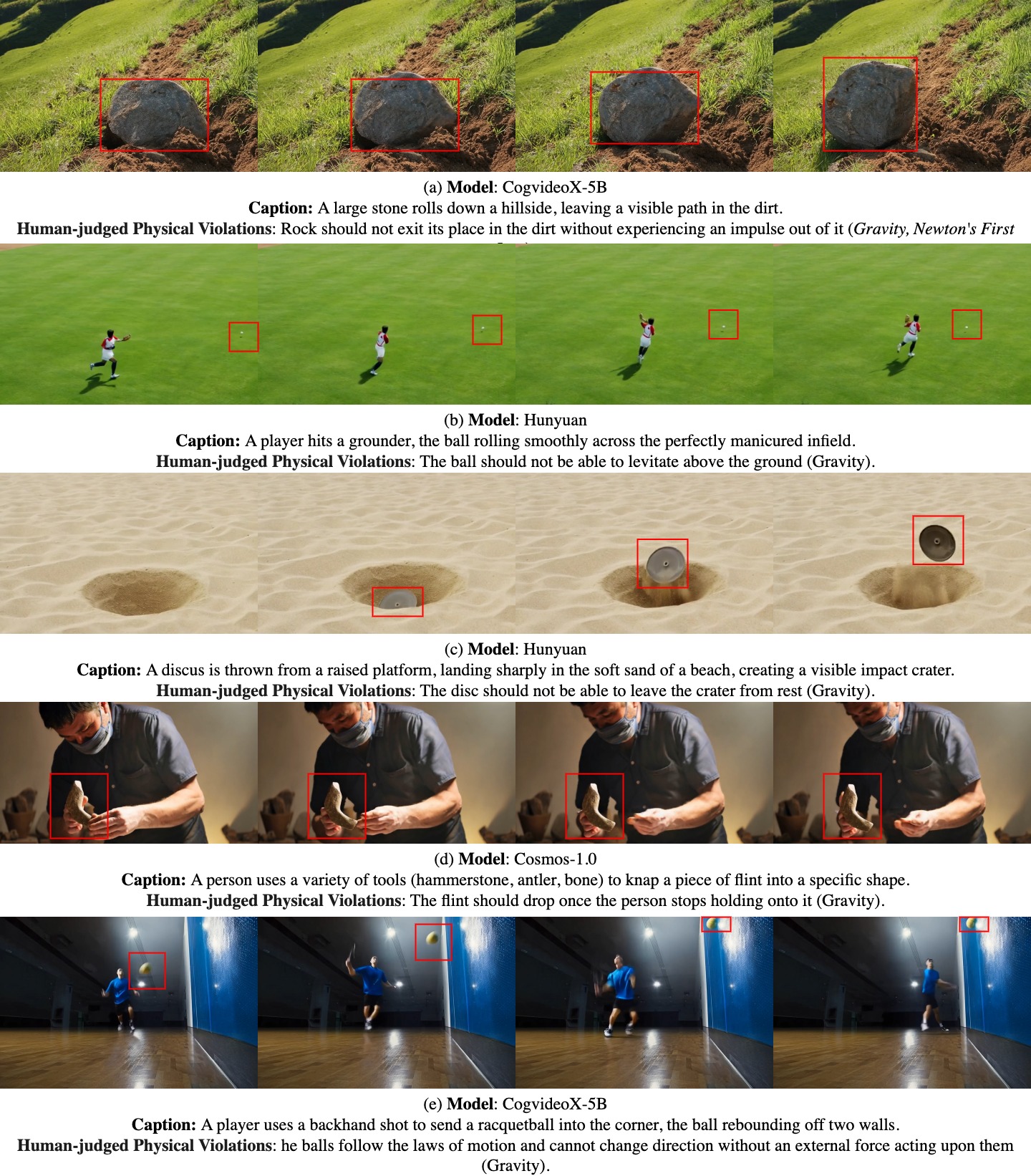}
    \caption{Examples of physically unlikely video generations where the physical law of gravity is violated}
    \label{fig:gravity_full_bad_examples}
\end{figure*}

\begin{figure*}[htbp]
    \centering
    \includegraphics[width=\textwidth]{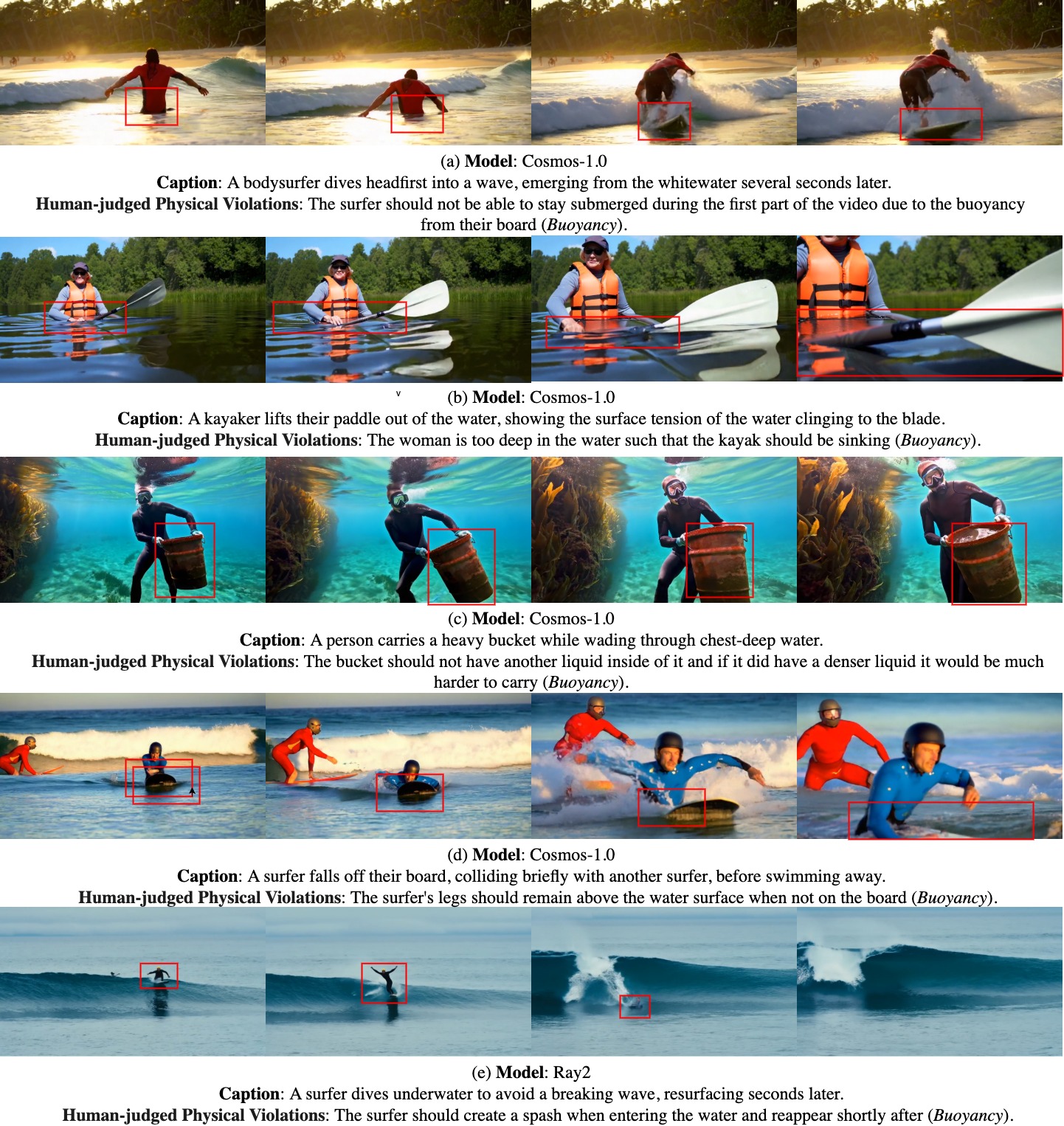}
    \caption{Examples of physically unlikely video generations where the physical law of Buoyancy is violated}
    \label{fig:buoyancy_full_bad_examples}
\end{figure*}

\begin{figure*}[htbp]
    \centering
    \includegraphics[width=\textwidth]{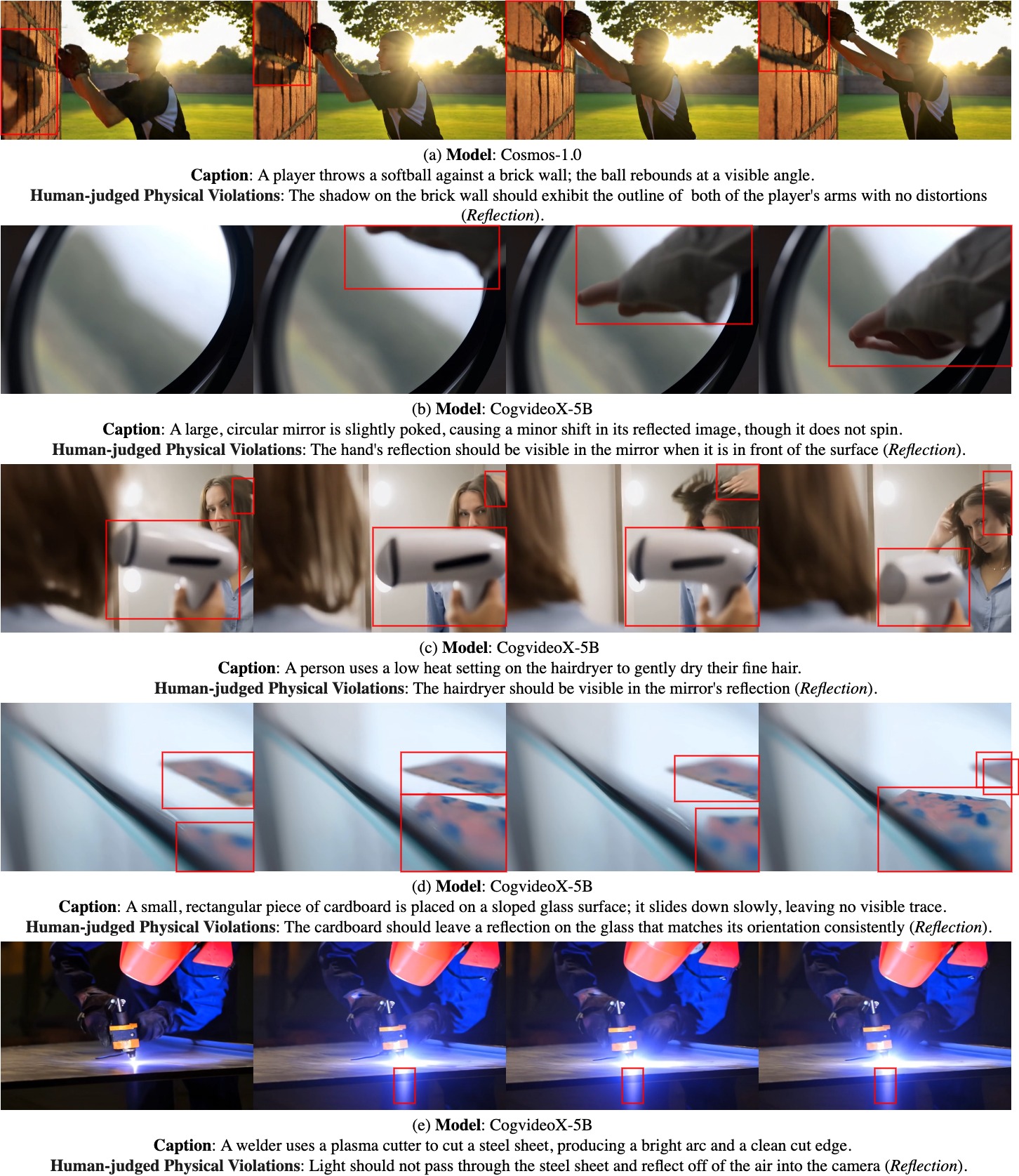}
    \caption{Examples of physically unlikely video generations where the physical law of reflection is violated}
    \label{fig:reflection_full_bad_examples}
\end{figure*}

\begin{figure*}[htbp]
    \centering
    \includegraphics[width=\textwidth]{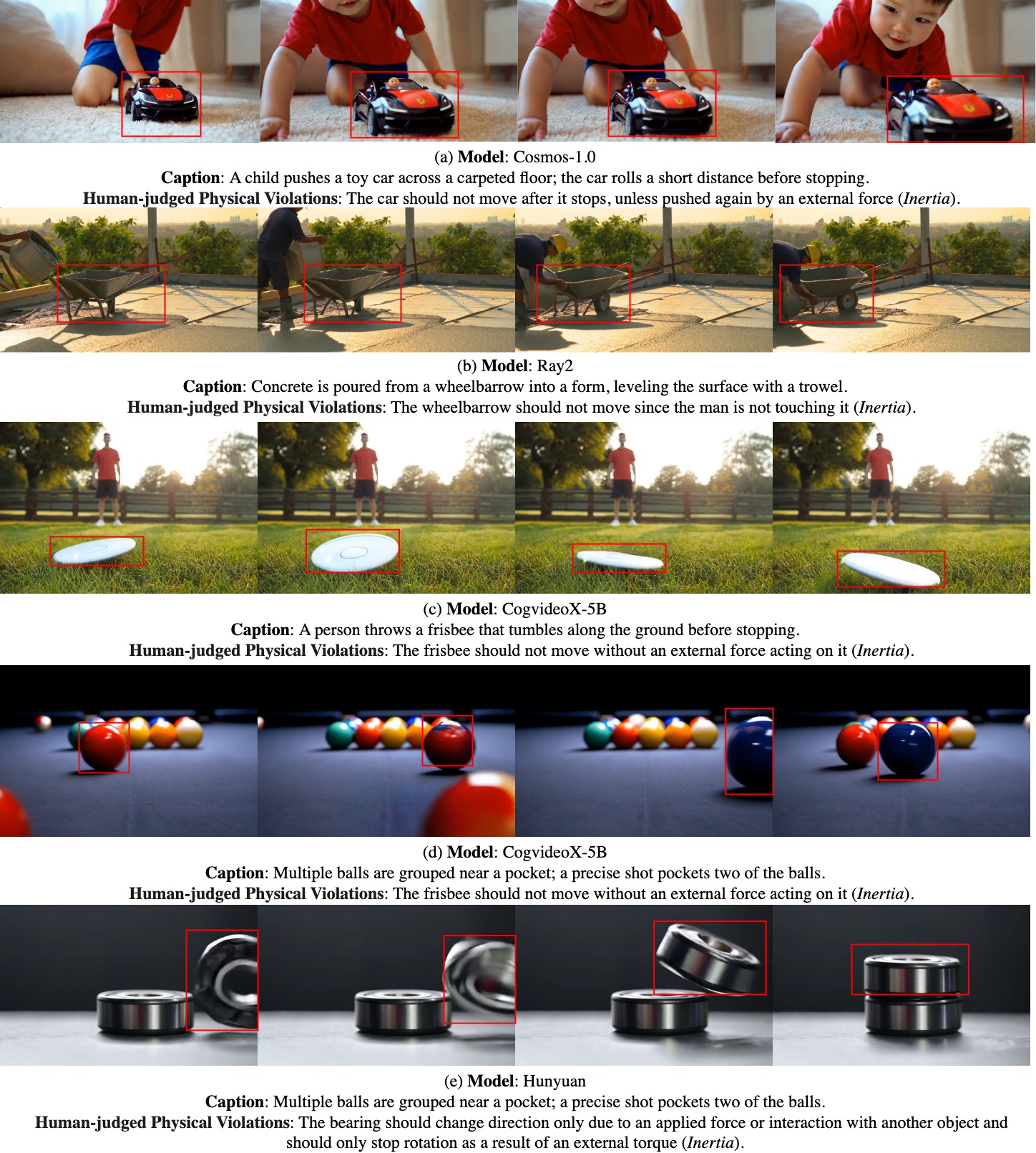}
    \caption{Examples of physically unlikely video generations where the physical law of inertia is violated}
    \label{fig:inertia_full_bad_examples}
\end{figure*}

We present a few qualitative examples highlighting instances where specific physical laws are violated in Figure \ref{fig:conservationmomentum_full_bad_examples} - Figure \ref{fig:inertia_full_bad_examples}.

\end{document}